\documentclass{article}
\usepackage[utf8]{inputenc}

\usepackage{microtype}
\usepackage{graphicx}
\usepackage{subcaption}
\usepackage{booktabs}
\usepackage[table]{xcolor}
\usepackage{colortbl}

\usepackage{amsmath, amssymb}
\usepackage{booktabs}
\usepackage{enumitem}
\usepackage{natbib}

\usepackage{times}
\usepackage{multirow}
\usepackage{hyperref}

\usepackage{pdfpages}
\usepackage{stfloats}
\usepackage{placeins}
\definecolor{criticalred}{RGB}{255,230,230}
\usepackage{hyperref}

\usepackage[preprint]{icml2026}

\usepackage{amsmath}
\usepackage{amssymb}
\usepackage{mathtools}
\usepackage{amsthm}

\definecolor{headerblue}{RGB}{66, 133, 244}
\definecolor{lightblue}{RGB}{232, 240, 254}
\definecolor{lightgray}{RGB}{248, 249, 250}
\definecolor{c1green}{RGB}{52, 168, 83}
\definecolor{c2blue}{RGB}{66, 133, 244}
\definecolor{c3orange}{RGB}{251, 188, 4}
\definecolor{c4red}{RGB}{234, 67, 53}

\usepackage[capitalize,noabbrev]{cleveref}

\theoremstyle{plain}

\theoremstyle{definition}

\theoremstyle{remark}

\usepackage[disable]{todonotes}

\icmltitlerunning{Representation as a Bottleneck for Mechanistic Interpretability}
\hypersetup{
  pdftitle={Representation as a Bottleneck for Mechanistic Interpretability: The Manifestation Unit Protocol},
  pdfauthor={Hussein Chouman, Wataru Sasaki, Tomokazu Matsui, Hirohiko Suwa, Keiichi Yasumoto}
}
\begin{document}

\twocolumn[
  \icmltitle{Representation as a Bottleneck for Mechanistic Interpretability:\\
The Manifestation Unit Protocol}

  \begin{icmlauthorlist}
    \icmlauthor{Hussein Chouman}{naist}
    \icmlauthor{Wataru Sasaki}{naist}
    \icmlauthor{Tomokazu Matsui}{naist}
    \icmlauthor{Hirohiko Suwa}{naist}
    \icmlauthor{Keiichi Yasumoto}{naist}
  \end{icmlauthorlist}

  \icmlaffiliation{naist}{Ubiquitous Computing Systems Laboratory, Nara Institute of Science and Technology, Nara, Japan}

  \icmlcorrespondingauthor{Hussein Chouman}{chouman.hussein.ck5@naist.ac.jp}

  \icmlkeywords{Machine Learning, ICML}

  \vskip 0.3in
]

\printAffiliationsAndNotice{}

\begin{abstract}
Mechanistic interpretability has produced a rich inventory of
component-level analyses that characterise what neural-network
components encode and how they interact.
Their outputs, however, are not easily reusable: selectivity tables,
circuit diagrams, and feature lists remain locked in per-study
notebooks---non-composable, not queryable in natural language, and
not directly actionable for downstream audit or intervention.
We study the \emph{representation layer} that sits between these
analyses and downstream use as a bottleneck that can be evaluated
independently and introduce
\textbf{Manifestation Units}, a typed tuple protocol $(E, S, R, D, G)$ extended with attention-head primitives $(T)$ for transformer architectures
organising per-component statistics into structured fields populated
automatically and queried through hybrid retrieval.
Instantiated across generative vision ($\beta$-VAE), discriminative
vision (CNN), and language (GPT-2), the protocol supports two
findings: typed structure substantially outperforms unstructured
baselines on retrieval, and CNN filters retrieved by the schema
satisfy causal sufficiency and necessity criteria under matched-budget
controls.
The schema absorbs attention-head primitives without modification,
set-recovers known IOI circuit members under retrieval-budget-matched
controls, and reveals an irreducible two-field core ($S{+}R$) with
remaining fields either redundant or actively interfering.
We present this as schema infrastructure for mechanistic
interpretability rather than frontier-scale validation.
\footnote{Transformer and CNN demos are available at \href{https://manifestation-xai.github.io/manifestation-transformers/}{https://manifestation-xai.github.io/manifestation-transformers/}\\ and \href{https://manifestation-xai.github.io/manifestation-cnn/}{https://manifestation-xai.github.io/manifestation-cnn/}.}
\end{abstract}

\section{Introduction}

\label{sec:introduction}

Mechanistic interpretability has produced a rich inventory of
analysis techniques for characterising what neural-network components
encode and how they interact: Network Dissection, probing classifiers,
sparse-autoencoder decompositions, and circuit
analyses~\citep{bau2017network, belinkov2022probing,
lieberum2024gemmascope, elhage2021framework, wang2023ioi,
olsson2022induction, conmy2023acdc}. Their outputs, however, are not
easily reusable. Selectivity tables, circuit diagrams, and SAE
feature lists remain in per-study notebooks and scripts: not
composable, not queryable in natural language, and not directly
actionable for downstream audit or intervention.

A practitioner asks of a trained CNN: \emph{``which filters encode 
`dog', and what happens if I amplify them?''} The needed 
statistics---selectivity, activation correlations, ablation 
effects---already exist, but there is no grounded interface to ask. 
Semantic search fails on numerical identifiers~\citep{li2023santa}, 
exact-match fails on conceptual queries, and free-form prose loses 
the typing that separates ``which filters detect dog'' from ``what 
does filter 81 detect.''

We study the representation layer that sits between component-level
analyses and downstream use, holding the analyses themselves fixed.
We introduce \emph{Manifestation Units}, a typed tuple protocol
$(E,S,R,D,G,T)$ that organises per-component statistics
into six fields aligned with distinct query primitives: entity
identity ($E$), semantic associations ($S$), cross-component
relations ($R$), quantified dynamics ($D$), intervention guidance
($G$), and attention-head primitives ($T$). Each field is extracted
automatically from activations and validation-set statistics; no
human annotation is involved. A hybrid retriever combines
exact-matching on $E$ with dense semantic search over the remaining
fields.

We evaluate two hypotheses about this representation layer. 
\textbf{H1 (structural accessibility).} Natural-language retrieval 
over component-level analyses requires typed \emph{structure}, not 
just the underlying content---random partitions of the same content 
fail to match the ESRDGT decomposition.
\\
\textbf{H2 (causal mediation, minimality, and optimality).}
On the CNN, components retrieved through the structured
representation are causal mediators of the targeted behaviour
in the MIB sense~\citep{mueller2024mib}, established by
sufficiency and necessity under matched-budget controls.
On the GPT-2 schema, a weaker analogue replaces strict causal
mediation: the decomposition admits a \emph{minimal-optimal}
core ($S{+}R$) confirmed via random-partition controls, with
remaining fields either redundant or actively interfering;
see \S\ref{sec:rq4} and App.~\ref{app:path_patching}.

\paragraph{Empirical findings.}
We instantiate the protocol across three architectures spanning
generative vision ($\beta$-VAE on CelebA), discriminative vision
(CNN on CIFAR-10), and language (GPT-2 attention heads). On the
GPT-2 schema, the $S{+}R$ subset reaches oracle recall@30 of
$0.411$, beating random partitions of identical content at $p<0.01$
(Fig.~\ref{fig:ablation_validation}D; H1); per-field ablation
isolates $S$ ($p<10^{-10}$) and $R$ ($p{=}.019$) as the
\emph{minimal-optimal core}, $G$ as redundant within that subset,
and $E$/$D$ as actively interfering when added to the full schema
(Fig.~\ref{fig:ablation_validation}A--C), with the two necessary
fields retrieving weakly anti-correlated head rankings
($\rho{=}{-}0.20$; Fig.~\ref{fig:redundancy_heatmaps}B) and
thereby complementing one another (H2, minimality and mechanism).
On the CNN, retrieved filters flip predictions in $76.3\%$ of trials
under amplification (vs.\ $6.7\%$ random matched-budget) and zeroing
them drops natural target-class accuracy by $38.3$pp (vs.\ $3.6$pp);
a four-way decomposition attributes $+17.4$pp of sufficiency to the
typed primary-class rule (H2, mediation). On the $\beta$-VAE and
CNN, structured units reach Precision@5 of $89.6\%$ and $96.8\%$
respectively, versus $18.0\%$ and $20.4\%$ for a BM25-hybrid
baseline over the same per-component statistics (H1, original
evaluation). On GPT-2, retrieved heads beat matched-budget random
retrieval by $2.0$--$4.5\times$ across four behaviours and recover
$5/7$ of the known IOI L7--9 name-mover heads at $k{=}30$.

\paragraph{Contributions.}
\begin{enumerate}[leftmargin=*, itemsep=0pt, topsep=2pt]
\item \textbf{Manifestation Units (ESRDG[T])}, a typed tuple protocol
for organising component-level analyses, with automatic extraction,
deterministic template grounding, and hybrid retrieval. The base
schema is $(E,S,R,D,G)$; the $T$ field extends it for transformer
architectures.
\item Matched-budget protocols establishing structural accessibility
(H1, random-partition controls) and the H2 cluster of causal mediation,
minimality, and optimality (CNN sufficiency/necessity; GPT-2 per-field
ablation)---identifying the \emph{minimal-optimal} core ($S{+}R$) and
characterising redundancy ($G$ within the core) and interference
($E$, $D$ on the full schema).
\item A GPT-2 attention-head instantiation that supplies the deepest
structural evidence---random-partition H1 confirmation, the
$S{+}R$ minimal-optimal core, and complementary $S$/$R$ rankings
($\rho{=}{-}0.20$)---and recovers known IOI circuit members via
set-overlap retrieval, with a principled $S$-primitive selection
procedure validated by reference-set recovery.
\end{enumerate}

\section{Related Work}
\label{sec:related}

\paragraph{Mechanistic interpretability and faithfulness.}
The MI toolkit characterises neural-component behaviour through 
complementary lenses: circuit analyses describe how components 
compose into algorithms~\citep{elhage2021framework, wang2023ioi, 
olsson2022induction, elhage2022toymodels}; activation and path 
patching trace causal pathways through 
transformers~\citep{meng2022rome, conmy2023acdc, goldowsky2023patch}; 
sparse autoencoders decompose polysemantic activations into 
monosemantic features at scale~\citep{lieberum2024gemmascope, 
cunningham2024sae}; Network Dissection and probing classifiers 
characterise what individual components 
encode~\citep{bau2017network, belinkov2022probing}. The MIB 
benchmark~\citep{mueller2024mib} formalises the two faithfulness 
notions our evaluation adopts (Sec.~\ref{sec:evaluation}): 
\emph{response faithfulness}---does the output reflect the 
evidence?---and \emph{causal faithfulness}---do the identified 
components causally mediate the behaviour, established by 
sufficiency and necessity under matched intervention?

\paragraph{Representation and retrieval over MI outputs.}
Given any of these analyses---SAE feature activations, circuit 
identifications, attribution-graph neighbourhoods, probing scores, 
correlational selectivity---how should the outputs be packaged so 
they can be queried, grounded, composed, and acted on? Existing 
community tooling targets \emph{programmatic access} 
(TransformerLens, nnsight) and \emph{visualisation} (Neuronpedia, 
Docent), leaving the typed-representation layer for natural-language 
querying largely unaddressed. \citet{orgad2024actionable} document 
that interpretability findings are frequently not acted on 
downstream; representation is part of the gap. Standard semantic 
search fails on neural-component queries because embedding models 
cannot reliably represent arbitrary numerical identifiers like 
``dimension 63''~\citep{li2023santa}, and exact match alone cannot 
handle conceptual queries. Structured and entity-aware 
retrieval~\citep{edge2024graphrag, fevry2020entities, 
yasunaga2022kgqa} shows that typed representations outperform 
unstructured prose for knowledge-intensive tasks, but targets 
general-domain knowledge rather than neural internals indexed by 
arbitrary identifiers. Manifestation Units are designed to compose 
with the MI toolkit above: $S$ can store SAE feature labels, $R$ 
can encode head--head or attribution-graph neighbourhoods, and $G$ 
can encode activation-patching or steering primitives. The 
Transformer instantiation (Sec.~\ref{sec:transformer-pilot}) 
demonstrates this composition on attention-head primitives drawn 
from \citet{elhage2021framework, olsson2022induction, wang2023ioi, 
goldowsky2023patch, heimersheim2024patching}.

\section{The MANIFESTATION Framework}
\label{sec:framework}

\begin{figure*}[t]
\centering
\includegraphics[width=0.95\linewidth]{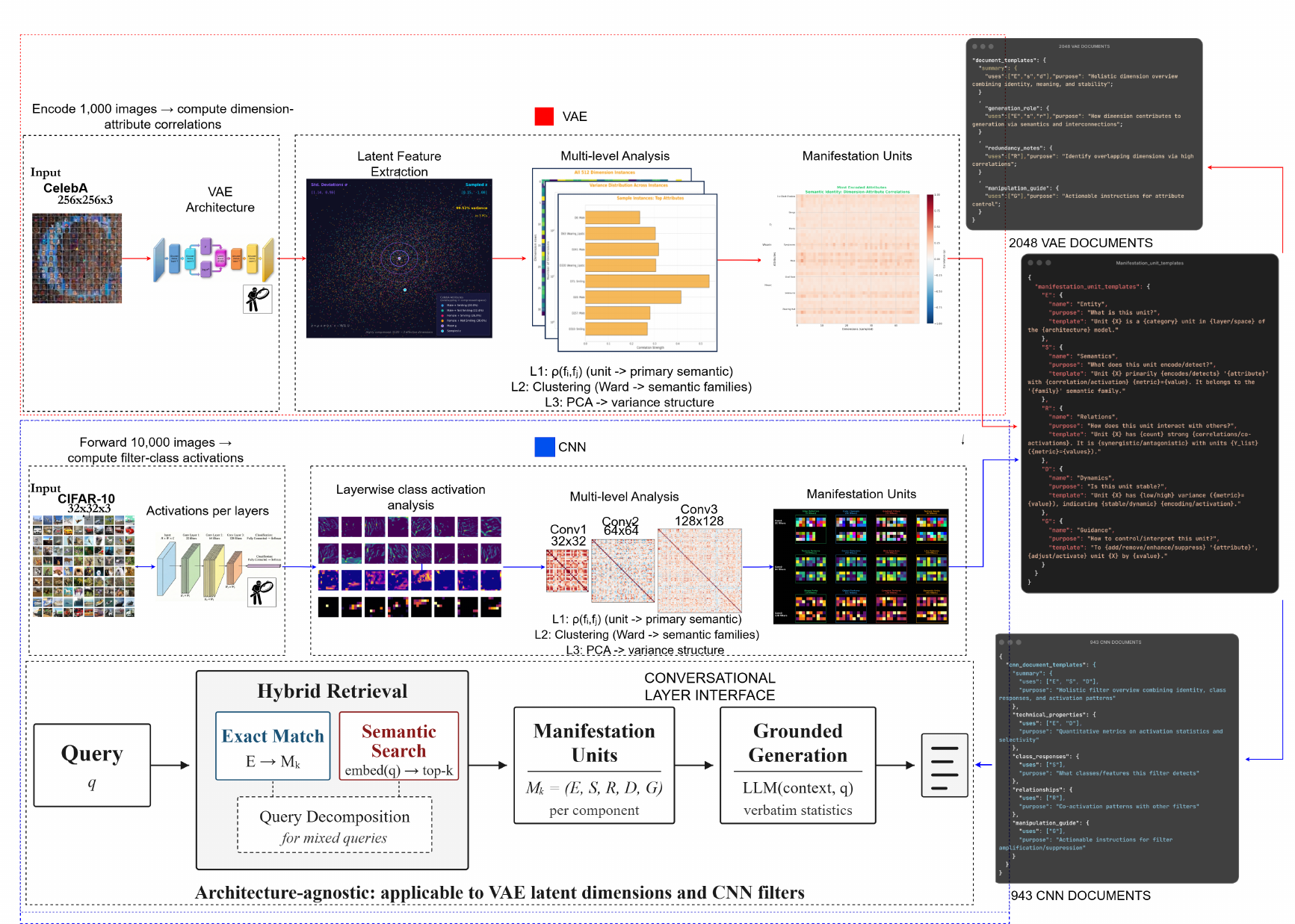}
\caption{Overview of MANIFESTATION. The pipeline extracts
Manifestation Units $(E, S, R, D, G, T)$ encoding entity identity
($E$), semantic associations ($S$), cross-component relationships
($R$), quantified dynamics ($D$), manipulation guidance ($G$), and
(for transformers) attention-head primitives ($T$). Hybrid retrieval
combines exact matching on $E$ with semantic search over the
remaining fields; the SLM generates responses grounded in retrieved
statistics from $D$.}
\label{fig:framework}
\end{figure*}

MANIFESTATION (Fig.~\ref{fig:framework}) populates Manifestation
Units automatically from model activations and exposes them through
a hybrid retriever for natural-language access.

\subsection{Manifestation Units}
\label{sec:mu-def}
Let $\mathcal{I}=\{1,\dots,n\}$ denote the set of neural network
components to be analysed (latent dimensions in a VAE, filters in a
CNN, attention heads in GPT-2). Each component $k\in\mathcal{I}$ is
represented as a structured tuple:
\begin{equation}
M_k = (E_k, S_k, R_k, D_k, G_k, T_k).
\end{equation}
The decomposition is functional: each field corresponds to a
recurring class of user query---$E$ for entity queries
(``what does dimension 63 do?''), $S$ for concept queries (``which
units encode smiling?''), $R$ for relational queries
(``what correlates with dimension 47?''), $D$ for grounded numeric
citation, $G$ for intervention queries, and $T$ for attention-head
primitives in transformer architectures. The six fields group
queries by the retrieval primitive each requires (typed identifier
matching, ranked semantic similarity, neighbourhood walks, scalar
lookup, parameterised intervention, attention-pattern features).
The decomposition is empirically minimal-optimal
(Sec.~\ref{sec:rq3}): $S{+}R$ is the highest-recall combination
across query types with both fields necessary within that subset;
$G$ is redundant, and $E$/$D$ actively interfere on the full schema.

\paragraph{Entity (E).}
$E_k$ is a unique, human-readable identifier (e.g., ``Dimension 63''
or ``Filter Conv2.15'') that enables entity-aware exact matching.
When a query contains an explicit component reference, pattern
extraction on $E$ guarantees retrieval of the correct unit---something
embedding-only semantic search cannot reliably achieve for arbitrary
identifiers.

\paragraph{Semantics (S).}
$S_k$ stores concept associations with quantified strengths over a
predefined concept vocabulary $\mathcal{C}$ (facial attributes for
CelebA, class labels for CIFAR-10, functional features for GPT-2).
Templates expose only the top-$m$ associations, ranked by magnitude:
\begin{equation}
\tilde{S}_k = \{(c_j, s_{kj})\}_{j=1}^{m} \ \text{ranked by}\ |s_{kj}|.
\end{equation}
The primary concept identity is the top-1 concept among the exposed
set:
\begin{equation}
\label{eq:primary-class}
c_k^{*}=\arg\max_{j\in[m]} |s_{kj}|,\quad s_k^{*}=s_{k,c_k^{*}}.
\end{equation}
The signed score $s_k^{*}$ is retained for faithful generation; the
absolute value is used only for ranking. $S_k$ is populated by 
Pearson correlation for VAE dimensions, entropy-based class 
selectivity for CNN filters, and attention-pattern primitives for 
GPT-2 heads (App.~\ref{app:extraction}).

\paragraph{Relationships (R).}
$R_k$ encodes cross-component dependencies derived from model
activations. Let $\rho_{ki}$ denote a dependency score between
components $k$ and $i$, and let
$t_{ki}\in\{\text{synergistic},\text{antagonistic},\text{independent}\}$
indicate relationship polarity. Templates expose a top-$r$
neighbourhood. We use these labels to describe statistical
co-activation, not causal mechanisms. Thresholds and tie-breaking
detailed in App.~\ref{app:extraction}.

\paragraph{Dynamics (D).}
$D_k$ is a typed dictionary of quantitative statistics (variance
contributions, selectivity summaries, cluster/family memberships,
cross-layer dependency scores). These values are rendered into the
LLM's context by deterministic templates
(Sec.~\ref{sec:hybrid-retrieval}), and a post-hoc check verifies
cited values against stored statistics.

\paragraph{Guidance (G).}
$G_k$ stores manipulation parameters (step sizes for VAE latent
edits in units of $\sigma$, amplification factors for CNN filter
activations). These parameters are pre-computed and retrieved,
grounding interventions in measured mappings rather than model
priors.

\paragraph{Attention-head primitives (T).}
For transformer architectures, $T_k$ stores attention-pattern
features---induction scores, copying scores, QK and OV circuit
norms, and previous/first-token attention ratios---computed from
the head's attention matrices over a reference corpus. Population
details in Sec.~\ref{sec:rq4} and App.~\ref{app:gpt2}.

\paragraph{Template-based grounding.}
Deterministic templates substitute field values directly, exposing
top-$m$/top-$r$ subsets; every rendered statement traces to a
stored field value (App.~\ref{app:template}).

\subsection{Hybrid Retrieval and Generation}
\label{sec:hybrid-retrieval}
The retriever combines exact-matching on $E$ (for entity-bearing
queries) with dense semantic search over templated $S$/$R$/$G$/$T$
documents (for conceptual queries); hits are aggregated at the unit
level before generation (App.~\ref{app:retrieval}).

\paragraph{Grounded generation.}
Generation is next-token prediction; faithfulness is enforced
upstream (deterministic retrieval and templating) and verified
downstream (post-hoc Factual Accuracy check, $>$94\% across 96
configs; App.~\ref{app:auto_eval}).

\subsection{Natural Language Control as a Causal Probe}
\label{sec:control-probe}
Beyond querying, the framework supports natural-language commands
that translate into targeted interventions. Given a command like
``make the face smile,'' the system identifies the target concept,
retrieves components with high association strength, and applies
modification parameters drawn from $G$. Retrieval here uses a typed
\emph{primary-class retrieval rule}: for target concept $c$, the
system selects only components whose top-1 entry in $\tilde{S}$ is
$c$, not those where $c$ merely appears in the top-$m$. This rule
is the central element of our H2 protocol (Sec.~\ref{sec:rq2}); it
is only definable because $S$ is a typed ranking, and it converts a
correlational ranking into a precise intervention target.

We use this capability as a causal probe in the MIB
sense~\citep{mueller2024mib}: causal mediation requires both
\emph{sufficiency} (intervening on the components reproduces the
targeted behavior) and \emph{necessity} (ablating them disrupts the
behavior under natural inputs). We test both directions on the CNN
under matched-budget comparisons against random selection
(Sec.~\ref{sec:rq2}).

\section{Applications}
\label{sec:applications}

We instantiate the protocol across three architectures: a 
$\beta$-VAE on CelebA ($512$ latent dimensions, $40$ binary 
attributes), a CNN on CIFAR-10 ($224$ filters, $10$ classes), and 
GPT-2 small ($144$ attention heads). This triple tests whether 
$(E,S,R,D,G,T)$ survives across generative vs.\ discriminative vs.\ 
language models. Measurement: Pearson correlation (VAE), 
entropy-based selectivity (CNN), attention-pattern primitives 
(GPT-2: induction/copying scores, QK/OV norms, position ratios), 
with pairwise activation correlations for $R$ throughout. 
Manipulation: $z_d \leftarrow z_d + \mathrm{sign}(s_{d,c}) \cdot 
2\sigma_d$ for VAE, $a'_f = \alpha \cdot a_f$ for CNN, 
zero/mean/path-patching ablation for GPT-2 (no amplitude semantics). 
Concept vocabularies: CelebA attributes, CIFAR-10 classes, MI-derived 
functional features (Sec.~\ref{sec:transformer-pilot}). Full 
instantiation matrix in App.~\ref{app:experiments}, 
Table~\ref{tab:mu_stats}.

\section{Evaluation}
\label{sec:evaluation}

We organise the evaluation around the two falsifiable hypotheses of
Sec.~\ref{sec:introduction}, with a brief response-faithfulness
check.

\paragraph{Setup.}
\label{sec:setup}
Architectures and instantiation details are in
Sec.~\ref{sec:applications}. Qwen3-8B is the conversational
interface; hybrid retrieval combines exact match on $E$ with
semantic search over $S/R/G/T$ templates. Metrics span retrieval
(Precision/Recall/NDCG), generation (BLEU, ROUGE), semantic
similarity (BERTScore), and faithfulness (Factual Accuracy = cited
statistics matching stored MU fields; Attribution = claims supported
by retrieved context); definitions in App.~\ref{app:metrics}.
VAE/CNN automatic metrics use $50$ single-turn queries per
architecture (App.~\ref{app:eval_query_sets}); GPT-2 schema-level
evaluation uses $8$ query paraphrases per IOI-style behaviour
(App.~\ref{app:gpt2}). Gold sets are pre-extracted offline and held
fixed across all methods.

\subsection{H1 --- structure beats content}
\label{sec:rq1}
H1 predicts that natural-language retrieval over component-level
analyses requires typed structure, not just the underlying content.
We test this in two complementary ways: against standard
sparse/dense baselines on VAE and CNN with the statistics held fixed
(Table~\ref{tab:ablation}), and against random partitions of the
same per-component text on the GPT-2 schema
(Fig.~\ref{fig:ablation_validation}D).

\paragraph{Baseline comparison (VAE/CNN).}
The structured system reaches Precision@5 of $89.6\%$/$96.8\%$ 
(VAE/CNN); a BM25-hybrid baseline with full access to the same 
per-component statistics reaches only $18.0\%$/$20.4\%$, and prose 
serialisation collapses to $3.7\%$/$2.8\%$ 
(Table~\ref{tab:ablation},~\ref{tab:gpt2_bm25}). Neither exact-match nor semantic-search 
alone suffices; the typed $E$ vs $S/R$ separation is what the 
hybrid retriever exploits.

\setlength{\abovecaptionskip}{5pt}
\setlength{\belowcaptionskip}{5pt}
\begin{table}[t]
\centering
\caption{Precision and Factual Accuracy at $k{=}5$, macro-averaged
over a $20$/$15$/$10$/$5$ entity/concept/relational/manipulation
mix. The structured system reaches 89.6\%/96.8\%; BM25 hybrid over
the same per-component statistics reaches 18.0\%/20.4\%, identifying
the central H1 gap. Prose serialisation collapses to 3.7\%/2.8\%,
controlling for serialisation choice. Full table
(14 configurations): App.~\ref{app:ablation}.}
\label{tab:ablation}
\small
\setlength{\tabcolsep}{3pt}
\resizebox{\columnwidth}{!}{%
\begin{tabular}{lcccc}
\toprule
& \multicolumn{2}{c}{VAE} & \multicolumn{2}{c}{CNN} \\
\cmidrule(lr){2-3} \cmidrule(lr){4-5}
Configuration & Prec & F-Acc & Prec & F-Acc \\
\midrule
\textbf{Full system (structured + hybrid)} & \textbf{89.6} & \textbf{97.9} & \textbf{96.8} & \textbf{96.8} \\
BM25 hybrid (same content)             & 18.0 & 45.5 & 20.4 & 48.5 \\
Unstructured text (prose)              & 3.7  & 0.7  & 2.8  & 41.9 \\
Exact match only                       & 62.4 & 78.3 & 71.2 & 82.1 \\
\bottomrule
\end{tabular}%
}
\end{table}

\paragraph{Structure-versus-content isolation (GPT-2).}
A stronger test isolates structure from content. On the GPT-2
schema, the $S{+}R$ subset reaches oracle recall@30 of $0.411$,
while random partitions of the \emph{same per-component text} into
two groups---identical content, structured arbitrarily---reach mean
oracle recall of $0.20$--$0.22$ ($p<0.01$ across three baseline
configurations; Fig.~\ref{fig:ablation_validation}D). Embedding
model and content are held fixed; only the field-level structure
varies. Together with the BM25 baselines (VAE/CNN Precision@5:
18.0\%/20.4\%; GPT-2 oracle recall@30: 0.143, below random
at 0.304; Table~\ref{tab:gpt2_bm25}, App.~N),

\subsection{H2 --- causal mediation on the CNN}
\label{sec:rq2}
H2 concerns the stricter MIB notion: not whether the response
reflects retrieved evidence, but whether retrieved components
\emph{causally mediate} the behaviour, established by both
\emph{sufficiency} (intervening on the components reproduces the
targeted behaviour) and \emph{necessity} (ablating them disrupts
behaviour under natural inputs). We test both directions on the
CNN under matched-budget comparisons against random selection.

\begin{table}[t]
\centering
\caption{CNN sufficiency and necessity vs.\ matched-budget random
selection. Sufficiency: $k{=}8$ filters, $\alpha \in \{30,50,70\}$,
$n{=}270$ trials, success $=$ target flip with
$\Delta p_{\text{target}} > 0.1$. Necessity: $k{=}8$, $1{,}000$
correctly classified images per class, ablation $=$ zero post-ReLU.
Random matched-budget uses $3$ seeds.}
\label{tab:rag_decompose}
\small
\setlength{\tabcolsep}{4pt}
\renewcommand{\arraystretch}{1.15}
\resizebox{\columnwidth}{!}{%
\begin{tabular}{@{}llrr@{}}
\toprule
Direction & Selection rule & Result & $\Delta$ random \\
\midrule
\multirow{4}{*}{Sufficiency}
  & A.\ MU + primary-class rule      & \textbf{76.3\%} & $+69.6$pp \\
  & C3.\ $S$-selectivity, no rule    & 58.9\%          & $+52.2$pp \\
  & C4.\ Raw-activation selectivity  & 58.9\%          & $+52.2$pp \\
  & R.\ Random                       & 6.7\%           & --- \\
\midrule
\multirow{2}{*}{Necessity}
  & B.\ MU + primary-class rule      & $-38.3$pp       & $+34.7$pp \\
  & R$_{\text{nec}}$.\ Random        & $-3.6$pp        & --- \\
\bottomrule
\end{tabular}%
}
\end{table}

\begin{figure}[t]
\centering
\IfFileExists{icml2026/fig/fig_necessity_v2.png}
  {\includegraphics[width=0.95\columnwidth]{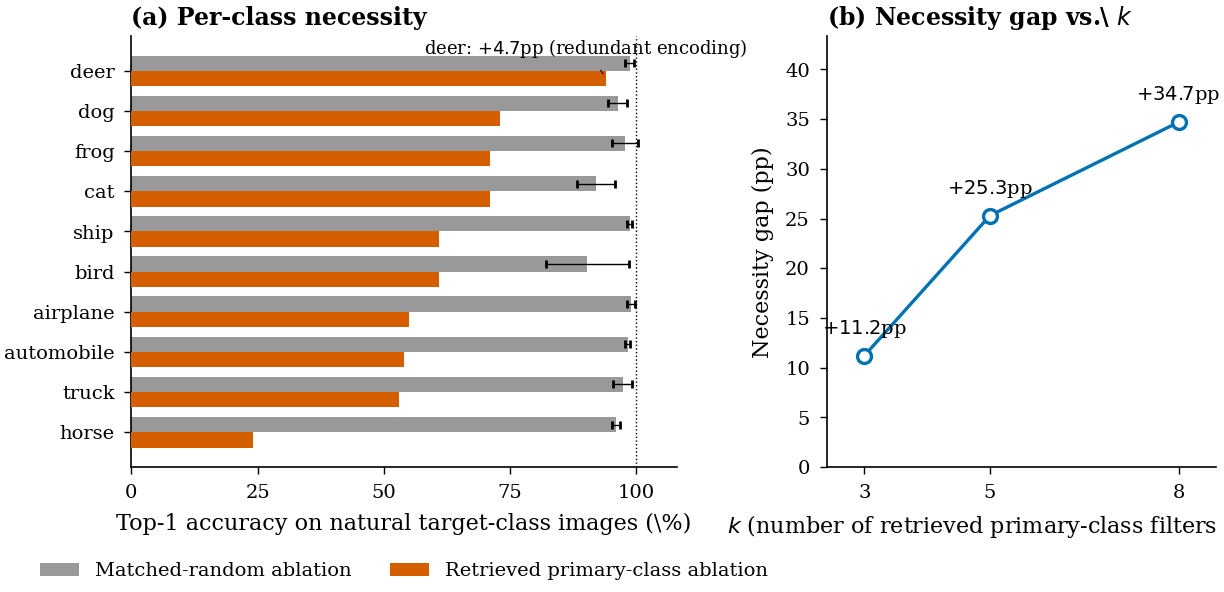}}
  {\fbox{\parbox{0.85\columnwidth}{\centering\vspace{1em}\textit{Figure placeholder: necessity-direction validation.}\vspace{1em}}}}
\caption{Necessity-direction validation. (a) Primary filters are 
necessary (larger drop); (b) necessity increases with $k$.}
\label{fig:necessity}
\end{figure}

\begin{figure}[t]
\centering
\includegraphics[width=0.95\columnwidth]{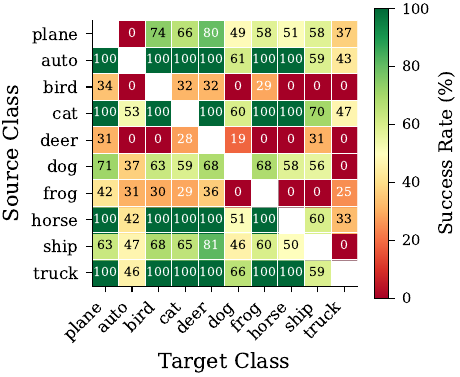}
\caption{Per-pair success rate, averaged across the full
amplification sweep $\alpha\in\{10,\dots,100\}$ (vs.\
Table~\ref{tab:rag_decompose}'s protocol $\{30,50,70\}$ for the H2
headline). Failures cluster within-category, tracing
filter-selectivity overlap. Full breakdown:
App.~\ref{app:manipulation}.}
\label{fig:main_heatmap}
\end{figure}

\paragraph{Sufficiency: targeted amplification flips predictions.}
For each of $90$ source-to-target class pairs, we retrieve
target-associated filters via RAG over the MUs, amplify their
post-ReLU activations by $\alpha$, and require
$\Delta p_{\text{target}} > 0.1$ with the prediction switching to
the target. At $k{=}8$ filters, $\alpha\!\in\!\{30,50,70\}$,
$n{=}270$ trials, RAG-retrieved filters flip the prediction in
\textbf{76.3\%} of trials, against \textbf{6.7\%} for matched-budget
random filters under identical conditions ($+69.6$pp;
Table~\ref{tab:rag_decompose}, row A vs R).

\paragraph{Decomposing sufficiency.}
The matched-budget gap decomposes (Table~\ref{tab:rag_decompose}): 
the typed primary-class rule contributes $+17.4$pp on top of 
selectivity-only ranking (A vs C3) operating over a 
faithfully-stored $S$ (C3$\,{=}\,$C4 at $58.9\%$); the $+52.2$pp 
gap from random to selectivity-only confirms amplification does 
not mechanically dominate this CNN.

\paragraph{Necessity: ablating retrieved filters disrupts natural classification.}
For each class $c$, we identify $100$ natural CIFAR-10 test images
that the unmodified model classifies correctly as $c$, then zero
the post-ReLU activations of the top-$k{=}8$ primary-class filters
retrieved by Eq.~\ref{eq:primary-class} and re-evaluate. Across
all $10$ classes, retrieved-filter ablation drops top-$1$ accuracy
from $100\%$ to $61.7\%$ (a $38.3$pp drop), while random-filter
ablation drops accuracy by only $3.6$pp
(Table~\ref{tab:rag_decompose}, B vs R$_\text{nec}$). The
$34.7$pp aggregate gap holds for $9/10$ classes
(Fig.~\ref{fig:necessity}a). The remaining class, deer, exhibits
a $4.7$pp gap; deer's stored selectivity is distributed across
a smaller pool of $24$ selective filters (vs.\ $29$--$57$ for
other classes), and the MU correctly surfaces this as a flatter
top-$m$ distribution in $S$. The necessity gap scales monotonically
with $k$ ($+11.2$pp at $k{=}3$, $+25.3$pp at $k{=}5$, $+34.7$pp at
$k{=}8$; Fig.~\ref{fig:necessity}b), indicating distributed
mediation by populations of filters rather than single units, with
the typed primary-class rule ordering the population by mediation
strength.

\paragraph{Joint reading.}
Sufficiency ($+69.6$pp) and necessity ($+34.7$pp) jointly establish 
MIB causal faithfulness for the CNN; the per-pair heatmap 
(Fig.~\ref{fig:main_heatmap}) shows failures cluster within-category, 
tracing filter-selectivity overlap. Per-$\alpha$ marginalisation
and the extended-search result ($88/90$ pairs after post-hoc
$\alpha$ tuning over resistant pairs; the primary H2 figure
remains $76.3\%$) in App.~\ref{app:manipulation_validation}.

\subsection{Decomposition mechanics on the GPT-2 schema}
\label{sec:rq3}

\begin{figure*}[t]
\centering
\includegraphics[width=0.95\linewidth]{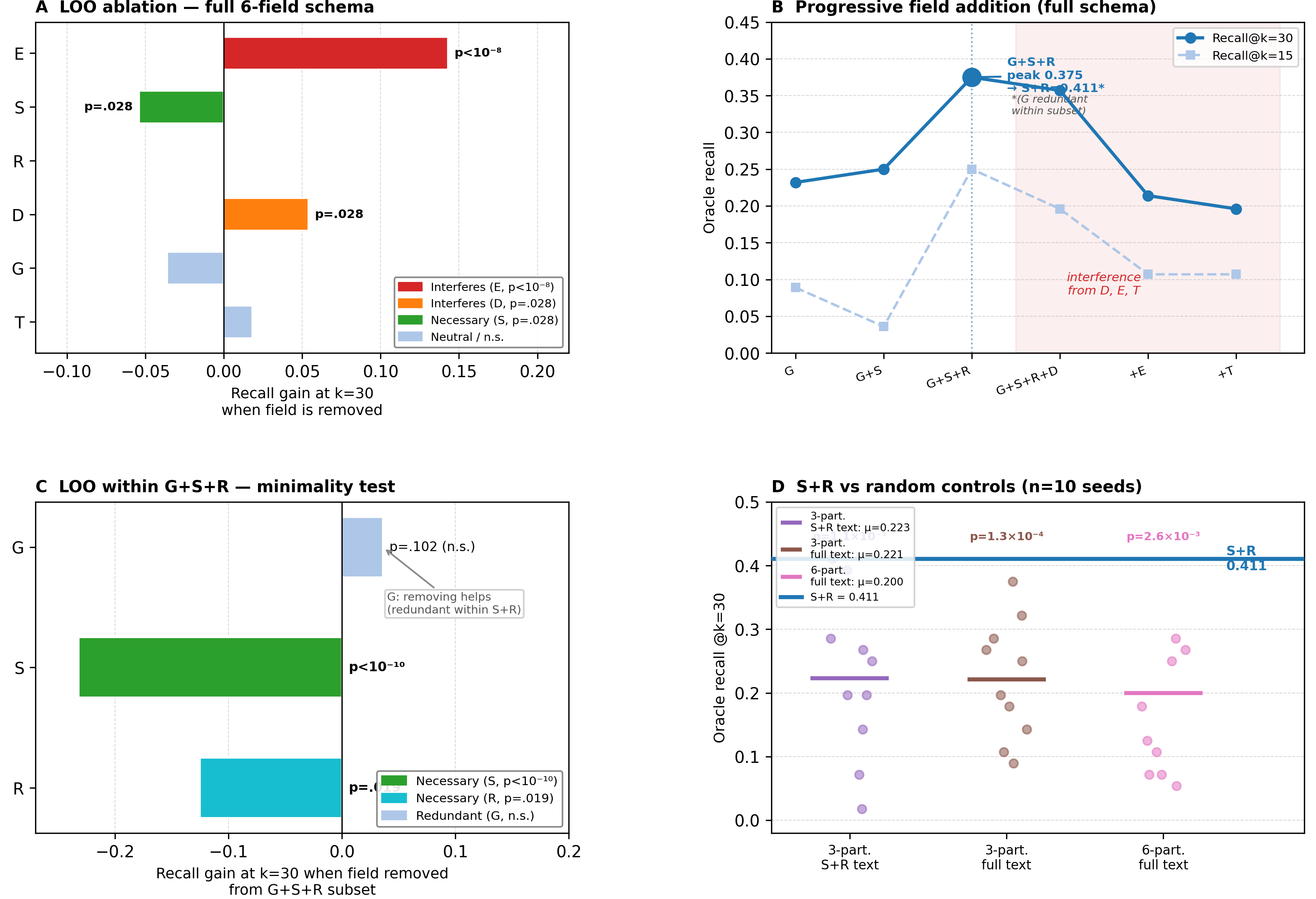}
\caption{Per-field ablation on the GPT-2 ESRDGT schema identifies 
$S{+}R$ as the minimal-optimal core. 
\textbf{(A)} Leave-one-out on the full six-field schema: $E$ 
($p<10^{-8}$) and $D$ ($p{=}.028$) actively interfere when present; 
$S$ ($p{=}.028$) is necessary; $R$, $G$, $T$ non-significant 
individually. 
\textbf{(B)} Progressive field addition: recall@30 reaches $0.375$
at $G{+}S{+}R$, drops in the interference zone when $D$, $E$, or
$T$ are added to that subset, and is highest at $S{+}R$ alone
($0.411$).
\textbf{(C)} Leave-one-out within the $G{+}S{+}R$ subset: $S$ 
($p<10^{-10}$) and $R$ ($p{=}.019$) are both necessary; $G$ is 
redundant within the subset. 
\textbf{(D)} $S{+}R$ versus random partitions of the same 
per-component text: $S{+}R$ recall@30 $=0.411$ beats three 
random-partition baselines ($\mu\in[0.20, 0.22]$) at $p<0.01$ 
across $10$ seeds, isolating the structural contribution from 
content.}
\label{fig:ablation_validation}
\end{figure*}

\begin{figure*}[t]
\centering
\includegraphics[width=0.95\linewidth]{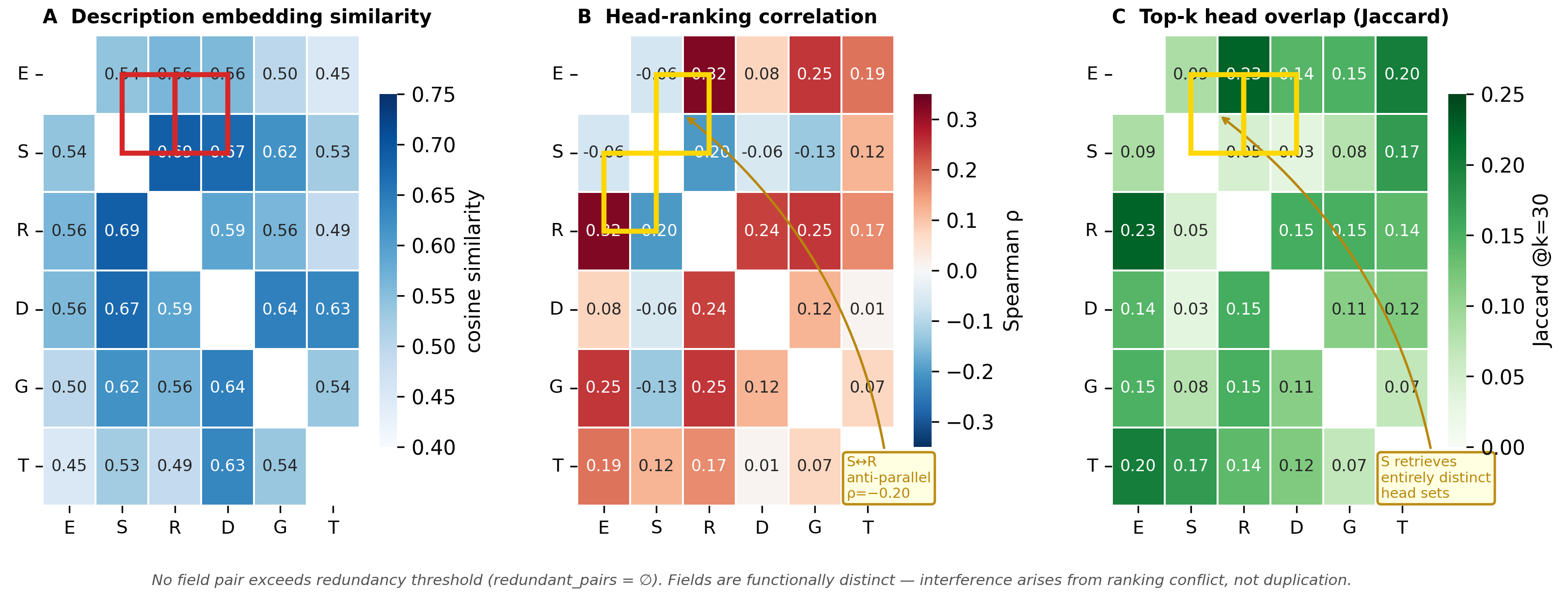}
\caption{Field redundancy analysis on the GPT-2 ESRDGT schema. 
\textbf{(A)} Pairwise embedding cosine similarity. 
\textbf{(B)} Head-ranking Spearman correlation: $S$ and $R$ show the strongest negative correlation ($\rho=-0.20$), indicating weakly anti-correlated head rankings.
\textbf{(C)} Top-$k$ head-set Jaccard overlap. No field pair exceeds redundancy thresholds, confirming functional distinctness. The $S\leftrightarrow R$ weak anti-correlation explains their complementarity, as they retrieve largely-disjoint head sets and jointly cover a larger functional space.}
\label{fig:redundancy_heatmaps}
\end{figure*}

With $T$ populated automatically from attention-pattern features
(induction/copying scores, QK and OV circuit norms, attention-position
ratios), we use the GPT-2 instantiation to ask whether the six-field
decomposition itself has identifiable internal structure. Three
findings emerge.

\paragraph{Minimal-optimal core.}
Leave-one-out ablation on the full six-field schema
(Fig.~\ref{fig:ablation_validation}A) identifies $S$ as necessary
($p{=}.028$); $E$ is removed without loss and---when re-added on top
of the $G{+}S{+}R$ subset---actively interferes with retrieval
($p<10^{-8}$), as does $D$ ($p{=}.028$). Within the $G{+}S{+}R$
subset (Fig.~\ref{fig:ablation_validation}C), $S$ ($p<10^{-10}$)
and $R$ ($p{=}.019$) are both necessary while $G$ is redundant
($p{=}.102$).
The progressive curve (Fig.~\ref{fig:ablation_validation}B) peaks
at $G{+}S{+}R$ (recall@30 $=0.375$) and degrades when $D$, $E$, or
$T$ are added to that subset, before rebounding at $S{+}R$ alone to
$0.411$---while the \emph{full} ESRDGT schema reaches only $0.196$,
confirming that interference from $E$, $D$, $T$ is substantial
(App.~\ref{app:ablation}). The \emph{minimal-optimal core} of the
decomposition is therefore $S{+}R$.

\paragraph{Non-redundancy and weak anti-correlation.}
Three measures of pairwise field overlap---description embedding
similarity, head-ranking correlation, and top-$k$ Jaccard---all
agree that no field pair exceeds the redundancy threshold
(Fig.~\ref{fig:redundancy_heatmaps}A,C). The two necessary fields,
$S$ and $R$, retrieve \emph{weakly anti-correlated} head rankings:
their head-ranking correlation is $\rho{=}{-}0.20$
(Fig.~\ref{fig:redundancy_heatmaps}B), the strongest negative
correlation in the matrix. This explains why their union is
maximally informative: $S$ and $R$ pull in different directions,
so combining them spans more of the relevant space than either
alone.
Per-field singleton retrieval and the field--function dissociation
between set-recall and first-rank MRR are in
App.~\ref{app:singleton}; field-level oracle discriminability
is in App.~\ref{app:het_analysis}.

\subsection{Schema transfer to attention heads}
\label{sec:rq4}
The protocol absorbs attention-head primitives without modification:
the $T$ field encodes induction scores, copying scores, QK/OV
circuit norms, and attention-position ratios extracted from each
head's attention matrices over a reference corpus
(App.~\ref{app:gpt2}). Across four IOI-style behaviours, retrieved
heads beat matched-budget random retrieval by $2.0$--$4.5\times$
on oracle recall metrics. At $k{=}30$, $S{+}R$ retrieval recovers
$5/7$ of the canonical IOI L7--9 name-mover heads of
\citet{wang2023ioi}---an oracle recall@30 of
$0.411$\footnote{Averaged over $8$ query paraphrases. This is set
membership with the canonical oracle, not causal restoration; see
the path-patching scope paragraph below.} and a $1.75\times$ lift
over random matched-budget retrieval. The required pivot from surface features (POS tags, syntactic
dependencies) to functional features (induction, copying, IOI
primitives) when populating $S$ defines a principled $S$-primitive
selection procedure: the schema is invariant across architectures
while $S$-content is chosen functionally, validated by reference-set
recovery ($3/16 \to 13/16$ heads; App.~\ref{app:gpt2}).

\paragraph{Path-patching scope.}
Framework-retrieved heads at $k{=}30$ collectively restore a
non-trivial fraction of the IOI logit-difference
(\texttt{frac\_restored}$\,{=}\,0.41$ vs random $0.23$), consistent
with the distributed nature of GPT-2's IOI circuit (the $7$ oracle
heads themselves restore only $0.12$). Per-field causal enrichment
under path-patching is not significant after Bonferroni correction
(App.~\ref{app:path_patching}); the framework's contribution on
GPT-2 is at the retrieval layer, not strict per-head causal mediation.
\paragraph{Response-faithfulness sanity check.}
A grounding-pipeline check ($n{=}15$ raters, $612$ ratings, scores 
$>4.0/5.0$; Qwen3-8B, Claude-Opus, and GPT-5.1 all reach $>$94\% 
factual accuracy on identical retrieved context) is in 
App.~\ref{app:human_eval}.

\section{Discussion and Limitations}
\label{sec:discussion}

\paragraph{Two kinds of faithfulness.}
We separate \emph{response faithfulness} (does the generated 
response reflect retrieved evidence?) from \emph{causal 
faithfulness}~\citep{mueller2024mib} (do identified components 
causally mediate the behaviour under both directions of 
intervention?). Sec.~\ref{sec:rq2} targets the latter and is the 
stronger claim; Sec.~\ref{sec:rq4}'s response-faithfulness check 
addresses the former. The four-way decomposition 
(Table~\ref{tab:rag_decompose}) localises the framework's H2 
contribution to the typed primary-class rule 
(Eq.~\ref{eq:primary-class}) operating over a faithfully-stored 
$S$; it does not trace causal paths, identify confounding 
interactions, or rule out multiple-mediator effects. Integrating 
MU-level evidence with circuit and SAE 
analyses~\citep{lieberum2024gemmascope, conmy2023acdc}---so 
retrieved units carry attribution-graph neighbourhoods or 
sparse-feature labels in $S$/$R$---tightens the claim without 
changing the interface.

\paragraph{Transformer instantiation (GPT-2).}
\label{sec:transformer-pilot}
The GPT-2 instantiation does three kinds of work simultaneously: 
structural-isolation test for H1 via random-partition controls 
(Sec.~\ref{sec:rq1}), per-field minimality and optimality with 
$S{+}R$ as the irreducible core (Sec.~\ref{sec:rq3}), and schema 
transfer to attention heads beating matched-budget random by 
$2.0$--$4.5\times$ across four behaviours (Sec.~\ref{sec:rq4}, 
App.~\ref{app:gpt2}). Each depends on automatic population of $T$ 
with attention-pattern primitives---no architecture-specific 
retrieval logic. Two qualifications: the concept vocabulary 
populating $S$ had to pivot from surface linguistic features 
($3/16$ reference heads recovered) to MI-derived functional features 
(attention-to-induction-target, IO-position, previous-token, self; 
$13/16$ recovered), validating a principled $S$-primitive selection
procedure for transformers; and on the strict path-patching causal-mediation
test, framework retrieval at $k{=}30$ collectively restores 
$0.409$ of the IOI logit-difference but does not show per-head 
causal enrichment (App.~\ref{app:path_patching}). The framework 
demonstrates set-overlap retrieval with the canonical IOI circuit, 
not strict per-head causal mediation; the latter would require 
retrieval criteria optimised directly for path-patching scores, 
which the current $T$ field does not provide. All five GPT-2 
experiments were pre-registered with SHA-256-hashed predictions 
before outcomes were computed.

\paragraph{Limitations.}
\label{sec:limits}
\textit{Scale.} No frontier-scale Transformer validation; SAE-based 
$S$~\citep{lieberum2024gemmascope} is the natural next step. 
\textit{Polysemanticity.} Top-concept $S$ is inadequate for 
polysemantic units; SAE primitives would address this. 
\textit{Concept vocabulary.} Resolution is conditional on 
$\mathcal{C}$ matched to the architecture; label-free domains need 
auto-labelling~\citep{oikarinen2023clipdissect} or SAE features as 
$S$ replacements. 
\textit{Per-head causal mediation on GPT-2.} Set-overlap with the 
canonical IOI circuit, not strict per-head path-patching enrichment 
(App.~\ref{app:path_patching}); the IOI circuit is distributed (the 
$7$-head oracle itself restores only $11.7\%$ of the logit-diff) 
and $T$ encodes attention-pattern features correlated with but not 
equal to path-patching scores. 
\textit{VAE causal.} Manipulation is qualitative; quantitative 
metrics for generative edits are future work. 
\textit{Queries.} Programmatic, not 
user-elicited~\citep{orgad2024actionable}.
\section{Conclusion}

We proposed Manifestation Units, a typed schema $(E, S, R, D, G, T)$
that organises component-level interpretability analyses into
structured, automatically populated tuples exposed through hybrid
retrieval --- creating a composable representation layer between
per-component statistics and downstream use.
Two hypotheses were confirmed under matched-budget controls.
\textbf{H1}: typed structure drives retrieval independently of
content --- structured retrieval reaches Precision@5 of
$89.6\%$/$96.8\%$ (VAE/CNN) versus $18.0\%$/$20.4\%$ for BM25 over
identical statistics, and $S{+}R$ beats random partitions of the same
GPT-2 text at $p{<}0.01$.
\textbf{H2}: on the CNN, retrieved components causally mediate the
targeted behaviour ($76.3\%$ sufficiency under amplification, $-38.3$pp
accuracy under ablation, both well above matched-random); on the GPT-2
schema the analogue is the minimal-optimal core ($S{+}R$), with per-head
causal enrichment under path-patching not significant after Bonferroni
correction (App.~\ref{app:path_patching}).
On the GPT-2 schema, ablation reveals that $S$ and $R$ form an
irreducible core, $G$ is redundant within it, and $E$/$D$ actively
interfere when added; the schema further recovers known IOI circuit
members and beats matched-budget random retrieval by
$2.0$--$4.5{\times}$ across four behaviours.
The immediate next steps are SAE-feature $S$ for polysemantic
components, frontier-scale validation, and label-free concept
vocabularies, none of which require changing the
$(E, S, R, D, G)$ interface or its extension~$(T)$.

\clearpage
\section*{Impact Statement}

Manifestation Units are interpretability \emph{infrastructure}, not
new analyses or new capabilities: a typed representation layer over
outputs that already exist in the field's toolkit. We expect three
groups to benefit. \emph{Interpretability researchers} gain a
composable substrate for organising per-component statistics across
studies; the schema is designed to accept SAE-feature
labels~\citep{lieberum2024gemmascope, cunningham2024sae},
attribution-graph neighbourhoods~\citep{conmy2023acdc}, and steering
or activation-patching primitives as future $S$/$R$/$G$ content
without changing the interface. \emph{Educators and students} gain a
natural-language interface to model internals that does not require
scripting around per-paper notebooks, lowering the entry cost for
engaging with mechanistic findings. \emph{Practitioners doing model
audit and debugging} gain retrieval-grounded evidence---every cited
statistic traces to a stored field value---which reduces the cost of
working with neural-network internals and supports reproducibility.
We see lowering these access costs as broadly aligned with safer
interpretability practice.

Three scope caveats are worth stating explicitly. \emph{(i) Domain.}
The $\beta$-VAE manipulation experiments edit facial attributes
(smiling, eyeglasses) on CelebA. Reconstruction quality is far below
current generative baselines, but the same retrieval-and-intervention
pattern, applied to higher-quality face models, would warrant
separate consideration as face-attribute manipulation tooling; we do
not develop or release such tooling. \emph{(ii) Inherited limits.}
The framework inherits the resolution of its underlying analyses;
extending to frontier models requires SAE-based or similarly scaled
$S$-primitives that this paper does not provide. \emph{(iii)
Settings.} All reported results are post-hoc analyses on small-scale
models---our own $\beta$-VAE on CelebA and CNN on CIFAR-10, together
with the publicly available GPT-2 small; we make no claim of readiness for safety-critical
deployment, and recommend domain-specific evaluation, access
controls, and auditing for any real-world use.

We adopt and recommend two methodological practices for
interpretability infrastructure work: pre-registration of experiments
via cryptographically hashed predictions before outcomes are computed
(all five GPT-2 experiments here), and explicit reporting of negative
results (the per-field path-patching null,
App.~\ref{app:path_patching}). Both reduce the risk that
interpretability tooling overclaims relative to its evidence---a
property of the field worth normalising.

\bibliography{ref}
\bibliographystyle{icml2026}

\newpage
\appendix
\onecolumn

\section*{Appendix Navigation}
\addcontentsline{toc}{section}{Appendix Navigation}

\begin{table}[h]
\caption{Navigation map: each main-paper section to its supporting appendix. Section and appendix references are clickable.}
\label{tab:navigation}
\small
\setlength{\tabcolsep}{4pt}
\renewcommand{\arraystretch}{1.15}
\begin{tabular}{p{0.32\linewidth} >{\centering\arraybackslash}p{0.04\linewidth} p{0.58\linewidth}}
\toprule
\textbf{Main paper} & \textbf{App.} & \textbf{Provides} \\
\midrule
All evaluations & \hyperref[app:eval-mapping]{A} & Evaluation-to-contribution mapping (C1--C3). \\
\hyperref[sec:mu-def]{\S3.1} MUs / extraction & \hyperref[app:extraction]{B} & Field-level math, thresholds ($m$, $r$, $\tau_{\text{syn}}$, $\tau_{\text{ant}}$), three-level analysis, sensitivity. \\
\hyperref[sec:hybrid-retrieval]{\S3.2} retrieval & \hyperref[app:retrieval]{C} & RAG architecture, templates, indexing, entity extraction, prompts. \\
\hyperref[sec:hybrid-retrieval]{\S3.2} SLM configuration & \hyperref[app:slm]{D} & SLM parameter studies (temperature, max tokens, prompt strategy). \\
\hyperref[sec:control-probe]{\S3.3} causal probe / \hyperref[sec:rq2]{\S5.2} H2 & \hyperref[app:manipulation]{E} & 90 CNN class-pair manipulations, RAG-vs-random, optimisation, filter overlap; VAE qualitative section. \\
\hyperref[sec:applications]{\S4} applications, \hyperref[sec:transformer-pilot]{\S6} GPT-2 & \hyperref[app:experiments]{F} & Architectures, dataset details, MU counts; per-architecture instantiation table; GPT-2 per-behaviour manipulation summary; CNN categorical structure; GPT-2 per-layer IOI mediation; concept-vocabulary pivot details; embedding stability across encoders; field-merging falsification. \\
\hyperref[sec:rq4]{\S5.4} path-patching / \hyperref[sec:limits]{\S6} limitations & \hyperref[app:path_patching]{G} & Path-patching causal-mediation analysis, budget-matched comparison, per-field causal-enrichment null result. \\
\hyperref[sec:rq3]{\S5.3} decomposition mechanics & \hyperref[app:het_analysis]{H} & Field-level oracle discriminability (Mann--Whitney, hand vs.\ learned modes). \\
\hyperref[sec:rq3]{\S5.3} decomposition mechanics & \hyperref[app:singleton]{I} & Per-field singleton retrieval; field--function dissociation (recall vs.\ MRR). \\
\hyperref[sec:setup]{\S5} setup & \hyperref[app:metrics]{J} & Formal metric definitions (retrieval, generation, semantic, faithfulness). \\
\hyperref[sec:rq4]{\S5.4} response-faithfulness sanity check & \hyperref[app:human_eval]{K} & Human study (612 ratings, expertise-stratified) with inter-rater reliability analysis. \\
\hyperref[sec:rq4]{\S5.4} response-faithfulness sanity check & \hyperref[app:llm_judge]{L} & LLM-as-judge full comparison tables (Claude-Opus, GPT-5.1). \\
\hyperref[sec:rq1]{\S5.1} H1 & \hyperref[app:auto_eval]{M} & Automatic-evaluation table across LLM backends and $k \in \{5, 10, 20\}$. \\
\hyperref[sec:rq1]{\S5.1} H1 & \hyperref[app:ablation]{N} & 14-configuration ablation (VAE) / 7-configuration ablation (CNN) with traditional retrieval baselines. \\
\bottomrule
\end{tabular}
\end{table}

\noindent\textbf{Reproducibility anchors.} The two anonymous interactive demos (\url{https://manifestation-xai.github.io/manifestation-transformers/} and \url{https://manifestation-xai.github.io/manifestation-cnn/}) walk through the extraction--retrieval--generation--manipulation pipeline with every intermediate output visible.

\vspace{0.5em}
\hrule
\vspace{0.5em}

\section{Evaluation-to-Contribution Mapping}
\label{app:eval-mapping}
\begin{table}[!ht]
\centering
\small
\setlength{\tabcolsep}{4pt}
\renewcommand{\arraystretch}{1.15}
\caption{Evaluation-to-contribution mapping.}
\label{tab:eval-mapping}
\begin{tabular}{p{0.18\textwidth} p{0.34\textwidth} p{0.34\textwidth} p{0.08\textwidth}}
\toprule
\textbf{Evaluation} & \textbf{What we did} & \textbf{What we wanted to see} & \textbf{Supports} \\
\midrule
Cross-architecture consistency &
Applied the same pipeline to a $\beta$-VAE (512 dims) and a CNN (224 filters). &
Whether the representation+retrieval interface generalizes across architectures. &
C1 \\
\midrule
\multicolumn{4}{l}{\textit{Hypothesis validation (main paper).}} \\
Structural isolation (H1, GPT-2) &
Compared $S{+}R$ retrieval against random partitions of identical per-component text on GPT-2 attention heads (Fig.~\ref{fig:ablation_validation}D; $p<0.01$ across 10 seeds and 3 baseline configurations). &
Whether typed structure beats arbitrary partitioning of the same content. &
C1, C3 \\
Causal mediation (H2, CNN) &
Targeted amplification ($k{=}8$, $\alpha \in \{30,50,70\}$, $n{=}270$) and natural-image ablation, vs.\ matched-budget random (Table~\ref{tab:rag_decompose}). &
Whether retrieved components causally mediate the targeted behaviour (MIB sense: sufficiency $+$ necessity). &
C1, C3 \\
Per-field ablation (H2, GPT-2) &
Leave-one-out and progressive field addition on the ESRDGT schema, plus field-merging falsification under coarser groupings (Fig.~\ref{fig:ablation_validation}A--C, Fig.~\ref{fig:figA4_merging}). &
Identify the minimal-optimal field combination; falsify the granularity choice. &
C1 \\
IOI circuit recovery (GPT-2) &
Set-overlap between framework-retrieved heads at $k \in \{7, 15, 30\}$ and the canonical IOI L7--9 name-mover heads of \citet{wang2023ioi}. &
Cross-architecture schema transfer to attention heads. &
C1 \\
Path-patching scope (GPT-2) &
Per-field causal enrichment under path-patching, Bonferroni-corrected across fields and retrieval strategies (App.~\ref{app:path_patching}). &
Honest disclosure of where retrieval enrichment does \emph{not} translate to per-head causal enrichment. &
C1 (scope) \\
\midrule
Automatic evaluation (single-turn) &
Evaluated retrieval, generation, semantic similarity, and faithfulness metrics across $k\in\{5,10,20\}$ on dims/filters queries. &
Correct retrieval of relevant units and grounded generation. &
C2 \\
LLM-as-judge evaluation &
Claude-Opus and GPT-5.1 rated multi-turn conversations (5 turns) on 7 metrics (Comprehensibility, Faithfulness, Correctness, Usefulness, Specificity, Trustworthiness, Coherence). &
Conversational quality under automated judging. &
C2 \\
Human evaluation &
15 evaluators across 3 expertise levels (Beginner, Intermediate, Expert), 612 assessments total (315 VAE + 297 CNN), multi-turn (5 turns), same 7 metrics. &
Usefulness and faithfulness across user expertise. &
C2 \\
SLM parameter studies &
Varied temperature (0.0--1.0), max tokens (256--2000), and prompting strategy (Basic, Detailed, Chain-of-Thought) while holding retrieval fixed. &
Robustness of grounded generation to decoding/prompt choices. &
C2 \\
Ablation: retrieval components &
Ablated query reformulation, reranking, boosting, over-retrieval, exact matching, and index composition (dims-only, Q\&A-only, interactions-only). &
Which retrieval components materially affect performance. &
C2 \\
\midrule
Ablation: structured vs.\ unstructured &
Replaced Manifestation Units with unstructured prose containing identical statistics. &
Whether structured $(E,S,R,D,G,T)$ representation is essential. &
C3 \\
Ablation: entity-aware retrieval &
Disabled exact matching (semantic-only) and compared against exact-only retrieval. &
Whether hybrid retrieval is necessary for mixed identifier/concept queries. &
C3 \\
SLM vs.\ frontier LLMs &
Compared Qwen3-8B against Claude-Opus and GPT-5.1 on identical automatic metrics with same retrieved context. &
Whether smaller models suffice when properly grounded. &
C3 \\
\bottomrule
\end{tabular}
\end{table}

This appendix maps each evaluation to the core contributions stated in the main paper (C1--C3). We also report manipulation experiments as additional validation evidence (Table~\ref{tab:additional-validation}), consistent with our framing in Sec.~3.4 that manipulation is used primarily as a validation mechanism.

\begin{table}[h]
\centering
\small
\setlength{\tabcolsep}{4pt}
\renewcommand{\arraystretch}{1.15}
\caption{Additional behavioral validation (not claimed as a separate contribution). The $97.8\%$ figure reflects post-hoc $\alpha$-tuning optimisation over resistant pairs; the primary H2 evaluation uses the fixed-protocol result of $76.3\%$ (Table~\ref{tab:rag_decompose}, main paper).}
\label{tab:additional-validation}
\begin{tabular}{p{0.18\textwidth} p{0.34\textwidth} p{0.30\textwidth} p{0.14\textwidth}}
\toprule
\textbf{Evaluation} & \textbf{What we did} & \textbf{What we wanted to see} & \textbf{Purpose} \\
\midrule
CNN manipulation (automatic) &
For 90 source$\to$target class pairs, conducted 540 tests (10 classes $\times$ 9 targets $\times$ 6 amplification levels: $\alpha \in \{10, 30, 50, 70, 90, 100\}$
). Retrieved target-associated filters via natural language queries and amplified activations. For 25 initially resistant pairs, applied systematic configuration search while holding the retrieved filter set fixed, resolving 23 additional pairs. &
Initial success: 72.2\% (65/90 pairs). Final success: \textbf{97.8\% (88/90 pairs)} after optimization. Strong gaps vs.\ random baselines, indicating retrieved components are causally relevant. Only 2 resistant pairs remain (airplane$\to$automobile, ship$\to$truck). &
Validation \\
\midrule
VAE manipulation (human) &
Human evaluation of natural-language attribute edits (e.g., ``make smile'', ``add eyeglasses'') via latent dimension modifications guided by retrieved units. Metrics: Visual Faithfulness (1--5), Degree of Change (1--5), Identity Preserved (Yes/No/Partial), Artifact Quality (None/Mild/Moderate/Severe). &
Perceptible, targeted edits with high identity preservation, noting reconstruction limitations inherent to VAE decoder quality. &
Validation \\
\bottomrule
\end{tabular}
\end{table}

\paragraph{Contributions summary.}
\begin{itemize}[leftmargin=*, itemsep=0.2em]
\item \textbf{C1:} Manifestation Units, a structured representation protocol $(E,S,R,D,G,T)$ that transforms statistical patterns from neural network analysis into retrievable, groundable knowledge units.
\item \textbf{C2:} A template-grounded hybrid retrieval architecture combining entity-aware exact matching with semantic search, enabling faithful natural language access to component-level knowledge.
\item \textbf{C3:} Empirical evidence that structured representation and hybrid retrieval each materially contribute in the studied settings: ablations show retrieval precision drops to near-zero when either component is removed; the response-faithfulness check is consistent across user expertise levels.
\end{itemize}
\section{Knowledge Extraction Details}
\label{app:extraction}

This appendix provides implementation details for the three-level analysis described in Sec.~3.3.

\subsection{Parameter Settings}

Table~\ref{tab:params} summarizes the key parameters used in our experiments.

\begin{table}[h]
\centering
\caption{Parameter settings for knowledge extraction.}
\label{tab:params}
\begin{tabular}{lcc}
\toprule
Parameter & VAE (CelebA) & CNN (CIFAR-10) \\
\midrule
Top-$m$ concepts per component & 5 & 3 \\
Top-$r$ relationships per component & 10 & 10 \\
Synergistic threshold $\tau_{\text{syn}}$ & 0.7 & 0.7 \\
Antagonistic threshold $\tau_{\text{ant}}$ & $-$0.5 & $-$0.5 \\
Clustering linkage & Ward's & Ward's \\
Clustering distance & $1 - |\rho_{ij}|$ & $1 - |\rho_{ij}|$ \\
Number of clusters & 20 & per-layer \\
\bottomrule
\end{tabular}
\end{table}

\subsection{Level 1: Individual Semantics}

\textbf{Activation extraction.} For each component, we compute a scalar activation per sample by spatially averaging its post-ReLU feature map (CNN) or spatially averaging the latent activation (VAE). All correlations and statistics are computed over $N$ test samples ($N=1{,}000$ for VAE, $N=10{,}000$ for CNN).

\textbf{VAE (CelebA).} For each latent dimension $d \in \{1, \ldots, 512\}$ and attribute $a \in \{1, \ldots, 40\}$, we compute Pearson correlation:
\begin{equation}
s_{d,a} = \text{corr}(\bar{z}_{\cdot,d}, \mathbf{a})
\end{equation}
where $\bar{z}_{\cdot,d} \in \mathbb{R}^{N}$ is the spatially-averaged activation of dimension $d$ across $N$ test samples, and $\mathbf{a} \in \{0,1\}^{N}$ is the binary attribute vector. We retain the top-$m$ attributes by $|s_{d,a}|$ for each dimension.

\textbf{CNN (CIFAR-10).} For each filter $f$ and class $c \in \{1, \ldots, 10\}$, we first compute class-wise mean activations $\mu_{f,c}$ over the validation set and normalize to obtain the distribution:
\begin{equation}
p_f(c) = \frac{\mu_{f,c}}{\sum_{c'} \mu_{f,c'}}.
\end{equation}
We then compute class selectivity as:
\begin{equation}
s_{f,c} = 1 - \frac{H(p_f)}{\log(10)}
\end{equation}
where $H(\cdot)$ is entropy. This entropy-based selectivity is 0 when activations are uniform across classes and approaches 1 when concentrated on a single class. We retain the top-$m$ classes by selectivity for each filter.

\subsection{Level 2: Pairwise Relationships}

\textbf{Formal definition of $N_r(k)$.} Let $\rho_{ki}$ denote the Pearson correlation between the activations of components $k$ and $i$ across the dataset:
\begin{equation}
\rho_{ki} = \text{corr}(\mathbf{a}_k, \mathbf{a}_i)
\end{equation}
where $\mathbf{a}_k, \mathbf{a}_i \in \mathbb{R}^{N}$ are activation vectors across $N$ samples.

The neighbor set $N_r(k)$ is defined as:
\begin{equation}
\mathcal{N}_r(k) = \{i_1, \ldots, i_r\} \subset I \setminus \{k\}
\end{equation}
where $|\rho_{k,i_1}| \geq |\rho_{k,i_2}| \geq \cdots \geq |\rho_{k,i_r}|$ and $|\rho_{k,i_r}| \geq |\rho_{k,j}|$ for all $j \notin \mathcal{N}_r(k)$. Ties are broken by component index (smaller index preferred).

\textbf{Relationship classification.} We label relationships as \emph{synergistic}, \emph{antagonistic}, or \emph{independent} using thresholds
$\tau_{\text{syn}}$ and $\tau_{\text{ant}}$:

\begin{equation}
t_{ki} = 
\begin{cases}
\text{synergistic} & \text{if } \rho_{ki} > \tau_{\text{syn}} \\
\text{antagonistic} & \text{if } \rho_{ki} < \tau_{\text{ant}} \\
\text{independent} & \text{otherwise}
\end{cases}
\end{equation}

\textbf{Computation and storage.} We compute correlations using minibatch accumulation and store only the top-$r$ neighbors per component. Full pairwise matrices are retained only for analysis layers with $\leq 512$ components. For VAEs, this yields a $512 \times 512$ correlation matrix. For CNNs, we compute both within-layer matrices (e.g., $32 \times 32$ for Conv1) and cross-layer matrices (e.g., $32 \times 64$ for Conv1$\to$Conv2).

\subsection{Level 3: Collective Organization}

\textbf{VAE: Semantic families.} We cluster using Ward's linkage on the distance $d_{ij} = 1 - |\rho_{ij}|$. The number of clusters (20) was determined by dendrogram inspection and silhouette analysis. Each dimension is assigned a cluster ID stored in $D$.

\textbf{VAE: Effective dimensionality.} We compute the effective dimensionality as:
\begin{equation}
\text{dim}_{\text{eff}}(\alpha) = \min\left\{k : \frac{\sum_{i=1}^{k} \lambda_i}{\sum_{i=1}^{n} \lambda_i} > \alpha \right\}
\end{equation}
where $\lambda_i$ are eigenvalues of the covariance matrix of latent activations across the dataset, sorted in descending order, and $\alpha = 0.95$. This measures how many dimensions capture 95\% of the variance.

\textbf{CNN: Cross-layer dependencies.} For consecutive layers $l$ and $l'$, we compute the cross-layer correlation matrix:
\begin{equation}
C^{(l \to l')} = [\rho_{ij}]_{i \in I_l, j \in I_{l'}}
\end{equation}
where $I_l$ and $I_{l'}$ are the filter sets for layers $l$ and $l'$. Strong correlations indicate feature dependencies across the hierarchy. These dependency scores are stored in $D$.

\begin{figure}[h]
\centering
\includegraphics[width=\textwidth]{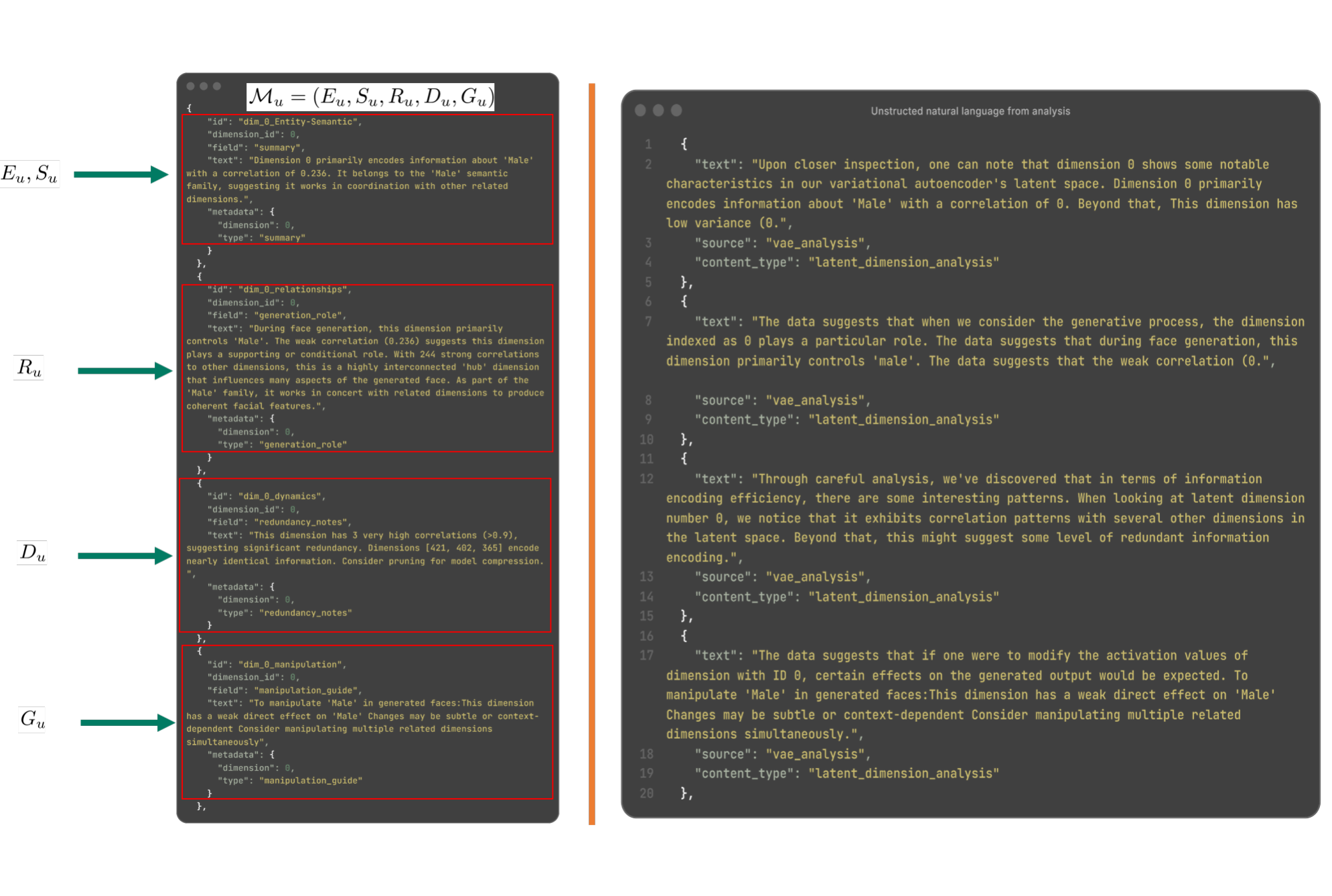}
\caption{Comparison of structured Manifestation Unit representation (left) versus unstructured natural language (right) for the same dimension. The structured format $(E, S, R, D, G)$ provides typed fields enabling exact-match retrieval on entity identifiers and faithful numeric grounding. The unstructured format buries the same information in prose, causing retrieval failures on entity-specific queries (e.g., ``dimension 0'' appears throughout the text but cannot be reliably matched).}
\label{fig:structured_vs_unstructured}
\end{figure}

\section{Retrieval Implementation Details}
\label{app:retrieval}

\begin{figure*}[t]
\centering
\includegraphics[width=\textwidth]{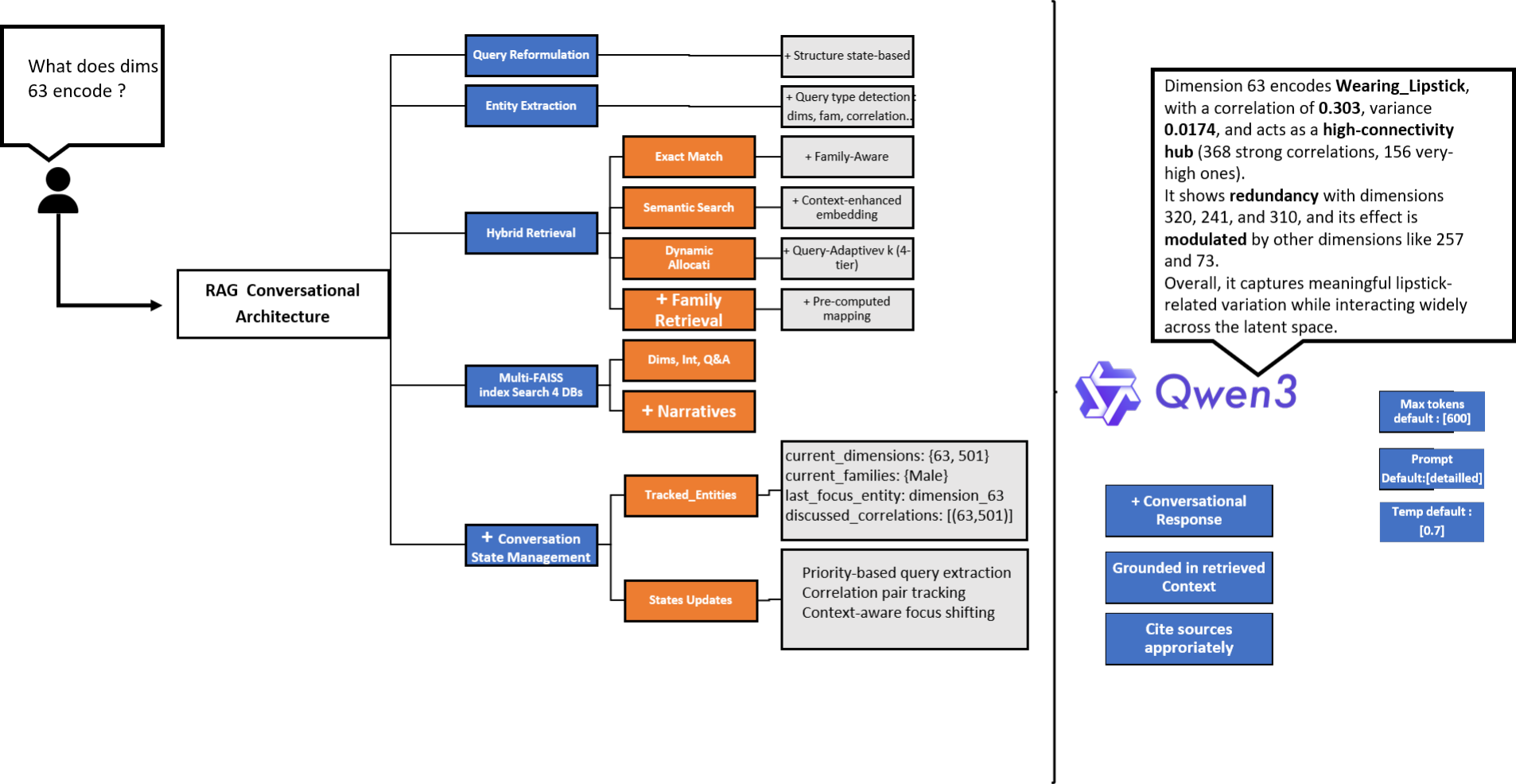}
\caption{RAG conversational architecture for the VAE system Conversational System. A user query (``What does dimension 63 encode?'') flows through: (1) query reformulation with state-based context tracking, (2) entity extraction detecting query type (dimension, family, correlation), (3) hybrid retrieval combining exact match, semantic search, and family-aware retrieval across four FAISS indices, (4) conversation state management tracking entities and focus, and (5) grounded response generation via Qwen3-8B. The output cites specific statistics from retrieved Manifestation Units (correlation $\rho=0.303$, variance 0.0174, relationship counts).}
\label{fig:rag_architecture}
\end{figure*}

Figure~\ref{fig:rag_architecture} illustrates the complete RAG conversational architecture for the VAE system.

\subsection{Template Structure}
\label{app:template}

Each Manifestation Unit $M_k=(E_k,S_k,R_k,D_k,G_k)$ is converted to natural language via deterministic templates.
These templates provide \emph{grounded views} of structured fields: each rendered statement is a direct instantiation
of values stored in the unit.

\subsubsection{Structured vs.\ Unstructured Representations}
\label{app:structured-vs-unstructured}

\begin{figure}[t]
\centering
\includegraphics[width=\textwidth]{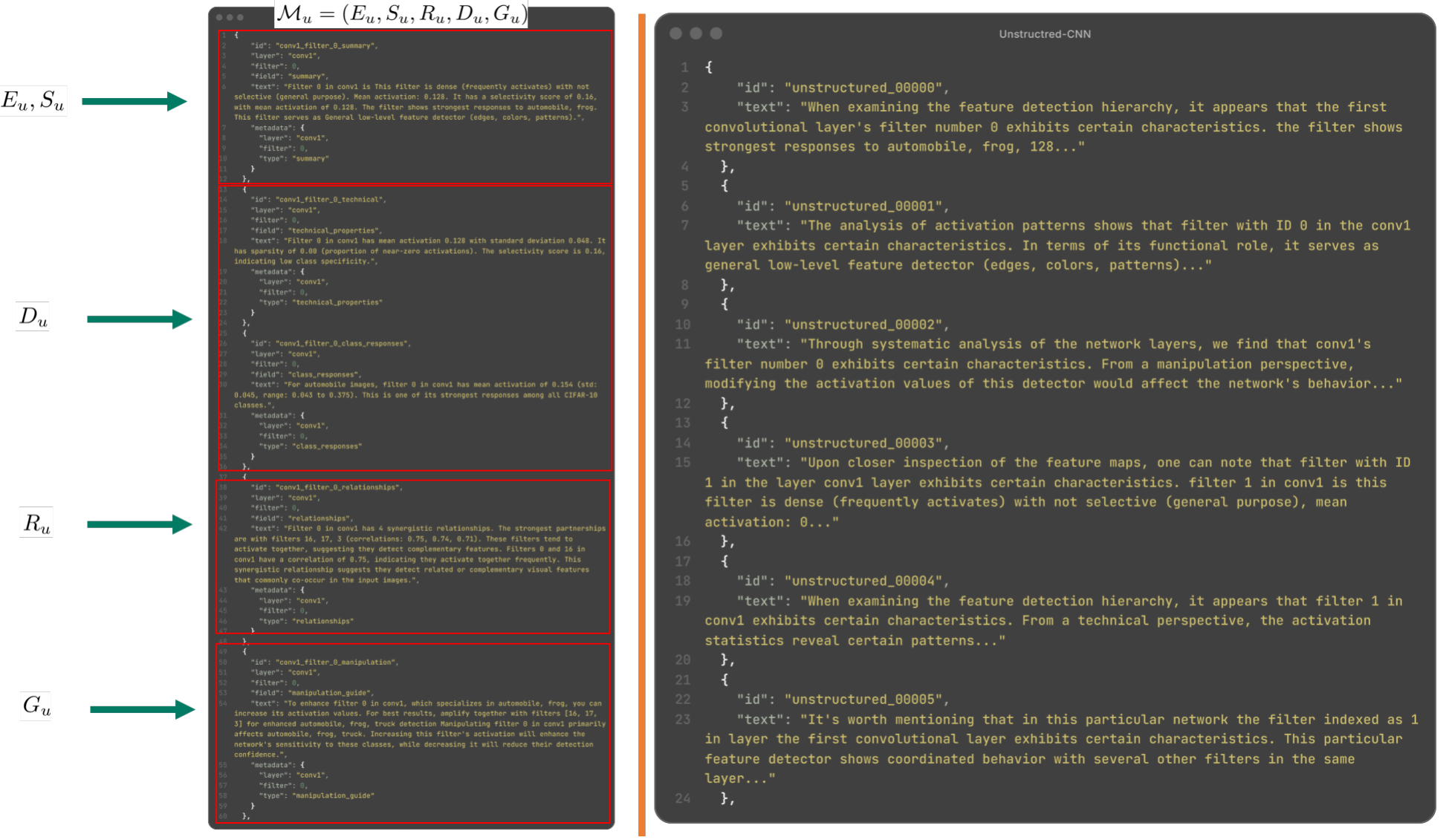}
\caption{Comparison of structured Manifestation Units (left) versus unstructured baseline (right) for CNN filter analysis. The structured representation decomposes knowledge into five specialized document types $(E_u, S_u, R_u, D_u, G_u)$, enabling field-specific retrieval. The unstructured baseline concatenates all information into narrative paragraphs, losing explicit structure.}
\label{fig:manifcnn}
\end{figure}

A key design decision in MANIFESTATION is the use of \emph{structured} templated documents derived from the five-tuple representation $(E,S,R,D,G)$, rather than unstructured narrative text. Figure~\ref{fig:manifcnn} illustrates this distinction for the CNN system.

\paragraph{Structured representation (MANIFESTATION).}
Each Manifestation Unit generates multiple specialized documents, with each document corresponding to a specific component of the five-tuple:
\begin{itemize}
  \item $E_u$ (Entity): Unique identifier enabling exact-match retrieval (e.g., ``Filter 0 in conv1'')
  \item $S_u$ (Summary): Primary semantic role and statistical properties
  \item $D_u$ (Technical Properties): Activation statistics, sparsity, class-specific responses
  \item $R_u$ (Relationships): Synergistic partnerships and correlation strengths
  \item $G_u$ (Manipulation Guide): Actionable instructions for enhancing or suppressing filter behavior
\end{itemize}

This decomposition enables \emph{field-specific retrieval}: queries about relationships retrieve relationship documents, while queries about manipulation retrieve guidance documents, improving precision and reducing irrelevant context.

\paragraph{Unstructured representation (Baseline).}
In contrast, unstructured baselines concatenate all information about an entity into free-form narrative text. As shown in the right panel of Figure~\ref{fig:manifcnn}, this approach loses the explicit field boundaries that enable targeted retrieval. Queries must rely entirely on semantic similarity, which struggles to distinguish between conceptually related but functionally distinct information (e.g., confusing ``what class does filter X detect?'' with ``what filters detect class Y?'').

\subsubsection{Templated Documents}
\label{app:templated-docs}

Rather than indexing a single concatenated text per unit, we generate \emph{multiple templated documents per unit},
each specializing in a distinct access pattern. This enables fine-grained retrieval where queries about encoding
retrieve summary documents while queries about redundancy retrieve redundancy-specific documents.

\paragraph{VAE template fields.}
For the VAE system (512 latent dimensions), each Manifestation Unit generates 4 templated documents:
\begin{itemize}
  \item \textbf{summary}: Primary attribute ($c^*$), correlation strength, semantic family membership, variance characteristics.
  \begin{quote}\small
  ``Dimension 63 primarily encodes `Wearing\_Lipstick' with correlation 0.303. Family: Wearing\_Lipstick. Variance: 0.0174. Strong correlations: 368.''
  \end{quote}
  
  \item \textbf{generation\_role}: Role during generation, number of strong correlations, hub/peripheral classification.
  \begin{quote}\small
  ``During face generation, this dimension primarily controls `Wearing\_Lipstick'. Hub dimension with 368 strong correlations across latent space.''
  \end{quote}
  
  \item \textbf{redundancy\_notes}: Very high correlations ($r>0.9$), redundant dimension IDs, pruning recommendations.
  \begin{quote}\small
  ``This dimension has 156 very high correlations ($>$0.9). Redundant with dimensions 320, 241, 310. Consider pruning for efficiency.''
  \end{quote}
  
  \item \textbf{manipulation\_guide}: How to manipulate the primary attribute, recommended co-manipulations.
  \begin{quote}\small
  ``To manipulate `Wearing\_Lipstick': adjust dimension 63. Co-manipulate correlated dimensions for stronger effect.''
  \end{quote}
\end{itemize}
This yields $512 \times 4 = 2{,}048$ dimension-level template entries in the primary Dimension Index.

\paragraph{CNN template fields.}
For the CNN system (224 filters: 32 in conv1, 64 in conv2, 128 in conv3), each unit generates up to 5 templated documents:
\begin{itemize}
  \item \textbf{summary}: Selectivity score, mean activation, top responding classes, layer role.
  \begin{quote}\small
  ``Filter 0 in conv1 is a general-purpose low-level feature detector. Selectivity: 0.16. Mean activation: 0.128.''
  \end{quote}
  
  \item \textbf{technical\_properties}: Sparsity coefficient, activation statistics (mean, std, range).
  \begin{quote}\small
  ``Sparsity: 0.00 (dense activation). Std: 0.048. Activation range: [0.0, 0.31].''
  \end{quote}
  
  \item \textbf{class\_responses}: Per-class activation statistics for CIFAR-10 classes.
  \begin{quote}\small
  ``Top responding classes: automobile ($\mu$=0.154, $\sigma$=0.045), frog ($\mu$=0.142), truck ($\mu$=0.138).''
  \end{quote}
  
  \item \textbf{relationships}: Synergistic filter partnerships and correlation strengths (when present).
  \begin{quote}\small
  ``4 synergistic partners. Strongest: filter 16 ($r$=0.75), filter 17 ($r$=0.74), filter 3 ($r$=0.71).''
  \end{quote}
  
  \item \textbf{manipulation\_guide}: How to enhance/suppress filter response, partner filters for amplification.
  \begin{quote}\small
  ``To enhance automobile/frog detection: amplify filter 0. Co-amplify filters 16, 17 for stronger effect.''
  \end{quote}
\end{itemize}
This yields 943 indexed template entries (not all filters have documented relationships).

\paragraph{Document metadata.}
Each templated document stores metadata for retrieval and traceability:
\begin{itemize}
  \item \textbf{Entity identifier}: \texttt{dimension\_id} (VAE) or \texttt{layer}+\texttt{filter} pair (CNN)
  \item \textbf{Field type}: Which template family this document instantiates
  \item \textbf{Text}: Natural language description derived from structured fields
  \item \textbf{Numeric values}: Original statistics for verbatim citation in responses
\end{itemize}

Table~\ref{tab:template-instantiation} shows variable instantiation by architecture.

\begin{table}[t]
\centering
\caption{Template variable instantiation by architecture.}
\label{tab:template-instantiation}
\begin{tabular}{lcc}
\toprule
Variable & VAE & CNN \\
\midrule
$E$ & ``Dimension \{d\}'' & ``Filter \{f\} in conv\{l\}'' \\
$c^*$ & Attribute name & Class name \\
$s^*$ & Correlation value & Selectivity score \\
Fields per unit & 4 & 5 (relationships optional) \\
Total documents & 2,048 (primary index) & 943 \\
\bottomrule
\end{tabular}
\end{table}

\subsection{RAG Architecture Details}
\label{app:rag-architecture}

The MANIFESTATION framework employs distinct RAG architectures for single-turn retrieval queries and multi-turn conversational interactions. This section details the architectural components illustrated in Figure~\ref{fig:manifvae}--\ref{fig:concnn}.

\subsubsection{Single-Turn Retrieval Architecture}
\label{app:single-turn-rag}

\begin{figure}[t]
\centering
\includegraphics[width=0.95\textwidth]{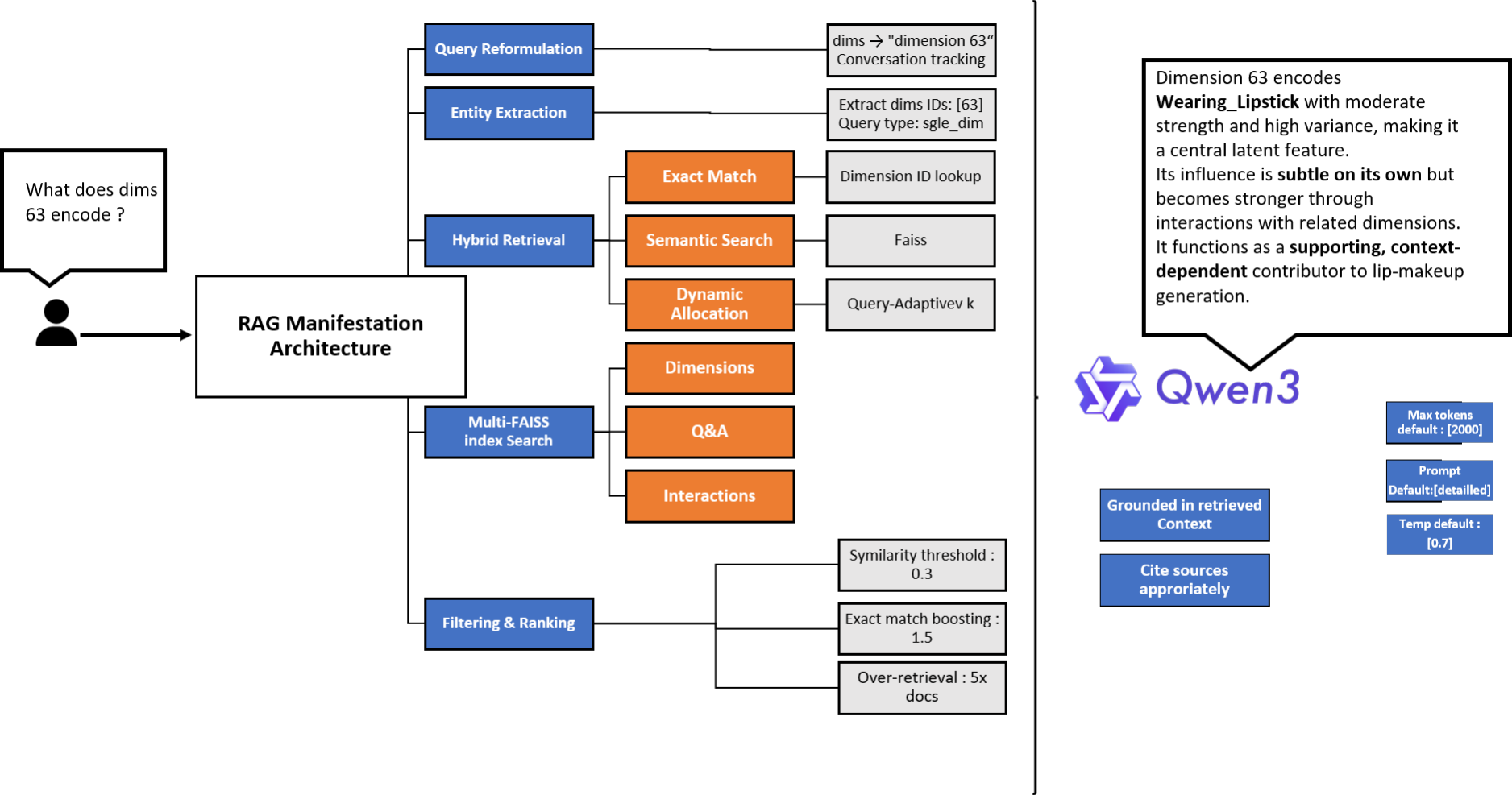}
\caption{VAE single-turn RAG architecture. The system processes queries through reformulation and entity extraction, then applies hybrid retrieval (exact match, semantic search, dynamic allocation) across multiple FAISS indices before filtering and ranking results for generation.}
\label{fig:manifvae}
\end{figure}

Figure~\ref{fig:manifvae} illustrates the RAG architecture for single-turn queries in the VAE system. Given a query such as ``What does dimension 63 encode?'', the system proceeds through the following stages:

\paragraph{Query processing.}
The \emph{Query Reformulation} module normalizes abbreviations and informal references (e.g., ``dims'' $\rightarrow$ ``dimension 63''). The \emph{Entity Extraction} module then identifies explicit dimension IDs using pattern matching, classifying the query type (single dimension, multi-dimension comparison, attribute exploration, etc.).

\paragraph{Hybrid retrieval.}
Three complementary retrieval strategies operate in parallel:
\begin{enumerate}
  \item \textbf{Exact Match}: Direct lookup of extracted entity IDs in the dimension index, guaranteeing retrieval of the target Manifestation Unit when explicitly referenced.
  \item \textbf{Semantic Search}: FAISS-based similarity search using sentence embeddings, enabling discovery of related dimensions even without explicit IDs.
  \item \textbf{Dynamic Allocation}: Query-adaptive $k$ selection based on query complexity---simple lookups retrieve fewer documents while comparative queries retrieve more.
\end{enumerate}

\paragraph{Multi-index search.}
The system maintains three specialized FAISS indices:
\begin{itemize}
  \item \textbf{Dimensions}: Primary index containing templated documents for all 512 VAE dimensions
  \item \textbf{Q\&A}: Pre-computed question-answer pairs for common query patterns
  \item \textbf{Interactions}: Pairwise correlation and relationship documents
\end{itemize}

\paragraph{Filtering and ranking.}
Retrieved documents undergo similarity thresholding (0.3), exact-match boosting (1.5$\times$), and over-retrieval with reranking (5$\times$ initial retrieval, then filter to final $k$).

\subsubsection{Conversational RAG Architecture}
\label{app:conversational-rag}

\begin{figure}[t]
\centering
\includegraphics[width=0.95\textwidth]{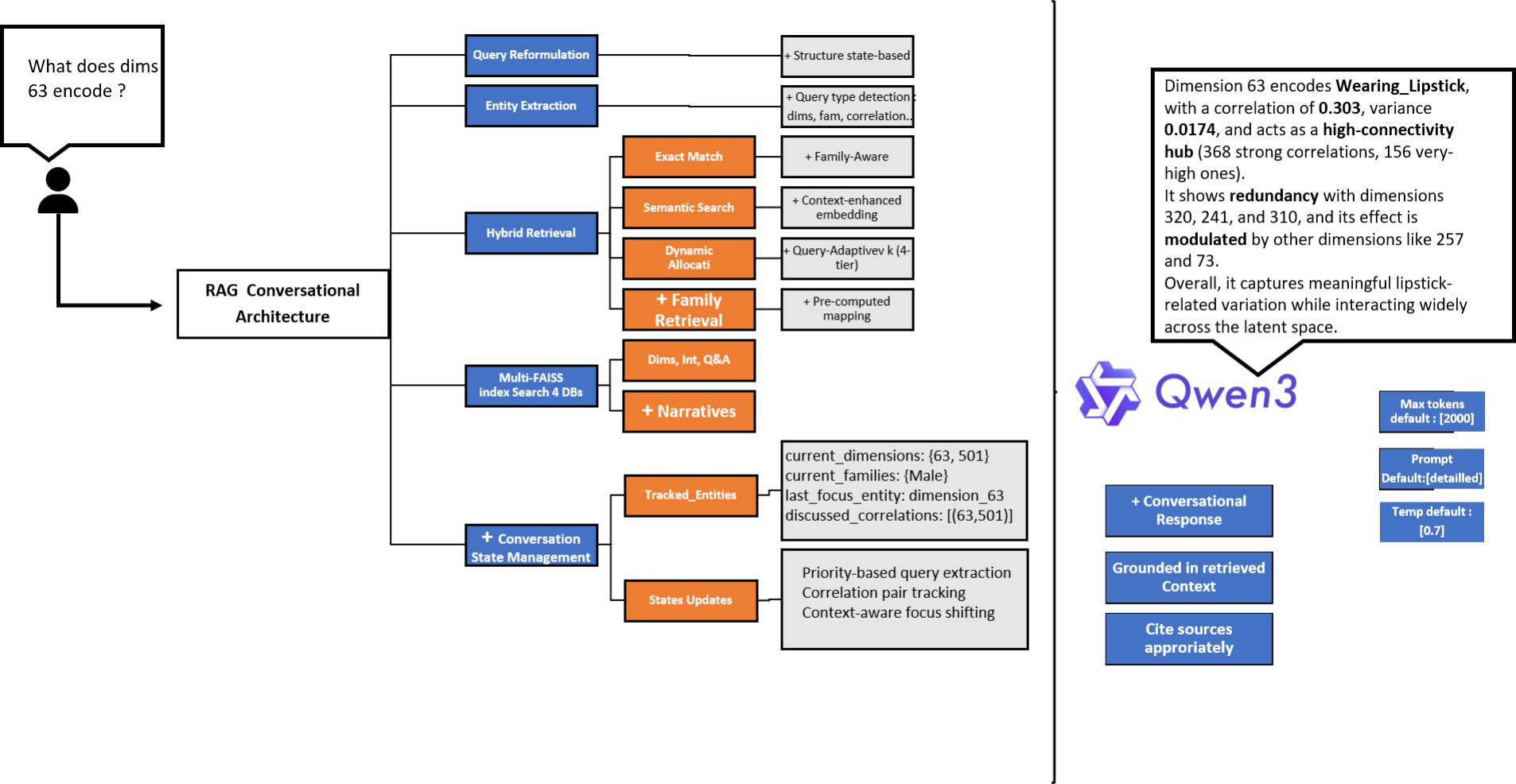}
\caption{VAE conversational RAG architecture. Extends the single-turn system with family retrieval, a narratives database, and conversation state management for multi-turn context tracking.}
\label{fig:vae-arch-exp}
\end{figure}

\begin{figure}[t]
\centering
\includegraphics[width=0.95\textwidth]{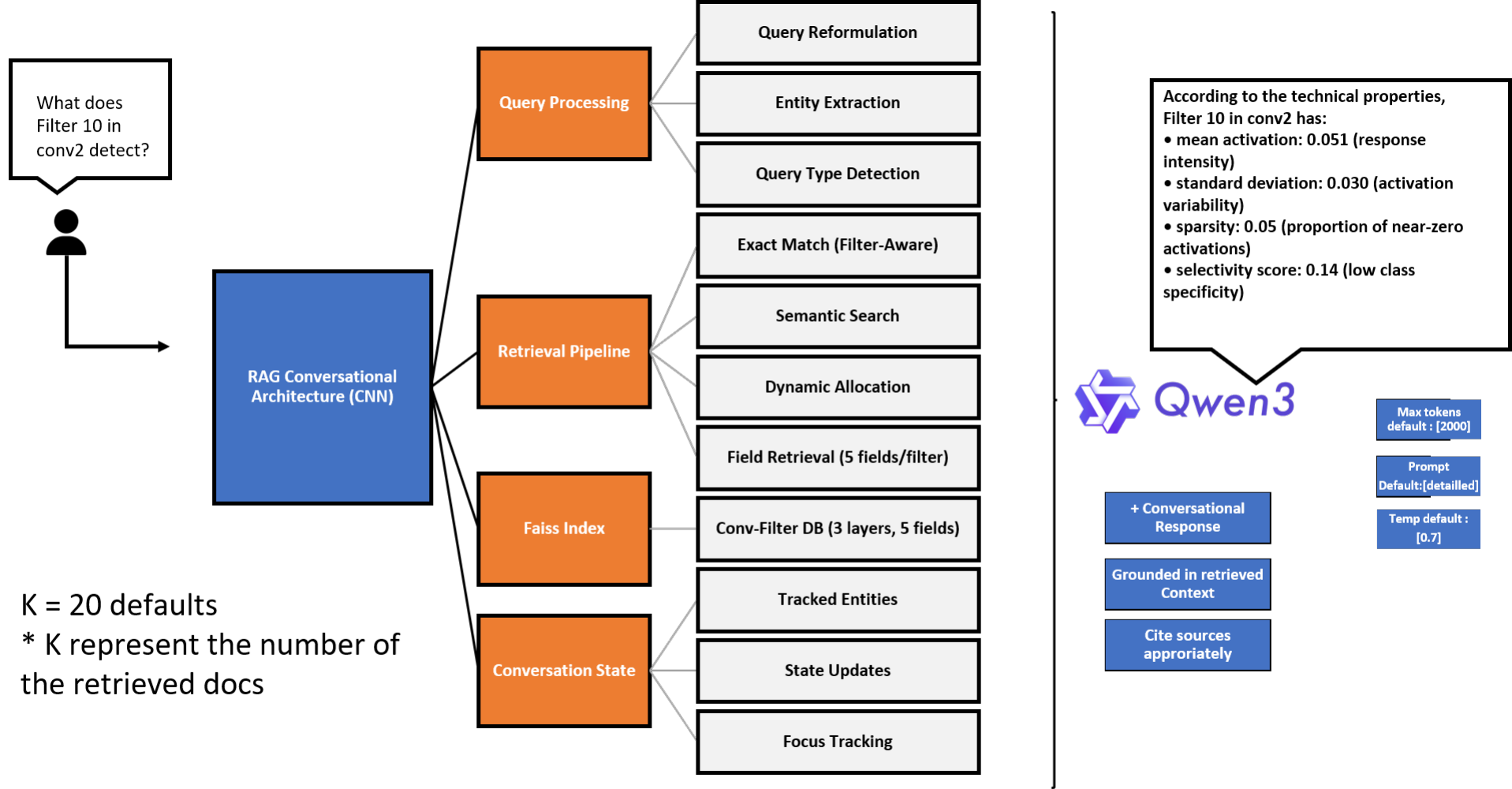}
\caption{CNN conversational RAG architecture. Processes filter queries through query reformulation and entity extraction, retrieves from a unified Conv-Filter database with 5 fields per filter, and maintains conversation state for multi-turn interactions.}
\label{fig:concnn}
\end{figure}

For multi-turn conversations, the architecture extends the single-turn system with state management and additional retrieval capabilities. Figure~\ref{fig:vae-arch-exp} and~\ref{fig:concnn} illustrate these conversational architectures for VAE and CNN systems respectively.

\paragraph{Extended VAE conversational architecture.}
As shown in Figure~\ref{fig:vae-arch-exp}, the conversational system adds three key components:

\begin{enumerate}
  \item \textbf{Family Retrieval}: Pre-computed mapping from semantic families (e.g., ``Male'', ``Wearing\_Lipstick'') to member dimensions, enabling efficient lookup for family-level queries.
  
  \item \textbf{Narratives Database}: A fourth FAISS index containing pre-generated narrative summaries that provide high-level overviews of dimension families and interaction patterns.
  
  \item \textbf{Conversation State Management}: Tracking of discussed entities, correlation pairs, and focus shifts to enable context-aware retrieval in follow-up turns.
\end{enumerate}

The state management component maintains:
\begin{itemize}
  \item \texttt{current\_dimensions}: Set of dimension IDs mentioned in conversation (e.g., \{63, 501\})
  \item \texttt{current\_families}: Semantic families under discussion (e.g., \{Male\})
  \item \texttt{last\_focus\_entity}: Most recently discussed entity for pronoun resolution
  \item \texttt{discussed\_correlations}: Pairs of dimensions whose relationships have been explored
\end{itemize}

\paragraph{CNN conversational architecture.}
Figure~\ref{fig:concnn} shows the analogous architecture for CNN filter analysis. The CNN system operates over a single database (Conv-Filter DB) organized by layer and filter, with 5 field types per filter. The conversation state tracks filter references across the three convolutional layers (conv1, conv2, conv3), maintaining focus for multi-turn exploration of filter hierarchies and cross-layer relationships.

\subsection{Query Structures and Evaluation Protocol}
\label{app:query-structures}

We evaluate the system using two complementary query types: single-turn retrieval queries and multi-turn conversational queries. Figure~\ref{fig:queries-cnn} and~\ref{fig:queriesvae} illustrate the JSON structures for both query types.

\subsubsection{Single-Turn Retrieval Queries}
\label{app:retrieval-queries}

Single-turn queries test the system's ability to retrieve and synthesize information from Manifestation Units in response to standalone questions. As shown in the left panels of Figure~\ref{fig:queries-cnn} and~\ref{fig:queriesvae}, each retrieval query specifies:

\begin{itemize}
  \item \textbf{id}: Unique query identifier (e.g., ``T1\_Q1'' for CNN tier-1, ``Q001'' for VAE)
  \item \textbf{tier}: Complexity level (1 = simple lookup, 2 = multi-entity, 3 = reasoning)
  \item \textbf{question}: Natural language query text
  \item \textbf{target\_layer/target\_filter} (CNN) or \textbf{dimensions} (VAE): Ground truth entities
  \item \textbf{target\_field}: Expected template field (e.g., ``technical\_properties'', ``relationships'')
  \item \textbf{answer\_type}: Expected format (float, integer, string, list)
  \item \textbf{expected\_format}: Granularity of expected answer (exact\_value, exact\_count, description)
\end{itemize}

Example CNN retrieval queries include:
\begin{quote}\small
``What is the mean activation value for Filter 0 in conv1?'' $\rightarrow$ technical\_properties, float \\
``How many synergistic relationships does Filter 1 in conv1 have?'' $\rightarrow$ relationships, integer
\end{quote}

Example VAE retrieval queries include:
\begin{quote}\small
``What does dimension 63 encode?'' $\rightarrow$ pairwise\_interaction, dimensions [373, 255] \\
``What does dimensions 299 and 335 encode?'' $\rightarrow$ pairwise\_interaction, multi-dimension
\end{quote}

\subsubsection{Multi-Turn Conversational Queries}
\label{app:conversational-queries}

\begin{figure}[t]
\centering
\includegraphics[width=\textwidth]{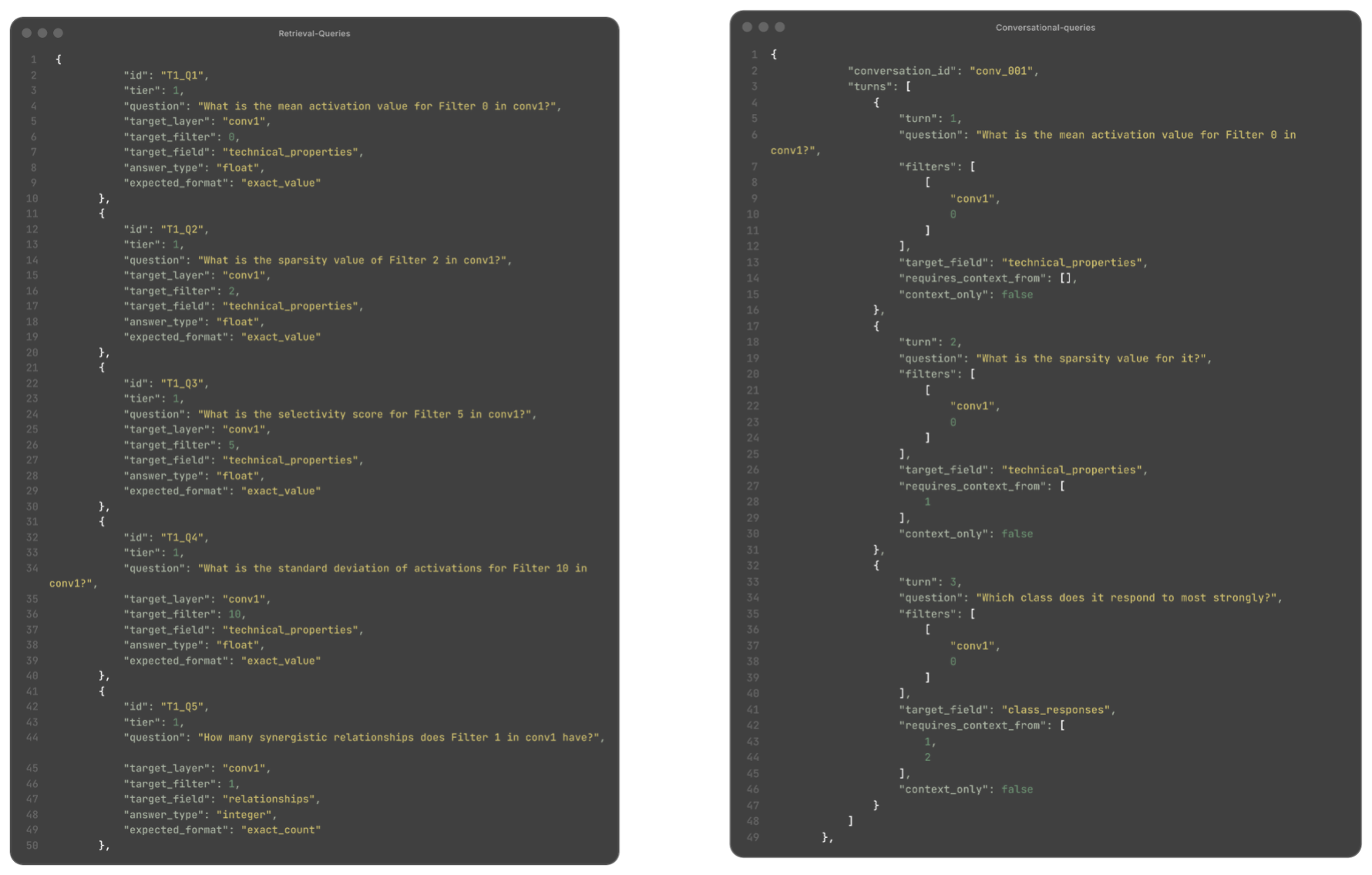}
\caption{CNN query structures. Left: Single-turn retrieval queries with explicit target specifications. Right: Multi-turn conversational queries with context dependencies indicated by \texttt{requires\_context\_from}.}
\label{fig:queries-cnn}
\end{figure}

\begin{figure}[t]
\centering
\includegraphics[width=\textwidth]{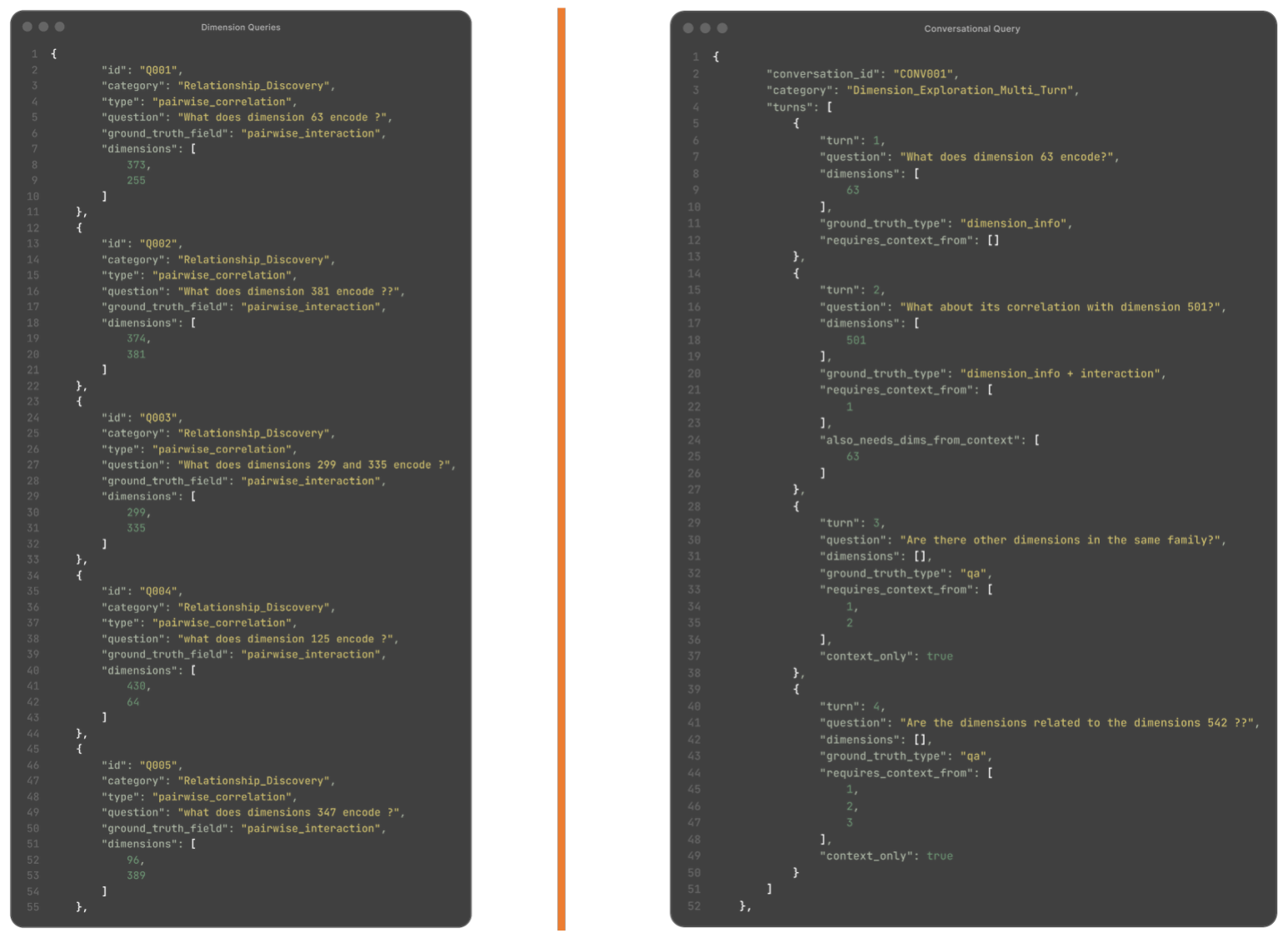}
\caption{VAE query structures. Left: Single-turn dimension queries categorized by relationship discovery type. Right: Multi-turn conversational queries demonstrating progressive exploration with context accumulation.}
\label{fig:queriesvae}
\end{figure}

Conversational queries test the system's ability to maintain context across multiple turns and resolve references to previously discussed entities. As shown in the right panels of Figure~\ref{fig:queries-cnn} and~\ref{fig:queriesvae}, each conversation specifies:

\begin{itemize}
  \item \textbf{conversation\_id}: Unique conversation identifier
  \item \textbf{category}: Conversation type (e.g., ``Dimension\_Exploration\_Multi\_Turn'')
  \item \textbf{turns}: Ordered list of turn objects, each containing:
  \begin{itemize}
    \item \textbf{turn}: Turn number (1-indexed)
    \item \textbf{question}: Natural language query for this turn
    \item \textbf{filters} (CNN) or \textbf{dimensions} (VAE): Entities relevant to this turn
    \item \textbf{target\_field}: Expected template field for retrieval
    \item \textbf{requires\_context\_from}: List of previous turn numbers needed to answer
    \item \textbf{context\_only}: Whether answer should come only from conversation context
  \end{itemize}
\end{itemize}

Example CNN conversational sequence:
\begin{quote}\small
Turn 1: ``What is the mean activation value for Filter 0 in conv1?'' \\
Turn 2: ``What is the sparsity value for it?'' (requires context from turn 1) \\
Turn 3: ``Which class does it respond to most strongly?'' (requires context from turns 1, 2)
\end{quote}

Example VAE conversational sequence:
\begin{quote}\small
Turn 1: ``What does dimension 63 encode?'' \\
Turn 2: ``What about its correlation with dimension 501?'' (requires context from turn 1) \\
Turn 3: ``Are there other dimensions in the same family?'' (context\_only: true) \\
Turn 4: ``Are the dimensions related to dimension 542?'' (requires context from turns 1, 2, 3)
\end{quote}

The \texttt{requires\_context\_from} field enables automatic evaluation of context-dependent retrieval by specifying which previous turns must be considered to correctly answer the current turn. The \texttt{context\_only} flag indicates queries that should be answerable purely from accumulated conversation context without additional retrieval.

\subsection{Entity Extraction Patterns}
\label{app:entity_patterns}

For hybrid retrieval, we use regular expressions to detect explicit component references in user queries.

\paragraph{VAE patterns.}
\begin{itemize}
  \item \texttt{dimension[s]?\textbackslash s*(\textbackslash d+)}   matches ``dimension 63'' or ``dimensions 63''
  \item \texttt{dim[s]?\textbackslash s*(\textbackslash d+)}   matches ``dim 63''
  \item \texttt{\textbackslash b(\textbackslash d\{1,3\})\textbackslash b}   standalone numbers in dimension context
\end{itemize}

\paragraph{CNN patterns.}
\begin{itemize}
  \item \texttt{filter\textbackslash s*(\textbackslash d+)\textbackslash s*in\textbackslash s*(conv[123])}   matches ``filter 5 in conv1''
  \item \texttt{(conv[123])\textbackslash s*filter\textbackslash s*(\textbackslash d+)}   matches ``conv1 filter 5''
  \item \texttt{filters?\textbackslash s*(\textbackslash d+(?:,\textbackslash s*\textbackslash d+)*)\textbackslash s*in\textbackslash s*(conv[123])}   matches filter lists
\end{itemize}

When a pattern matches, the system performs exact lookup by iterating through documents and matching
on the entity identifier field (\texttt{dimension\_id} for VAE, \texttt{layer}+\texttt{filter} for CNN).
Exact matches receive a similarity boost ($\times 1.5$) during retrieval ranking, ensuring that
explicitly referenced components appear at the top of retrieved results.
When no pattern matches, the system falls back to pure semantic search.

\subsection{Embedding, Indexing, and Retrieval}
\label{app:indexing}

\paragraph{Embedding model.}
We use Sentence-BERT (\texttt{all-MiniLM-L6-v2}, 384 dimensions) to embed templated documents.
We chose this lightweight sentence encoder for efficiency; preliminary experiments with larger encoders
(e.g., \texttt{all-mpnet-base-v2}) yielded similar retrieval performance.

\paragraph{Index structure.}
We use FAISS \texttt{IndexFlatIP} for similarity search. Document and query embeddings are
L2-normalized, so inner product corresponds to cosine similarity.

The VAE and CNN systems use different index architectures reflecting their scale and query patterns:

\subparagraph{VAE: Four specialized indices.}
The VAE system maintains four FAISS indices to support heterogeneous query types:
\begin{enumerate}
  \item \textbf{Dimension Index}: 2{,}048 templated documents from Manifestation Units (512 dimensions $\times$ 4 fields). 
        This is the primary index containing the core unit representations.
  \item \textbf{Q\&A Index}: 144 curated question-answer pairs covering general VAE architecture concepts, 
        latent space interpretation, and common user questions.
  \item \textbf{Interaction Index}: 2{,}607 documents describing pairwise dimension relationships, 
        conditional dependencies, and correlation patterns between specific dimension pairs.
  \item \textbf{Narrative Index}: 658 high-level analysis documents including semantic family summaries, 
        variance distribution insights, and correlation cluster narratives.
\end{enumerate}
This yields a total of \textbf{5{,}457} indexed documents across all VAE indices.

\subparagraph{CNN: Single unified index.}
The CNN system uses a single FAISS index containing all \textbf{943} templated documents.
The smaller scale (224 filters vs.\ 512 dimensions) and simpler relationship structure
do not require index separation; all template fields are searchable within one index.

\paragraph{Hybrid retrieval strategy.}
Given a user query $q$, retrieval proceeds as follows:
\begin{enumerate}
  \item \textbf{Entity extraction}: Apply regex patterns to detect explicit component references 
        (e.g., ``dimension 63'', ``filter 5 in conv1'').
  \item \textbf{Query type classification}: Classify query as conceptual, correlation-focused, 
        family-related, or component-specific based on keywords and extracted entities.
  \item \textbf{Dynamic budget allocation} (VAE only): Allocate retrieval slots across indices based on query type.
        For specific dimension queries with $k=20$ total slots:
        \begin{itemize}
          \item Dimension Index: $k_{\text{dim}} = 12$ (60\%)
          \item Q\&A Index: $k_{\text{qa}} = 4$ (20\%)
          \item Interaction Index: $k_{\text{int}} = 2$ (10\%)
          \item Narrative Index: $k_{\text{narr}} = 2$ (10\%)
        \end{itemize}
  \item \textbf{Exact match retrieval}: For queries with explicit entity references, retrieve \emph{all}
        documents for the referenced components and apply similarity boost ($\times 1.5$).
  \item \textbf{Semantic search}: Fill remaining slots via cosine similarity search across relevant indices.
  \item \textbf{Re-ranking}: Sort final results by (exact\_match, boosted\_similarity) in descending order.
\end{enumerate}

\paragraph{Why multi-index architecture (VAE).}
Separating indices by content type enables targeted retrieval for different query intents:
\begin{itemize}
  \item \emph{``What does dimension 63 encode?''} $\rightarrow$ Prioritizes Dimension Index for unit-specific data.
  \item \emph{``How do VAE latent spaces work?''} $\rightarrow$ Prioritizes Q\&A Index for conceptual answers.
  \item \emph{``What dimensions interact with 63?''} $\rightarrow$ Prioritizes Interaction Index for relationship evidence.
  \item \emph{``What is the Wearing\_Lipstick family?''} $\rightarrow$ Prioritizes Narrative Index for family summaries.
\end{itemize}
The ablation study (Table~\ref{tab:ablation_full}) confirms that removing any single index degrades performance,
with the Dimension Index being most critical (removing it causes recall to drop to 0\% on dimension-specific queries).

\paragraph{Why multi-field indexing.}
Each Manifestation Unit generates multiple documents (one per field), enabling queries to retrieve
the most relevant \emph{view} of a unit. For example, ``What does dimension 63 encode?'' retrieves
the \texttt{summary} document, while ``Which dimensions are redundant with dimension 63?'' retrieves the
\texttt{redundancy\_notes} document. This yields higher precision than concatenating all fields into a
single document (see ablation: Unstructured Text baseline achieves only 20.0\% recall vs.\ 94.3\% for structured retrieval).

\subsection{Example Queries and Retrieval}
\label{app:query_examples}

\paragraph{Retrieval traces.}
To illustrate retrieval behavior, we present traces from 50 example queries per system, focusing on the \texttt{Relationship\_Discovery} category.
This illustrative set is separate from the main evaluation benchmark (Appx.~\ref{app:eval_query_sets}).
The 50 example queries are distributed across three query types that test different retrieval capabilities.

\paragraph{Query distribution.}
\begin{itemize}
  \item \textbf{Pairwise correlation} (25 queries): Single or paired component queries.
        Ground truth field: \texttt{pairwise\_interaction}.
  \item \textbf{Triple correlation} (20 queries): Three-component relationship queries.
        Ground truth field: \texttt{multi\_interaction}.
  \item \textbf{Family members} (5 queries): Semantic grouping queries.
        Ground truth field: \texttt{summary}.
\end{itemize}

Table~\ref{tab:query_examples} shows representative queries and their retrieval behavior.

\begin{table}[h]
\centering
\caption{Example queries and retrieval traces. Exact matches (EM) receive $1.5\times$ similarity boost.}
\label{tab:query_examples}
\resizebox{\textwidth}{!}{%
\begin{tabular}{p{4.5cm}p{1.8cm}p{3.5cm}p{4.5cm}}
\toprule
\textbf{Query} & \textbf{Type} & \textbf{Retrieved Fields} & \textbf{Retrieval Behavior} \\
\midrule
\multicolumn{4}{l}{\textit{VAE System (512 dimensions, 4 indices, 5{,}457 total documents)}} \\
\midrule
``What does dimension 63 encode?'' & pairwise & Dim 63: summary, generation\_role, redundancy\_notes, manipulation\_guide (all EM) & Exact match on dim\_id=63; all 4 fields retrieved with $1.5\times$ boost \\
\addlinespace
``Why are dimensions 49, 4, and 215 correlated?'' & triple & Dims 49, 4, 215: summary (EM) + Interaction docs for pairs & Exact match on 3 dims + semantic search in Interaction Index \\
\addlinespace
``Why are dimensions 154, 186, and 308 correlated?'' & family & Dims 154, 186, 308: summary (EM) + Narrative family insights & Family detected; Narrative Index allocation increased \\
\midrule
\multicolumn{4}{l}{\textit{CNN System (224 filters, 1 index, 943 documents)}} \\
\midrule
``What does filter 0 in conv1 encode?'' & pairwise & conv1-f0: summary, technical, class\_responses, relationships, manipulation (all EM) & Exact match on (conv1, 0); all 5 fields retrieved \\
\addlinespace
``Why are filters 0, 16, and 17 in conv1 correlated?'' & triple & conv1-f0, f16, f17: summary, relationships (EM) & Exact match on 3 filters + semantic expansion \\
\addlinespace
``Why are filters 0, 1, and 2 in conv1 correlated?'' & family & conv1-f0, f1, f2: summary (EM) & Summary fields contain family/grouping information \\
\bottomrule
\end{tabular}%
}
\end{table}

\paragraph{Retrieval workflow trace (VAE).}
For the query ``What does dimension 63 encode?'':
\begin{enumerate}
  \item \textbf{Entity extraction}: Regex detects ``dimension 63'' $\rightarrow$ \texttt{target\_dims = [63]}
  \item \textbf{Query classification}: Single dimension query $\rightarrow$ \texttt{type = specific}
  \item \textbf{Budget allocation}: $k=20$ allocated as: dim=12, qa=4, int=2, narr=2
  \item \textbf{Exact match}: All 4 documents for dimension 63 retrieved with $1.5\times$ boost
  \item \textbf{Semantic fill}: Remaining 8 dim slots filled with similar dimension summaries (dims 67, 351, 332, ...);
        Q\&A slots filled with related questions (``What does dimension 64 encode?'')
  \item \textbf{Final retrieval}: 20 documents total, 4 exact matches, avg similarity 60.15\%
  \item \textbf{Context assembly}: Documents grouped by dimension ID for coherent LLM input
\end{enumerate}

\paragraph{Retrieval workflow trace (CNN).}
For the query ``What does filter 0 in conv1 detect?'':
\begin{enumerate}
  \item \textbf{Entity extraction}: Regex detects ``filter 0 in conv1'' $\rightarrow$ \texttt{target = [(conv1, 0)]}
  \item \textbf{Field detection}: Query contains ``what does'' $\rightarrow$ primary field = \texttt{summary}
  \item \textbf{Exact match}: All 5 documents for conv1-filter0 retrieved with $1.5\times$ boost
  \item \textbf{Semantic fill}: Remaining slots filled with related filters (conv1-f1, conv1-f16, conv2-f0, ...)
  \item \textbf{Final retrieval}: 10 documents total, 1 exact match in required field, avg similarity 58.86\%
\end{enumerate}

\subsection{Generation Prompt}
\label{app:prompt}

The language model (Qwen3-8B) receives the query along with retrieved evidence grouped by entity.
The prompt enforces grounded answering: the model must answer based only on retrieved content
and must cite numeric values from retrieved fields.

\paragraph{System prompt structure.}
The system prompt specifies:
\begin{itemize}
  \item \textbf{Model context}: Architecture description (VAE with 512 dimensions / CNN with 224 filters across 3 layers)
  \item \textbf{Grounding rules}: Cite specific data from retrieved documents; distinguish facts from interpretation
  \item \textbf{Domain knowledge}: Architecture-specific concepts (semantic families, layer hierarchy, correlation semantics)
  \item \textbf{Response format}: Quote exact numeric values; explain relationships using retrieved evidence
\end{itemize}

\paragraph{VAE context format.}
Retrieved documents are provided grouped by dimension:
\begin{verbatim}
=== RETRIEVED EVIDENCE ===

  Dimension 63  
  • summary: Dimension 63 primarily encodes 'Wearing_Lipstick' 
    with correlation 0.303. Family: Wearing_Lipstick. 
    Variance: 0.0174. Strong correlations: 368.
  • generation_role: During face generation, this dimension 
    primarily controls 'Wearing_Lipstick'. Hub dimension with 
    extensive connections to other dimensions.
  • redundancy_notes: This dimension has 156 very high 
    correlations (>0.9). Redundant with dims 320, 241, 310.
  • manipulation_guide: To manipulate 'Wearing_Lipstick': 
    adjust dimension 63. Co-manipulate correlated dims.

  Q&A Knowledge  
  • Q: What does dimension 64 encode? [similarity: 72.2%]
  • Q: What does dimension 96 encode? [similarity: 70.5%]

  Narrative Insights  
  • Family Wearing_Lipstick: 35 dimensions encoding makeup...

=== END EVIDENCE ===
\end{verbatim}

\paragraph{CNN context format.}
For CNN queries, evidence is grouped by layer and filter:
\begin{verbatim}
=== RETRIEVED EVIDENCE ===

  CONV1 Filter 0  
  • summary: Filter 0 in conv1 is a general-purpose low-level 
    feature detector. Selectivity: 0.16. Mean activation: 0.128.
  • technical_properties: Sparsity: 0.0009 (dense activation). 
    Std: 0.048. Range: [0.0, 0.31].
  • class_responses: Top classes - automobile ($\mu$=0.154), 
    frog ($\mu$=0.142), truck ($\mu$=0.138).
  • relationships: 4 synergistic partners. Strongest: 
    filter 16 (r=0.75), filter 17 (r=0.74), filter 3 (r=0.71).
  • manipulation_guide: Enhance automobile/frog detection 
    by amplifying filter 0. Co-amplify filters 16, 17.

  CONV1 Filter 16   [semantic match, similarity: 65.2%]
  • summary: Filter 16 in conv1 specializes in edge detection...

=== END EVIDENCE ===
\end{verbatim}

The model is instructed to quote numeric values verbatim from these fields (e.g., ``correlation 0.303'', 
``selectivity 0.16''), ensuring traceability from response to extracted statistics.

\section{SLM Configuration and Ablation Study}
\label{app:slm}

This appendix documents the Small Language Model (SLM) configuration used in the conversational interface
and presents comprehensive ablation studies evaluating the impact of generation parameters on response quality.

\subsection{Conversational Architecture}
\label{app:slm_architecture}

\begin{figure*}[t]
\centering
\includegraphics[width=\textwidth]{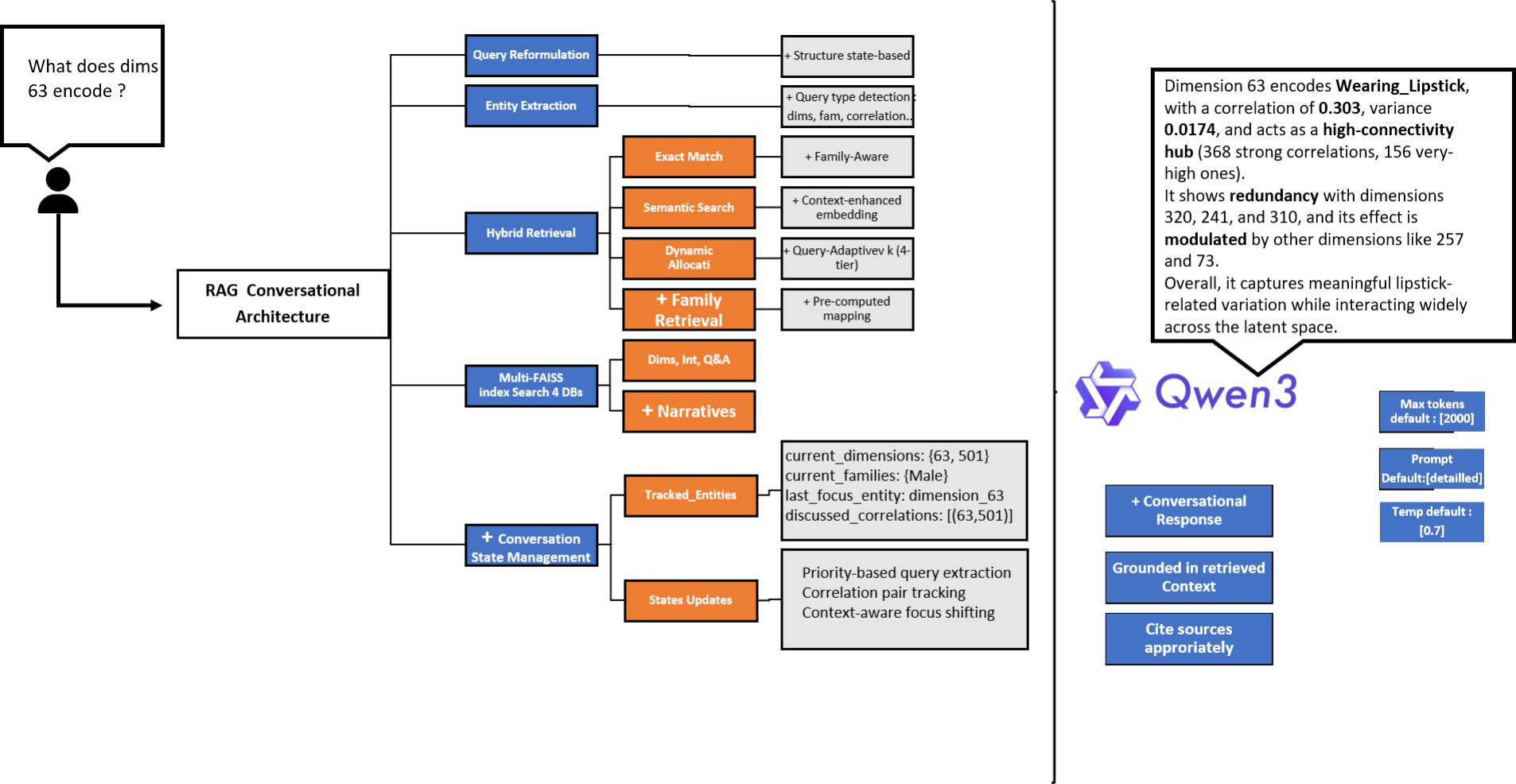}
\caption{VAE RAG Conversational Architecture and SLM Ablation Study Design. \textbf{Top}: The system processes user queries (e.g., ``What does dimension 63 encode?'') through: (1) Query Reformulation with structure-based state tracking, (2) Entity Extraction detecting query types (dimensions, families, correlations), (3) Hybrid Retrieval combining Exact Match (family-aware), Semantic Search (context-enhanced embedding), Dynamic Allocation (query-adaptive $k$ with 4-tier budget), and Family Retrieval (pre-computed mapping), (4) Multi-FAISS Index Search across 4 databases (Dimensions, Interactions, Q\&A, Narratives), and (5) Conversation State Management tracking current dimensions, families, focus entities, and discussed correlations. The SLM (Qwen3-8B) generates grounded responses citing specific statistics from retrieved Manifestation Units. \textbf{Bottom}: Ablation study design showing parameter configurations for Temperature Study (varying 0.1--1.0), Prompt Study (minimal, structured, cot, detailed), and Max Tokens Study (500--4000).}
\label{fig:vae_slm_arch}
\end{figure*}

\begin{figure*}[t]
\centering
\includegraphics[width=\textwidth]{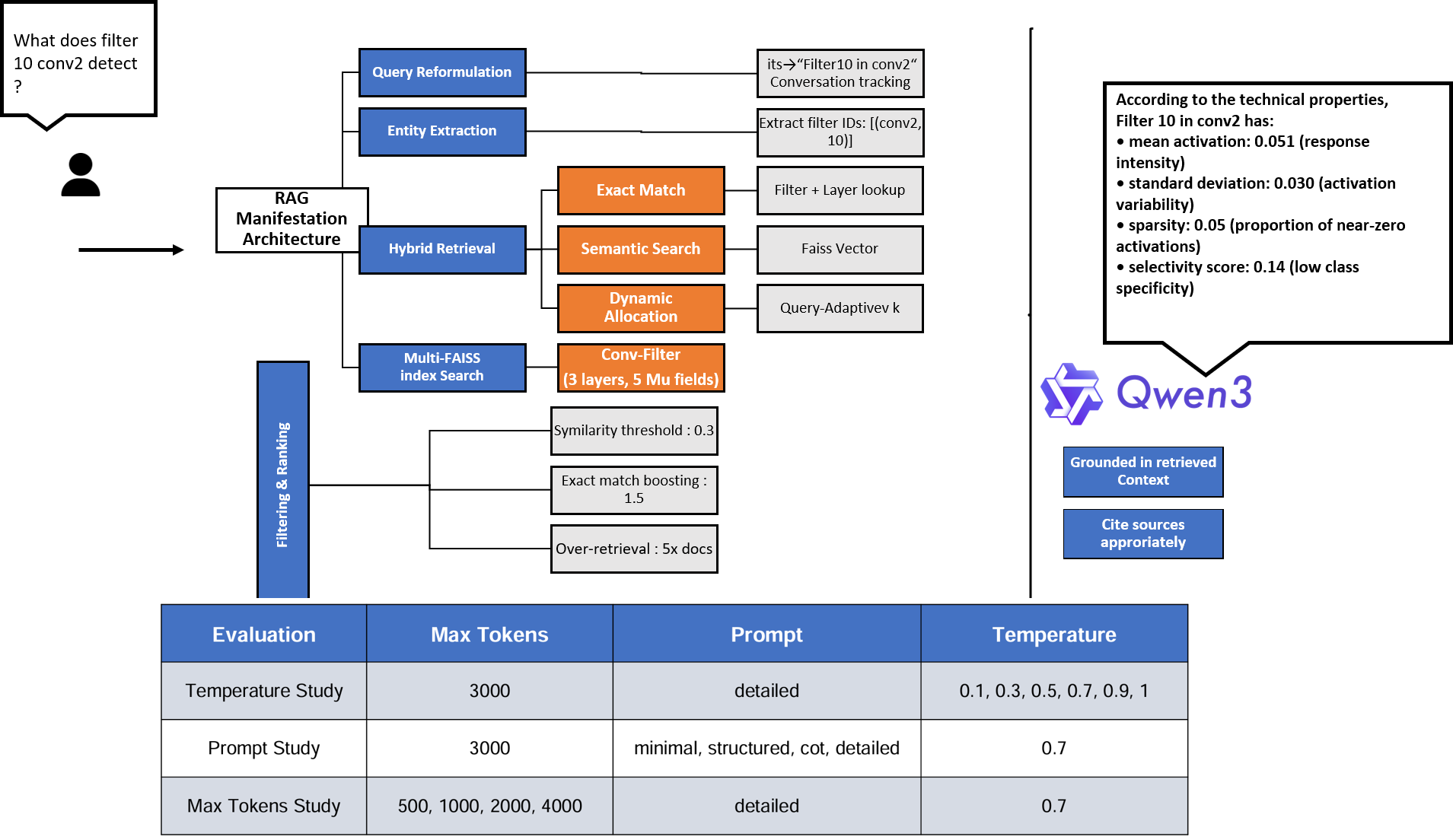}
\caption{CNN RAG Manifestation Architecture and SLM Ablation Study Design. \textbf{Top}: The architecture processes filter queries (e.g., ``What does filter 10 in conv2 detect?'') through: (1) Query Reformulation with conversation tracking (pronoun resolution), (2) Entity Extraction identifying filter IDs and layers, (3) Hybrid Retrieval with Exact Match (filter + layer lookup), Semantic Search (FAISS vector), and Dynamic Allocation (query-adaptive $k$), (4) Single unified FAISS index (Conv-Filter: 3 layers, 5 MU fields), and (5) Filtering \& Ranking with similarity threshold (0.3), exact match boosting ($1.5\times$), and over-retrieval ($5\times$ candidates). The SLM generates responses grounded in retrieved context with appropriate source citations. \textbf{Bottom}: Identical ablation study design to VAE, enabling cross-architecture comparison.}
\label{fig:cnn_slm_arch}
\end{figure*}

Figure~\ref{fig:vae_slm_arch} and~\ref{fig:cnn_slm_arch} illustrate the RAG conversational architectures
for the VAE and CNN systems, respectively.

\subsection{SLM Configuration Parameters}
\label{app:slm_params}

Both conversational systems use Qwen3-8B as the generation backbone with configurable parameters
for ablation studies. Table~\ref{tab:slm_params} summarizes the parameter space explored.

\begin{table}[h]
\centering
\caption{SLM configuration parameters and their ablation ranges.}
\label{tab:slm_params}
\begin{tabular}{llll}
\toprule
\textbf{Parameter} & \textbf{Default} & \textbf{Ablation Range} & \textbf{Description} \\
\midrule
\texttt{model\_name} & qwen3:8b &   & Base language model \\
\texttt{temperature} & 0.7 & [0.1, 0.3, 0.5, 0.7, 0.9, 1.0] & Sampling temperature \\
\texttt{max\_tokens} & 2000 & [500, 1000, 2000, 4000] & Maximum output tokens \\
\texttt{num\_predict} & 3000 &   & Ollama prediction budget \\
\texttt{prompt\_method} & detailed & [minimal, structured, cot, detailed] & Prompt engineering strategy \\
\texttt{top\_p} & 0.9 &   & Nucleus sampling threshold \\
\texttt{top\_k} & 20--40 &   & Top-$k$ sampling cutoff \\
\bottomrule
\end{tabular}
\end{table}

\paragraph{Ablation study experimental design.}
Table~\ref{tab:ablation_design} summarizes the experimental configurations for each ablation study.
Each study varies one parameter while holding others constant, enabling isolated analysis of parameter effects.

\begin{table}[h]
\centering
\caption{Ablation study experimental design. Each study varies one parameter (bold) while holding others at default values.}
\label{tab:ablation_design}
\begin{tabular}{lccc}
\toprule
\textbf{Evaluation Study} & \textbf{Max Tokens} & \textbf{Prompt Method} & \textbf{Temperature} \\
\midrule
Temperature Study & 3000 & detailed & \textbf{0.1, 0.3, 0.5, 0.7, 0.9, 1.0} \\
Prompt Study & 3000 & \textbf{minimal, structured, cot, detailed} & 0.7 \\
Max Tokens Study & \textbf{500, 1000, 2000, 4000} & detailed & 0.7 \\
\bottomrule
\end{tabular}
\end{table}

\paragraph{Parameter definitions.}
\begin{itemize}
  \item \textbf{temperature}: Controls randomness in token sampling. Lower values (0.1--0.3) produce
        more deterministic, focused outputs; higher values (0.7--1.0) increase diversity and creativity.
  \item \textbf{max\_tokens}: Maximum number of tokens the model can generate per response.
        Affects response comprehensiveness and detail level.
  \item \textbf{prompt\_method}: Strategy for constructing the generation prompt from retrieved context:
        \begin{itemize}
          \item \emph{minimal}: Raw context concatenation with simple question-answer format
          \item \emph{structured}: Organized sections with clear field delineation
          \item \emph{cot} (Chain-of-Thought): Step-by-step reasoning instructions
          \item \emph{detailed}: Comprehensive system instructions with grounding rules, domain knowledge, and structured output format (default)
        \end{itemize}
\end{itemize}

\subsection{Prompt Engineering Methods}
\label{app:prompt_methods}

We evaluate four prompt engineering strategies, each representing a different level of instruction specificity.

\paragraph{Minimal prompt.}
Direct context injection with no system instructions:
\begin{verbatim}
Context:
{retrieved_documents}

Question: {query}

Answer:
\end{verbatim}

\paragraph{Structured prompt.}
Organized sections separating context by source type:
\begin{verbatim}
You are a {VAE/CNN} analysis assistant.

[DIMENSION/FILTER DATA]
{dimension/filter context grouped by entity}

[REFERENCE Q&A]
{qa context}

[QUESTION]
{query}

[ANSWER]
\end{verbatim}

\paragraph{Chain-of-Thought (CoT) prompt.}
Explicit reasoning steps before final answer:
\begin{verbatim}
You are analyzing a {VAE/CNN} trained on {CelebA/CIFAR-10}.

Available information:
{bulleted context items}

Question: {query}

Think step by step:
1. What specific information from the context is relevant?
2. What does this information tell us?
3. What can we conclude?

Final Answer:
\end{verbatim}

\paragraph{Detailed prompt.}
Comprehensive instructions including grounding rules, domain knowledge, and output format specifications.
This method includes:
\begin{itemize}
  \item System role definition with architecture-specific expertise
  \item Explicit grounding rules (``cite specific data'', ``distinguish facts from interpretation'')
  \item Domain terminology (semantic families, layer hierarchy, correlation semantics)
  \item Structured output format with metric tags for automatic verification
  \item Example question-answer pairs demonstrating expected response style
\end{itemize}
See Appx.~\ref{app:prompt} for the complete detailed prompt template.

\subsection{Ablation Study: Maximum Token Length}
\label{app:ablation_tokens}

We evaluate the impact of maximum token limits on generation quality while holding retrieval parameters constant ($k=20$).
All experiments use 50 queries per system.

Table~\ref{tab:vae_token_results} presents the VAE system results across token settings.

\begin{table}[h]
\centering
\caption{VAE system: Maximum token ablation results. Retrieval metrics remain constant across all token settings. Best results per metric are \textbf{bolded}.}
\label{tab:vae_token_results}
\resizebox{\textwidth}{!}{%
\begin{tabular}{l|ccc|ccc|ccc}
\toprule
 & \multicolumn{3}{c|}{\textbf{Retrieval ($k$=20)}} & \multicolumn{3}{c|}{\textbf{Generation Quality}} & \multicolumn{3}{c}{\textbf{Faithfulness \& Factual}} \\
\textbf{Max Tokens} & Recall & Prec & NDCG & F1$\uparrow$ & RG-L$\uparrow$ & BERT$\uparrow$ & Attr$\uparrow$ & Sem$\uparrow$ & Fact$\uparrow$ \\
\midrule
500 & 1.000 & 0.468 & 0.963 & 0.259 & 0.166 & 0.810 & \textbf{0.618} & \textbf{0.448} & 0.902 \\
1000 & 1.000 & 0.468 & 0.963 & \textbf{0.394} & \textbf{0.230} & \textbf{0.837} & 0.612 & 0.356 & \textbf{0.984} \\
2000 & 1.000 & 0.468 & 0.963 & 0.383 & 0.219 & 0.836 & 0.597 & 0.339 & 0.980 \\
4000 & 1.000 & 0.468 & 0.963 & 0.393 & 0.225 & \textbf{0.837} & 0.571 & 0.321 & 0.985 \\
\bottomrule
\end{tabular}%
}
\end{table}

Table~\ref{tab:vae_token_stability} shows the stability analysis across token settings.

\begin{table}[h]
\centering
\caption{VAE system: Stability analysis across maximum token settings. Retrieval metrics show zero variance (perfect stability), while generation metrics exhibit controlled variability.}
\label{tab:vae_token_stability}
\begin{tabular}{llcccc}
\toprule
\textbf{Category} & \textbf{Metric} & \textbf{Mean} & \textbf{Std Dev} & \textbf{Min} & \textbf{Max} \\
\midrule
\multirow{3}{*}{Retrieval} & Recall@20 & 1.000 & 0.000 & 1.000 & 1.000 \\
 & Precision@20 & 0.468 & 0.000 & 0.468 & 0.468 \\
 & NDCG@20 & 0.963 & 0.000 & 0.963 & 0.963 \\
\midrule
\multirow{3}{*}{Generation} & F1 Score & 0.357 & 0.061 & 0.259 & 0.394 \\
 & ROUGE-L & 0.210 & 0.029 & 0.166 & 0.230 \\
 & BERTScore & 0.830 & 0.012 & 0.810 & 0.837 \\
\midrule
\multirow{2}{*}{Faithfulness} & Attribution & 0.600 & 0.020 & 0.571 & 0.618 \\
 & Semantic Overlap & 0.366 & 0.054 & 0.321 & 0.448 \\
\midrule
Factual & Accuracy & 0.963 & 0.039 & 0.902 & 0.985 \\
\bottomrule
\end{tabular}
\end{table}

\paragraph{Key findings.}
\begin{itemize}
  \item \textbf{Retrieval stability}: All retrieval metrics show zero variance across token settings,
        confirming that generation parameters do not affect retrieval quality.
  \item \textbf{Optimal balance}: Token limit of 1000 achieves the best trade-off between generation quality
        (highest F1, ROUGE-L) and faithfulness (highest factual accuracy at 98.4\%).
  \item \textbf{Diminishing returns}: Increasing beyond 1000 tokens improves BERTScore marginally (0.837)
        but decreases Attribution (0.571 at 4000 vs.\ 0.618 at 500), suggesting longer responses
        include more content not directly grounded in retrieved evidence.
\end{itemize}

\subsection{Ablation Study: Prompt Engineering Methods}
\label{app:ablation_prompt}

We evaluate four prompt engineering strategies while holding other parameters constant
(temperature=0.7, max\_tokens=3000, $k=20$).

Table~\ref{tab:prompt_ablation} compares prompt methods across key metrics for both systems.

\begin{table}[h]
\centering
\caption{Prompt engineering comparison across VAE and CNN systems. Best values per metric are \textbf{bolded}.}
\label{tab:prompt_ablation}
\resizebox{\textwidth}{!}{%
\begin{tabular}{l|ccc|ccc|ccc|ccc}
\toprule
 & \multicolumn{6}{c|}{\textbf{VAE System}} & \multicolumn{6}{c}{\textbf{CNN System}} \\
\cmidrule(lr){2-7} \cmidrule(lr){8-13}
\textbf{Method} & F1$\uparrow$ & RG-L$\uparrow$ & BERT$\uparrow$ & Attr$\uparrow$ & Sem$\uparrow$ & Fact$\uparrow$ & F1$\uparrow$ & RG-L$\uparrow$ & BERT$\uparrow$ & Attr$\uparrow$ & Sem$\uparrow$ & Fact$\uparrow$ \\
\midrule
Minimal & 0.330 & 0.198 & 0.841 & \textbf{0.619} & 0.389 & 0.988 & 0.162 & 0.120 & 0.838 & \textbf{1.000} & \textbf{0.653} & \textbf{1.000} \\
Structured & 0.359 & 0.200 & \textbf{0.842} & 0.596 & 0.347 & 0.988 & 0.176 & 0.127 & 0.838 & 0.997 & 0.656 & 0.935 \\
CoT & 0.359 & \textbf{0.224} & 0.834 & 0.587 & 0.380 & \textbf{0.991} & 0.264 & 0.169 & 0.832 & 0.516 & 0.404 & 0.959 \\
Detailed & \textbf{0.390} & \textbf{0.224} & \textbf{0.842} & 0.609 & 0.368 & 0.985 & \textbf{0.396} & \textbf{0.219} & \textbf{0.850} & 0.682 & 0.363 & 0.939 \\
\bottomrule
\end{tabular}%
}
\end{table}

Table~\ref{tab:prompt_stability} shows the stability analysis across prompt methods.

\begin{table}[h]
\centering
\caption{Stability analysis across prompt engineering methods. Retrieval metrics show zero variance, while generation metrics exhibit controlled variability.}
\label{tab:prompt_stability}
\begin{tabular}{llcccc}
\toprule
\textbf{Category} & \textbf{Metric} & \textbf{Mean} & \textbf{Std Dev} & \textbf{Min} & \textbf{Max} \\
\midrule
\multirow{3}{*}{Retrieval} & Recall@20 & 1.000 & 0.000 & 1.000 & 1.000 \\
 & Precision@20 & 0.468 & 0.000 & 0.468 & 0.468 \\
 & NDCG@20 & 0.963 & 0.000 & 0.963 & 0.963 \\
\midrule
\multirow{3}{*}{Generation} & F1 Score & 0.360 & 0.024 & 0.330 & 0.390 \\
 & ROUGE-L & 0.211 & 0.014 & 0.198 & 0.224 \\
 & BERTScore & 0.838 & 0.004 & 0.834 & 0.842 \\
\midrule
\multirow{2}{*}{Faithfulness} & Attribution & 0.603 & 0.014 & 0.587 & 0.619 \\
 & Semantic Overlap & 0.371 & 0.018 & 0.347 & 0.389 \\
\midrule
Factual & Accuracy & 0.988 & 0.002 & 0.985 & 0.991 \\
\bottomrule
\end{tabular}
\end{table}

\paragraph{Key findings.}
\begin{itemize}
  \item \textbf{Generation quality}: Detailed prompting consistently achieves highest F1 and ROUGE-L
        scores across both systems, with particularly strong gains on CNN (+0.234 F1 vs.\ Minimal).
  \item \textbf{Faithfulness trade-off}: Minimal prompting achieves perfect Attribution on CNN (1.000)
        but lower generation quality. This suggests minimal prompts produce responses that closely
        mirror retrieved content but lack interpretive depth.
  \item \textbf{Factual accuracy}: CoT achieves highest factual accuracy on VAE (0.991), while
        Minimal achieves perfect accuracy on CNN (1.000). Both demonstrate that explicit grounding
        instructions improve factual reliability.
  \item \textbf{System differences}: CNN shows higher variance across prompt methods than VAE,
        indicating the simpler architecture is more sensitive to prompt engineering choices.
\end{itemize}

\subsection{Ablation Study: Temperature}
\label{app:ablation_temp}

We evaluate six temperature settings from deterministic (0.1) to high creativity (1.0)
while holding other parameters constant (prompt\_method=detailed, max\_tokens=3000, $k=20$).

Table~\ref{tab:vae_temp_results} presents the VAE temperature ablation results.

\begin{table}[h]
\centering
\caption{VAE system: Temperature ablation results. Best results per metric are \textbf{bolded}.}
\label{tab:vae_temp_results}
\resizebox{\textwidth}{!}{%
\begin{tabular}{l|ccc|ccc|ccc}
\toprule
 & \multicolumn{3}{c|}{\textbf{Retrieval ($k$=20)}} & \multicolumn{3}{c|}{\textbf{Generation Quality}} & \multicolumn{3}{c}{\textbf{Faithfulness \& Factual}} \\
\textbf{Temp.} & Recall & Prec & NDCG & F1$\uparrow$ & RG-L$\uparrow$ & BERT$\uparrow$ & Attr$\uparrow$ & Sem$\uparrow$ & Fact$\uparrow$ \\
\midrule
0.1 & 1.000 & 0.468 & 0.963 & 0.386 & 0.219 & 0.835 & 0.580 & 0.351 & 0.975 \\
0.3 & 1.000 & 0.468 & 0.963 & 0.391 & 0.223 & 0.837 & 0.587 & 0.309 & \textbf{0.985} \\
0.5 & 1.000 & 0.468 & 0.963 & \textbf{0.395} & \textbf{0.231} & \textbf{0.838} & 0.617 & \textbf{0.379} & 0.978 \\
0.7 & 1.000 & 0.468 & 0.963 & 0.381 & 0.221 & 0.834 & \textbf{0.626} & 0.372 & 0.972 \\
0.9 & 1.000 & 0.468 & 0.963 & 0.391 & 0.220 & 0.836 & 0.606 & 0.337 & 0.981 \\
1.0 & 1.000 & 0.468 & 0.963 & 0.387 & 0.223 & 0.837 & 0.574 & 0.321 & 0.981 \\
\bottomrule
\end{tabular}%
}
\end{table}

Table~\ref{tab:cnn_temp_results} presents the CNN temperature ablation results.

\begin{table}[h]
\centering
\caption{CNN system: Temperature ablation results. Best results per metric are \textbf{bolded}.}
\label{tab:cnn_temp_results}
\resizebox{\textwidth}{!}{%
\begin{tabular}{l|ccc|ccc|ccc}
\toprule
 & \multicolumn{3}{c|}{\textbf{Retrieval ($k$=20)}} & \multicolumn{3}{c|}{\textbf{Generation Quality}} & \multicolumn{3}{c}{\textbf{Faithfulness \& Factual}} \\
\textbf{Temp.} & Recall & Prec & NDCG & F1$\uparrow$ & RG-L$\uparrow$ & BERT$\uparrow$ & Attr$\uparrow$ & Sem$\uparrow$ & Fact$\uparrow$ \\
\midrule
0.1 & 1.000 & 0.468 & 0.963 & 0.388 & \textbf{0.224} & 0.851 & 0.655 & 0.376 & 0.918 \\
0.3 & 1.000 & 0.468 & 0.963 & 0.382 & 0.214 & 0.851 & 0.662 & 0.359 & \textbf{0.956} \\
0.5 & 1.000 & 0.468 & 0.963 & 0.382 & 0.220 & 0.849 & 0.629 & 0.337 & 0.927 \\
0.7 & 1.000 & 0.468 & 0.963 & 0.386 & 0.221 & 0.849 & 0.691 & 0.375 & 0.939 \\
0.9 & 1.000 & 0.468 & 0.963 & \textbf{0.401} & \textbf{0.226} & \textbf{0.854} & 0.648 & 0.358 & 0.945 \\
1.0 & 1.000 & 0.468 & 0.963 & 0.388 & 0.217 & 0.850 & \textbf{0.704} & \textbf{0.372} & 0.927 \\
\bottomrule
\end{tabular}%
}
\end{table}

Table~\ref{tab:temp_stability} shows stability analysis with coefficient of variation.

\begin{table}[h]
\centering
\caption{Temperature stability analysis. All generation metrics exhibit CV below 5\%, indicating high stability across temperature range.}
\label{tab:temp_stability}
\begin{tabular}{llcccc}
\toprule
\textbf{Category} & \textbf{Metric} & \textbf{Mean} & \textbf{Std Dev} & \textbf{CV (\%)} \\
\midrule
\multirow{3}{*}{Retrieval} & Recall@20 & 1.000 & 0.000 & 0.0 \\
 & Precision@20 & 0.468 & 0.000 & 0.0 \\
 & NDCG@20 & 0.963 & 0.000 & 0.0 \\
\midrule
\multirow{3}{*}{Generation} & F1 Score & 0.389 & 0.005 & 1.3 \\
 & ROUGE-L & 0.223 & 0.004 & 1.8 \\
 & BERTScore & 0.836 & 0.002 & 0.2 \\
\midrule
\multirow{2}{*}{Faithfulness} & Attribution & 0.598 & 0.019 & 3.2 \\
 & Semantic Overlap & 0.345 & 0.026 & 7.5 \\
\midrule
Factual & Accuracy & 0.979 & 0.004 & 0.4 \\
\bottomrule
\end{tabular}
\end{table}

\paragraph{Key findings.}
\begin{itemize}
  \item \textbf{High stability}: All generation metrics exhibit coefficient of variation (CV) below 5\%
        across the temperature range, indicating the SLM produces consistent quality outputs regardless of sampling temperature.
  \item \textbf{System-specific optima}: VAE prefers moderate temperatures (0.5--0.7) while CNN
        prefers higher temperatures (0.9--1.0) for generation quality. This difference may reflect
        the complexity of the underlying analysis: VAE's richer semantic structure benefits from
        more focused generation, while CNN's simpler filter descriptions benefit from more diverse phrasing.
  \item \textbf{Factual accuracy}: Both systems achieve highest factual accuracy at lower temperatures (0.3),
        suggesting deterministic generation reduces hallucination risk.
  \item \textbf{Default recommendation}: Temperature 0.7 provides a robust default balancing generation
        quality, faithfulness, and response diversity for both systems.
\end{itemize}

\subsection{Implementation Details}
\label{app:slm_impl}

\paragraph{Conversation state management.}
Both systems maintain structured conversation state for context-aware query handling:

\begin{verbatim}
# VAE conversation state
conversation_state = {
    "current_dimensions": set(),      # e.g., {63, 501}
    "current_families": set(),        # e.g., {"Male"}
    "last_focus_entity": str,         # e.g., "dimension_63"
    "discussed_correlations": list    # e.g., [(63, 501)]
}

# CNN conversation state
conversation_state = {
    "current_filters": set(),         # e.g., {("conv2", 10)}
    "current_layers": set(),          # e.g., {"conv2"}
    "current_classes": set(),         # e.g., {"automobile"}
    "last_focus_entity": str,         # e.g., "filter_conv2_10"
    "discussed_relationships": list   # e.g., [(("conv1", 5), ("conv1", 16))]
}
\end{verbatim}

\paragraph{Query reformulation.}
The system resolves pronouns and implicit references using conversation state:
\begin{itemize}
  \item ``its correlation'' $\rightarrow$ ``dimension 63's correlation'' (VAE)
  \item ``in the same family'' $\rightarrow$ ``in the Male family'' (VAE)
  \item ``filter 10's partners'' $\rightarrow$ ``filter 10 in conv2's partners'' (CNN)
  \item ``other filters in this layer'' $\rightarrow$ ``other filters in conv2 besides filter 10'' (CNN)
\end{itemize}

\paragraph{Response generation.}
The SLM is invoked via Ollama's local API:
\begin{verbatim}
response = requests.post(
    "http://localhost:11434/api/generate",
    json={
        "model": "qwen3:8b",
        "prompt": grounded_prompt,
        "stream": False,
        "options": {
            "temperature": 0.7,
            "num_predict": 3000,
            "max_tokens": 2000,
            "top_p": 0.9,
            "top_k": 20
        }
    }
)
\end{verbatim}

\paragraph{Thinking tag removal.}
Qwen3's reasoning traces (enclosed in \texttt{<think>...</think>} tags) are stripped from responses
unless explicitly requested, ensuring clean output for evaluation and user interaction.

\subsection{Architectural Differences Summary}
\label{app:arch_diff}

Table~\ref{tab:arch_comparison} summarizes the key architectural differences between the VAE and CNN conversational systems.

\begin{table}[h]
\centering
\caption{Architectural comparison between VAE and CNN conversational systems.}
\label{tab:arch_comparison}
\resizebox{\textwidth}{!}{%
\begin{tabular}{lcc}
\toprule
\textbf{Component} & \textbf{VAE System} & \textbf{CNN System} \\
\midrule
\multicolumn{3}{l}{\textit{Model Scope}} \\
Components & 512 latent dimensions & 224 filters (32+64+128) \\
Attributes & 40 CelebA facial attributes & 10 CIFAR-10 classes \\
Semantic groupings & 20 families & 3 layers (conv1, conv2, conv3) \\
\midrule
\multicolumn{3}{l}{\textit{Index Architecture}} \\
Number of indices & 4 (Dimension, Q\&A, Interaction, Narrative) & 1 (unified) \\
Total documents & 5,457 & 943 \\
Fields per unit & 4 & 5 \\
\midrule
\multicolumn{3}{l}{\textit{Retrieval Strategy}} \\
Dynamic allocation & Yes (4-tier query-adaptive $k$) & Yes (query-adaptive $k$) \\
Family-aware retrieval & Yes (pre-computed mapping) & No \\
Exact match boosting & $1.5\times$ & $1.5\times$ \\
Similarity threshold & 0.3 & 0.3 \\
\midrule
\multicolumn{3}{l}{\textit{Conversation State}} \\
Entity tracking & dimensions, families, correlations & filters, layers, classes \\
Relationship tracking & correlation pairs & synergistic partnerships \\
\midrule
\multicolumn{3}{l}{\textit{Default SLM Configuration}} \\
Temperature & 0.7 & 0.7 \\
Max tokens & 2000 & 2000 \\
Prompt method & detailed & detailed \\
\bottomrule
\end{tabular}%
}
\end{table}

\section{Manipulation Details}
\label{app:manipulation}

\subsection{VAE Manipulation}

Given a target concept $c_{\text{target}}$ (e.g., ``Smiling''):

\begin{enumerate}
    \item Retrieve dimensions where $|s_{d, c_{\text{target}}}| > \tau_s$ (default $\tau_s = 0.3$).
    \item Select top-$n$ dimensions by $|s_{d, c_{\text{target}}}|$ (default $n=3$).
    \item For each selected dimension $d$, modify:
    \begin{equation}
    z'_d = z_d + \text{sign}(s_{d, c_{\text{target}}}) \cdot \text{step} \cdot \sigma_d
    \end{equation}
    where $\sigma_d$ is the standard deviation of dimension $d$ across the training set, and step is from $G$ (default 2.0).
    \item Decode: $x' = p_\theta(z')$.
\end{enumerate}

\subsection{CNN Manipulation}

Given a target class $c_{\text{target}}$ (e.g., ``dog''):

\begin{enumerate}
    \item Retrieve filters where selectivity for $c_{\text{target}}$ exceeds threshold.
    \item Select top-$n$ filters by selectivity (default $n=3$).
    \item For each selected filter $f$, we apply amplification to the post-ReLU feature map:
    \begin{equation}
    a'_f = \alpha \cdot a_f
    \end{equation}
    where $\alpha_0$ is a conservative starting gain stored in $G$ for interactive use (default 1.5). For evaluation, we override this and sweep $\alpha \in \{10, 30, 50, 70, 90, 100\}$ (see Sec.~\ref{sec:evaluation}).
    \item Forward pass with modified activations and measure probability shift.
\end{enumerate}

\subsection{Manipulation Validation}
\label{app:manipulation_validation}

To validate that excavated knowledge is both \textbf{actionable} (enables successful manipulations) and \textbf{faithful} (retrieved components are semantically correct), we conducted comprehensive evaluation across both architectures.

\subsubsection{CNN Automatic Evaluation}

We systematically tested all 90 source$\to$target class pairs (10 classes $\times$ 9 targets) across 6 amplification levels ($\alpha \in \{10, 30, 50, 70, 90, 100\}$), yielding 540 total manipulation attempts.

\begin{figure}[h]
\centering
\includegraphics[width=0.75\linewidth]{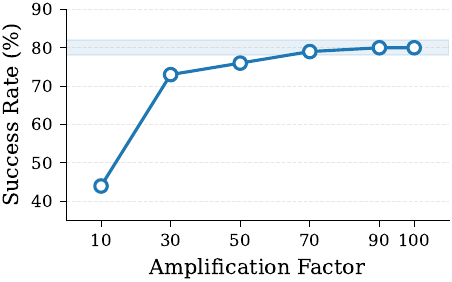}
\caption{CNN manipulation success rate by amplification factor. Performance plateaus around 80\% for $\alpha \geq 70$.}
\label{fig:amplification}
\end{figure}

\paragraph{Success Rate by Amplification.} Figure~\ref{fig:amplification} shows success rates by amplification level. Performance increases sharply from 44\% at $\alpha=10$ to approximately 80\% at $\alpha \geq 70$, with optimal performance in the 70--100$\times$ range. The overall success rate across all amplification levels is 72\%.

\begin{figure}[h]
\centering
\includegraphics[width=0.65\linewidth]{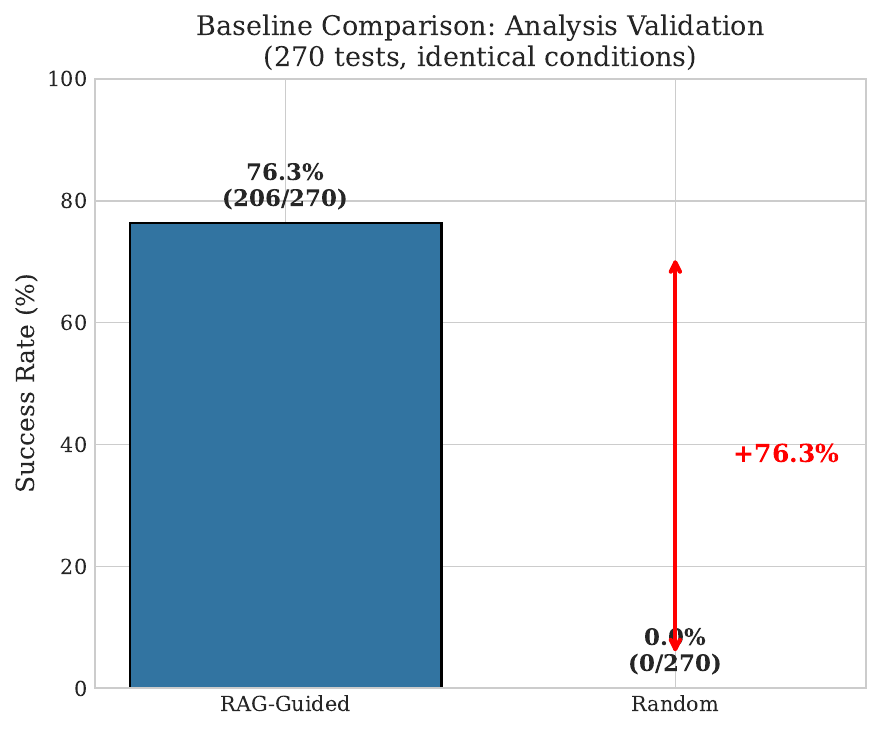}
\caption{RAG-guided vs.\ random baseline (single seed, $n{=}270$). Random filter selection achieves 0\% success; RAG-guided achieves 76.3\%. The three-seed mean for random (Table~\ref{tab:rag_decompose}) is 6.7\%, confirming the ${\approx}70$pp gap is robust. Correct filter selection---not arbitrary amplification---is what drives the difference.}
\label{fig:rag_vs_random}
\end{figure}

\paragraph{Baseline Comparison: RAG-Guided vs Random Selection.}
To validate that success stems from meaningful filter-class mappings rather than noise exploitation, we compared RAG-guided filter selection against a random baseline. Both methods used identical conditions: same amplification factors ($\alpha \in \{30, 50, 70\}$), same number of filters, and same source images. The \textit{only} difference was the filter selection criterion (Figure~\ref{fig:rag_vs_random}).

\begin{table}[h]
\centering
\caption{Baseline comparison validates that excavated knowledge is meaningful. This visualisation run used a single random seed and obtained 0\% success (0/270). The three-seed mean reported in Table~\ref{tab:rag_decompose} is 6.7\% (18/270); single seeds vary between 0 and $\sim$10\%, confirming the large gap to RAG-guided selection (76.3\%) is robust to seed choice.}
\label{tab:baseline_comparison}
\small
\begin{tabular}{lccc}
\toprule
Method & Success Rate & Tests & Description \\
\midrule
RAG-Guided & 76.3\% & 270 & Filters from MANIFESTATION analysis \\
Random Baseline & 0.0\% & 270 & Random filters, single seed (3-seed mean: 6.7\%) \\
\bottomrule
\end{tabular}
\end{table}

\begin{table}[h]
\centering
\caption{Detailed baseline comparison by amplification factor.}
\label{tab:baseline_by_amp}
\small
\begin{tabular}{lccc}
\toprule
Amplification & RAG-Guided & Random & $\Delta$ \\
\midrule
30$\times$ & 73.3\% & 0.0\% & +73.3\% \\
50$\times$ & 75.6\% & 0.0\% & +75.6\% \\
70$\times$ & 80.0\% & 0.0\% & +80.0\% \\
\bottomrule
\end{tabular}
\end{table}

Table~\ref{tab:baseline_by_amp} shows the breakdown by amplification factor for this single-seed visualisation run. The near-zero random success (0\% here; 6.7\% three-seed mean in Table~\ref{tab:rag_decompose}) provides strong evidence that MANIFESTATION's excavated filter-class mappings capture genuine semantic structure rather than reflecting noise exploitation.

\subsubsection{Class-Pair Analysis}

\begin{figure}[h]
\centering
\includegraphics[width=0.85\linewidth]{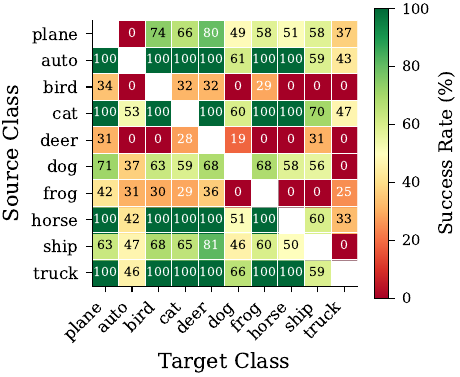}
\caption{Manipulation success rate by class pair. Values indicate success rate (\%) across amplification levels. Notable patterns include high success for horse and truck as sources, and difficulty with automobile as target.}
\label{fig:heatmap}
\end{figure}

Figure~\ref{fig:heatmap} shows the manipulation success rate for each source$\to$target class pair, revealing systematic patterns in manipulation difficulty.

\begin{figure}[h]
\centering
\includegraphics[width=0.6\linewidth]{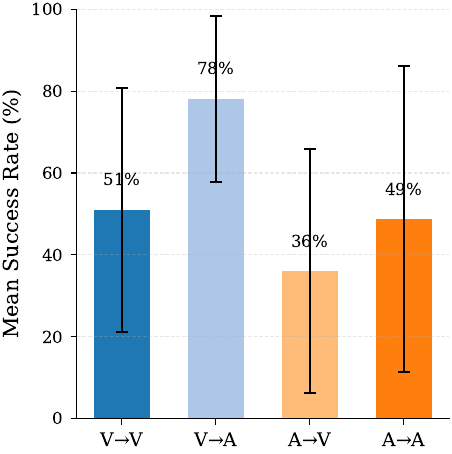}
\caption{Success rate by transformation category. V$\to$A (vehicle to animal) achieves highest success (78\%), while A$\to$V (animal to vehicle) is most challenging (36\%).}
\label{fig:category}
\end{figure}

\paragraph{Category-Level Analysis.}
We grouped CIFAR-10 classes into vehicles (airplane, automobile, ship, truck) and animals (bird, cat, deer, dog, frog, horse) to analyze cross-category manipulation patterns (Figure~\ref{fig:category}).

\begin{table}[h]
\centering
\caption{Manipulation success rate by category transformation.}
\label{tab:category}
\small
\begin{tabular}{lcc}
\toprule
Transformation & Success Rate & Interpretation \\
\midrule
V $\to$ A (Vehicle to Animal) & 78\% & Easiest \\
V $\to$ V (Vehicle to Vehicle) & 51\% & Moderate \\
A $\to$ A (Animal to Animal) & 49\% & Moderate \\
A $\to$ V (Animal to Vehicle) & 36\% & Hardest \\
\bottomrule
\end{tabular}
\end{table}

Table~\ref{tab:category} summarizes the results. The asymmetry between V$\to$A (78\%) and A$\to$V (36\%) suggests that vehicle features are more easily suppressed than animal features, possibly because the CNN encodes animal classes with more distributed, redundant filter responses.

\begin{figure}[h]
\centering
\includegraphics[width=0.85\linewidth]{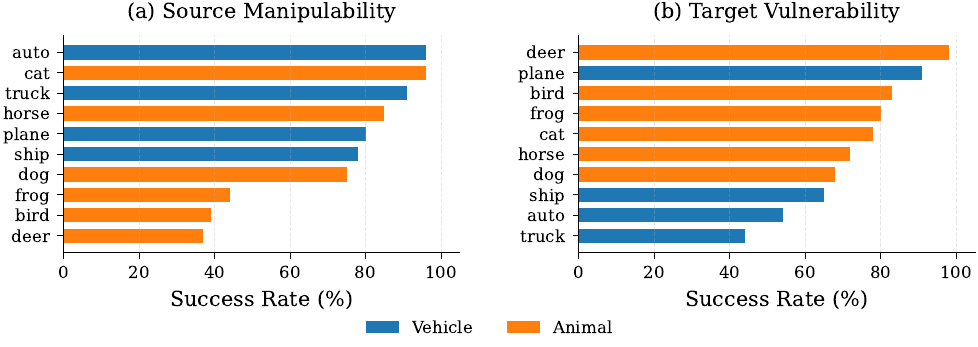}
\caption{(a) Source manipulability: automobile and cat are easiest to transform. (b) Target vulnerability: deer and airplane are easiest targets.}
\label{fig:asymmetry}
\end{figure}

\paragraph{Asymmetric Manipulability.}
Figure~\ref{fig:asymmetry} reveals that classes differ significantly in both source manipulability (how easily they can be transformed away from) and target vulnerability (how easily other images can be transformed into them).

Key observations:
\begin{itemize}[itemsep=1pt]
    \item \textbf{Most manipulable sources}: automobile, cat, truck, horse
    \item \textbf{Least manipulable sources}: deer, bird, frog
    \item \textbf{Most vulnerable targets}: deer, airplane, bird
    \item \textbf{Least vulnerable targets}: truck, automobile
\end{itemize}

\subsubsection{CNN Human Evaluation}

To validate that automatic metrics align with human perception, evaluators assessed manipulation quality on a subset of tasks (14 evaluations).

\paragraph{Metrics.} Evaluators rated each manipulation on:
\begin{itemize}[itemsep=1pt]
    \item \textbf{Classification Success}: Did the manipulation achieve the requested class change? (Yes/Partially/No)
    \item \textbf{Filter Explanation Quality}: Were the selected filters meaningfully explained? (Yes/Partially/No)
    \item \textbf{Manipulation Coherence}: Did the manipulation make logical sense? (1--5 scale)
    \item \textbf{Task Difficulty}: Perceived difficulty of the transformation (Easy/Medium/Hard/Very Hard)
    \item \textbf{Iterations Needed}: Number of attempts required (1/2/3--4/5+)
\end{itemize}

\paragraph{Results.}
Table~\ref{tab:cnn_human_eval} summarizes the CNN human evaluation results.

\begin{table}[h]
\centering
\caption{CNN human manipulation evaluation results (14 evaluations).}
\label{tab:cnn_human_eval}
\small
\begin{tabular}{lc}
\toprule
Metric & Result \\
\midrule
Classification Success & Yes: 79\%, No: 21\% \\
Filter Explanation Quality & Yes: 64\%, Partially: 29\%, No: 7\% \\
Manipulation Coherence & 3.64 $\pm$ 1.28 / 5.0 \\
Task Difficulty & Easy: 36\%, Medium: 50\%, Hard: 14\% \\
Iterations Needed & 1: 57\%, 2: 21\%, 3--4: 14\%, 5+: 7\% \\
\bottomrule
\end{tabular}
\end{table}

\paragraph{Human-Automatic Alignment.}
Table~\ref{tab:human_auto_alignment} shows that human coherence ratings correlate with automatic classification success, validating that automatic metrics serve as reliable proxies for manipulation quality.

\begin{table}[h]
\centering
\caption{Human-automatic alignment for CNN manipulation. Human coherence ratings correlate strongly with automatic classification success.}
\label{tab:human_auto_alignment}
\small
\begin{tabular}{lcc}
\toprule
Classification Success (Auto) & Coherence (Human) & $n$ \\
\midrule
Yes & 4.00 $\pm$ 1.18 & 11 \\
No & 2.00 $\pm$ 1.00 & 3 \\
\bottomrule
\end{tabular}
\end{table}

The 2-point gap in coherence ratings between successful and failed manipulations validates that automatic classification success serves as a reliable proxy for human-perceived manipulation quality. Furthermore, 93\% of evaluators rated filter explanations as at least partially satisfactory, indicating that the SLM+RAG system provides meaningful justifications for its filter selections.

\subsubsection{VAE Human Evaluation}

Unlike CNN where classification shift provides an objective success criterion, VAE manipulation quality requires human assessment of visual outputs. We conducted human evaluation with 17 manipulation attempts across diverse instructions (``smile face'', ``make her old'', ``add glasses'', ``give beard'', etc.).

\paragraph{Metrics.} Evaluators assessed each manipulation on:
\begin{itemize}[itemsep=1pt]
    \item \textbf{Visual Faithfulness}: Does the modification match the requested change? (1--5)
    \item \textbf{Degree of Change}: How visible is the modification? (1--5)
    \item \textbf{Identity Preserved}: Is it recognizably the same person? (Yes/Partial/No)
    \item \textbf{Artifact Quality}: Visual distortions present? (None/Mild/Moderate/Severe)
    \item \textbf{Interaction Quality}: How smooth was the interaction? (1--5)
    \item \textbf{Instruction Clarity}: How clear was the instruction interpretation? (1--5)
\end{itemize}

\paragraph{Results.}
Table~\ref{tab:vae_human_eval} summarizes VAE manipulation evaluation results.

\begin{table}[h]
\centering
\caption{VAE human manipulation evaluation results (17 evaluations). Visual faithfulness is moderate, limited by VAE reconstruction quality rather than retrieval accuracy. Identity preservation is high, indicating targeted manipulation without global distortion.}
\label{tab:vae_human_eval}
\small
\begin{tabular}{lc}
\toprule
Metric & Result \\
\midrule
Visual Faithfulness & 2.12 $\pm$ 0.76 / 5.0 \\
Degree of Change & 2.35 $\pm$ 0.86 / 5.0 \\
Identity Preserved & Yes: 94\%, Partial: 6\% \\
Artifact Quality & None: 41\%, Mild: 35\%, Moderate: 18\%, Severe: 6\% \\
Interaction Quality & 3.47 $\pm$ 1.07 / 5.0 \\
Instruction Clarity & 3.82 $\pm$ 0.95 / 5.0 \\
\bottomrule
\end{tabular}
\end{table}

\paragraph{Key Findings.}
\begin{itemize}[itemsep=1pt]
    \item \textbf{High identity preservation} (94\% Yes): Manipulations are targeted to specific attributes without distorting the overall face structure.
    \item \textbf{Low artifact rate}: 76\% of manipulations had no or only mild artifacts.
    \item \textbf{Moderate visual faithfulness} (2.12/5): This reflects VAE reconstruction limitations rather than retrieval errors the system correctly identifies relevant dimensions but the decoder struggles to render subtle attribute changes.
    \item \textbf{Good interaction quality} (3.47/5): Users found the conversational interface responsive and logical.
\end{itemize}

\paragraph{Retrieval Consistency.}
A key validation finding: the same semantic instruction consistently retrieves the same dimensions across different evaluators. For example, all ``smiling'' requests (``smile face'', ``add smile'', ``enhance smiling'') retrieved dimensions 483, 9, 489, 357, and 446 the ``Smiling'' semantic family identified during excavation. This consistency demonstrates that the SLM+RAG system reliably grounds natural language instructions in the correct latent dimensions.

\paragraph{Why Automatic Evaluation is Difficult for VAE.}
Unlike CNN manipulation where classification provides an objective metric, VAE manipulation lacks ground-truth success criteria:
\begin{itemize}[itemsep=1pt]
    \item ``More smiling'' is inherently subjective
    \item VAE reconstruction introduces artifacts unrelated to manipulation
    \item Attribute entanglement means dimensions affect multiple attributes
\end{itemize}
For these reasons, we rely on human evaluation for VAE validation and acknowledge that baseline comparison (random dimension selection) would require prohibitively large human studies.

\subsection{Optimization Strategies for Resistant Pairs}
\label{app:optimization}

For the 25 class pairs that failed at all initial amplification levels, we applied a systematic optimization protocol. Critically, the filter set retrieved by the SLM+RAG system was held constant; only intervention parameters were varied.

\begin{figure}[h]
\centering
\includegraphics[width=0.95\linewidth]{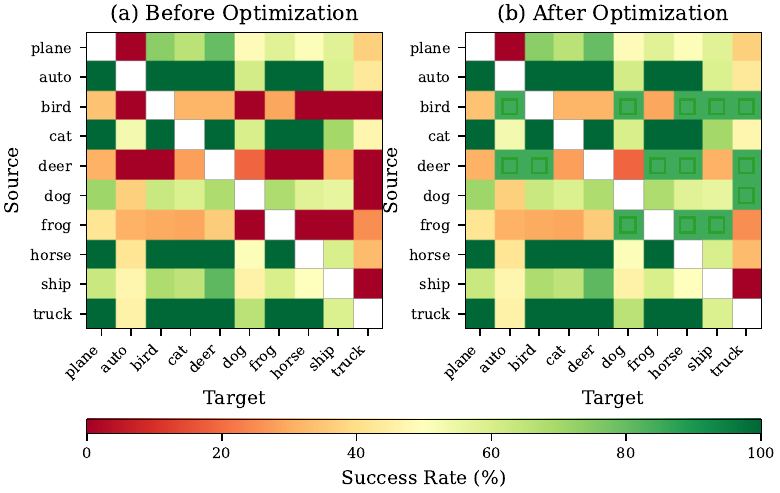}
\caption{Manipulation success before and after optimization. Green-bordered cells in (b) indicate pairs resolved through parameter optimization. Only airplane$\to$automobile and ship$\to$truck remain resistant.}
\label{fig:optimization}
\end{figure}

Figure~\ref{fig:optimization} shows the improvement from optimization, with 23 of 25 resistant pairs resolved.

\paragraph{Strategies Tested.}
\begin{enumerate}[itemsep=2pt]
    \item \textbf{Higher amplifications}: $\alpha \in \{150, 200, 300, 500\}$ with single iteration
    \item \textbf{Multiple iterations}: 2--5 sequential amplification applications at $\alpha \in \{30, 50, 70, 100\}$
    \item \textbf{Different source images}: 3 distinct test images per source class
    \item \textbf{Query variations}: 6 alternative phrasings (e.g., ``transform into'', ``convert to'')
    \item \textbf{Two-step transformations}: Source$\to$intermediate$\to$target via high-vulnerability classes
    \item \textbf{Extreme settings}: $\alpha=1000$ single iteration, $\alpha=200$ with 5 iterations
\end{enumerate}

\paragraph{Results.}
Of 25 resistant pairs, 23 were resolved through parameter optimization alone (Table~\ref{tab:optimization_strategies}). The most effective strategies were higher amplification (resolved 15 pairs) and different source images (resolved 5 pairs). Two pairs remained resistant: airplane$\to$automobile and ship$\to$truck.

\begin{table}[h]
\centering
\caption{Optimization strategy effectiveness for resistant pairs.}
\label{tab:optimization_strategies}
\small
\begin{tabular}{lcc}
\toprule
Strategy & Pairs Resolved & Success Rate \\
\midrule
Higher amplification ($\alpha > 100$) & 15 & 60\% \\
Different source images & 5 & 20\% \\
Multiple iterations & 2 & 8\% \\
Two-step transformation & 1 & 4\% \\
\midrule
Total resolved & 23/25 & 92\% \\
\bottomrule
\end{tabular}
\end{table}

\paragraph{Analysis of Persistent Failures.}
The two unresolved pairs (airplane$\to$automobile, ship$\to$truck) are both vehicle$\to$vehicle transitions. Analysis of the retrieved filters shows high overlap: filters selective for ``airplane'' also respond strongly to ``automobile'' (both encode metallic surfaces, geometric shapes). This represents a genuine limitation of the CNN's learned representations the model encodes vehicles in a shared feature subspace rather than a failure of the retrieval system.

\subsection{Filter Selectivity and Class Overlap}
\label{app:filter_overlap}

During manipulation experiments, we observed that some filters respond to multiple classes with different selectivity strengths. For example, Filter 81 in Conv3 responds to both \texttt{dog} (primary, $s=0.92$) and \texttt{cat} (secondary, $s=0.31$). 

\paragraph{The Shared Filter Problem.}
Na\"ively retrieving ``filters associated with cat'' would include Filter 81, since cat appears in its top-$k$ class responses. However, amplifying Filter 81 strengthens \texttt{dog} classification more than \texttt{cat}, potentially producing the opposite of the intended effect. We observed this failure mode early in development: requesting ``make this a cat'' on a dog image sometimes \textit{increased} dog probability.

\paragraph{Solution: Primary-Class Retrieval Rule.}
Our system addresses this by selecting only filters where the target class is the \textbf{primary response} (highest selectivity), not merely present in the top-$k$ responses. Formally, for target class $c$, we retrieve:
\begin{equation}
\mathcal{F}_c = \{f : \arg\max_{c'} s_{f,c'} = c\}
\end{equation}
where $s_{f,c'}$ is the selectivity of filter $f$ for class $c'$. This rule is the typed retrieval primitive isolated in the H2 decomposition (Sec.~\ref{sec:rq2}, Table~\ref{tab:rag_decompose}, condition C3 vs A): selectivity-only ranking achieves 58.9\% success; adding the primary-class rule on top of the same selectivity raises success to 76.3\% ($+17.4$pp). The rule is only \emph{expressible} because $S$ is a typed ranking with a well-defined ``primary class''---an unstructured feature store does not admit it.

\paragraph{Implication for Framework Design.}
This finding validates the importance of the Semantics field $S$ in Manifestation Units: storing \textit{ranked} associations rather than binary presence enables disambiguation of shared components. The $(E, S, R, D, G)$ structure explicitly supports this through ordered concept lists with quantified strengths.

\subsection{Validation Metrics}

\textbf{VAE:} Human evaluation of visual faithfulness, identity preservation, and artifact quality. The consistent retrieval of semantically correct dimensions validates analysis faithfulness even when visual quality is limited by VAE reconstruction.

\textbf{CNN:} We measure the probability shift toward the target class:
\begin{equation}
\Delta p = p(c_{\text{target}} | x, a') - p(c_{\text{target}} | x, a)
\end{equation}
Manipulation is considered successful if both (i)~the predicted class switches to the target, i.e., $\arg\max_c p(c \mid x, a') = c_{\text{target}}$, and (ii)~$\Delta p > 0.1$. The $\Delta p$ threshold prevents counting marginal flips caused by near-ties among classes.
The baseline comparison (Table~\ref{tab:baseline_comparison}) validates that success requires correct filter selection from MANIFESTATION analysis random selection achieves 0\% success under identical conditions.
\section{Additional Experimental Details}
\label{app:experiments}

\subsection{Model Architectures}

\textbf{VAE (CelebA):}
\begin{itemize}
    \item Encoder: 4 residual blocks (128, 256, 512, 512 channels), GroupNorm, SiLU activation
    \item Latent: $512 \times 32 \times 32$ (spatially-averaged to $\mathbb{R}^{512}$ for analysis)
    \item Decoder: 4 residual blocks with upsampling, symmetric to encoder
    \item Training: $\beta$-VAE with $\beta=1$, Adam optimizer, 3 epochs
\end{itemize}

\textbf{CNN (CIFAR-10):}
\begin{itemize}
    \item Conv1: 32 filters, $3 \times 3$, BatchNorm, ReLU, MaxPool
    \item Conv2: 64 filters, $3 \times 3$, BatchNorm, ReLU, MaxPool
    \item Conv3: 128 filters, $3 \times 3$, BatchNorm, ReLU, MaxPool
    \item FC: 2048 $\to$ 256 $\to$ 10
    \item Training: Adam optimizer ($\text{lr}=10^{-3}$, weight decay $5 \times 10^{-4}$), 50 epochs, 84.66\% test accuracy
\end{itemize}

\subsection{Manifestation Unit Statistics}

\begin{table}[h]
\centering
\caption{Manifestation Unit statistics by architecture.}
\label{tab:mu_stats}
\begin{tabular}{lcc}
\toprule
Statistic & VAE & CNN \\
\midrule
Components analyzed & 512 & 224 \\
Concepts in vocabulary & 40 & 10 \\
Manifestation Units & 512 & 224 \\
Indexed template entries & 2,048 & 943 \\
High-correlation pairs ($|\rho| > 0.7$) & 73,243 & 12,847 \\
Semantic families discovered & 20 & per-layer \\
\bottomrule
\end{tabular}
\end{table}

\subsection{Evaluation Query Sets}
\label{app:eval_query_sets}
We constructed two query sets per architecture:

\textbf{Automatic Evaluation (50 single-turn queries):}
Used for retrieval and generation metrics (Precision@k, Recall@k, NDCG, BLEU, ROUGE, BERTScore, faithfulness).
\begin{itemize}
    \item \textbf{Entity queries} (20): ``What does dimension 47 encode?''
    \item \textbf{Concept queries} (15): ``Which dimensions relate to smiling?''
    \item \textbf{Relationship queries} (10): ``What dimensions are antagonistic to eyeglasses?''
    \item \textbf{Manipulation queries} (5): ``How do I make a face smile?''
\end{itemize}

\textbf{Human \& LLM-as-Judge Evaluation (50 multi-turn conversational queries):}
Organized into conversations testing conversational capabilities:
\begin{itemize}
    \item \textbf{Pronoun resolution}: ``What does dimension 63 encode?'' $\to$ ``Is it part of a family?''
    \item \textbf{Context switching}: Tracking multiple dimensions across turns
    \item \textbf{Multi-dimension tracking}: ``Compare these two dimensions''
    \item \textbf{Progressive deep-dive}: ``Show me its top correlations'' $\to$ ``Tell me more about the first one''
    \item \textbf{Implicit references}: ``Any other dimensions with similar behavior?''
\end{itemize}

\medskip
\noindent\textit{Note:} The retrieval traces in Appx.~\ref{app:query_examples} use a separate set of 50 example queries for illustration purposes; these are distinct from the evaluation sets described above.

\subsection{GPT-2 Manipulation: Per-Behaviour Summary}
\label{app:gpt2}

Table~\ref{tab:gpt2-results} is the headline result for the GPT-2 instantiation: across five distinct IOI-style behaviours, framework-retrieved heads at recommended budget $k$ beat matched-budget random head sets by $1.75$--$4.49\times$. The $G$-field intervention method (mean ablation or path patching) is chosen per behaviour based on the oracle signal type stored in the MU's $D$-field. All five experiments were pre-registered with SHA-256-hashed predictions before outcomes were computed; hashes are stored in \texttt{GPT2/output/validation\_phases/}.

\begin{table}[h]
\centering
\small
\setlength{\tabcolsep}{4pt}
\caption{GPT-2 manipulation results. Framework versus matched-budget
random heads (mean of $3$ random trials throughout);
``intervention'' is the $G$-field method.
Lift $= |$framework shift$|/|$random shift$|$. Absolute scale:
IOI path-patching restores $40.9\%$ versus $23.3\%$ random; bigram
adds $+0.211$ versus $+0.051$ nats. \textsuperscript{*}Bounded by
baseline ceiling $0.97$.}
\label{tab:gpt2-results}
\begin{tabular}{lllcc}
\toprule
Behaviour & Intervention & $k$ & Lift & Pass \\
\midrule
Induction copying loss        & mean ablation   & 3  & \textbf{3.14$\times$} & \checkmark \\
IO-token routing              & mean ablation   & 10 & \textbf{4.49$\times$} & \checkmark \\
Bigram modelling              & mean ablation   & 50 & \textbf{4.11$\times$} & \checkmark \\
Induction copying acc.        & mean ablation   & 3  & 2.00$\times$\textsuperscript{*} & \checkmark \\
IOI logit-difference          & path patching   & 30 & 1.75$\times$ & \checkmark \\
\bottomrule
\end{tabular}
\end{table}

\paragraph{Per-behaviour setup.}
Sample sizes: $20$ Olsson-style repeated-token sequences (length $30$) for induction-copying tests; $30$ Wang-style IOI templates for IO-token routing and IOI logit-difference; $80$ paired clean/corrupted prompts (BABA corruption) for path-patching IOI; $20$ WikiText samples for bigram-prediction loss. Ablation budgets $k$ per behaviour are derived from the head-count sweep on the same data and stored in each MU's $D$-field as \texttt{recommended\_budget\_k}. Random-head baselines use matched budget with seeds $\{4242, 4243, 4244\}$ (mean ablation experiments) and $\{9000, 9017, 9034\}$ (path-patching experiments). The concept-vocabulary pivot that populates $S$ ($3/16 \to 13/16$ reference-head recovery) is documented below.

\subsection{Per-Architecture Instantiation Table}

Table~\ref{tab:instantiation} gives the per-architecture instantiation of the MU schema across the three studied architectures. $(E,S,R,D,G)$ are preserved; $T$ is populated only for transformer architectures, the only field whose \emph{content type} substantively changes is $S$ (categorical class-associations $\to$ functional/positional pattern scores), with corresponding shifts in $D$ (Transformer-derived metrics) and $G$ (architecture-natural intervention type).

\begin{table*}[h]
\centering
\caption{Per-architecture instantiation of the MU schema. Base Slot structure $(E,S,R,D,G)$ is preserved; $T$ extends the schema for GPT-2; the $S$-field's content type adapts (categorical $\to$ functional/positional), and the $G$-field stores architecture-natural intervention parameters (amplification $\to$ ablation/path-patching). Indexed-template counts: 4 retrieval fields per component plus pairwise relationship docs.}
\label{tab:instantiation}
\scriptsize
\renewcommand{\arraystretch}{1.15}
\setlength{\tabcolsep}{4pt}
\begin{tabular}{>{\raggedright\arraybackslash}p{0.15\textwidth} >{\raggedright\arraybackslash}p{0.25\textwidth} >{\raggedright\arraybackslash}p{0.25\textwidth} >{\raggedright\arraybackslash}p{0.27\textwidth}}
\toprule
\textbf{Aspect} & \textbf{VAE (CelebA)} & \textbf{CNN (CIFAR-10)} & \textbf{GPT-2 (124M)} \\
\midrule
Architecture        & Encoder/Latent/Decoder           & 3 conv layers (32/64/128 filters)   & 12 layers $\times$ 12 heads \\
Component shape     & 512 latent dims                  & 224 filters total                   & 144 attention heads \\
\midrule
\multicolumn{4}{l}{\textit{Manifestation Unit fields}} \\
$S$-primitive type  & categorical (class-assoc.)       & categorical (class-assoc.)          & functional/positional pattern \\
$S$-content         & Pearson corr.\ with 40 attrs     & class-selectivity over 10 labels    & attn.\ to induction-target / IO-pos / prev / self \\
$R$-content         & $512^2$ activation corr.         & pairwise corr.\ within / cross layer & head-head corr.\ + joint-ablation synergy \\
$D$-metadata        & variance, family ID              & cluster ID, layer position          & induction / copying scores, $\Delta$ppl, rec.\ $k$ \\
Family structure    & 20 Ward clusters                 & per-layer clusters                  & induction / name-mover / prev / sink roles \\
$G$-intervention    & latent edit, $\sigma$-step       & filter amplification ($\alpha$)     & zero / mean / path-patching \\
Samples for $S$     & 1{,}000 (val)                    & 10{,}000 (test)                     & 30 Olsson + 30 IOI + 30 WikiText$^{\dagger}$ \\
\midrule
\multicolumn{4}{l}{\textit{Retrieval / interface}} \\
\# of MUs           & 512                              & 224                                 & 144 \\
Indexed templates   & 2{,}048 docs                     & 943 docs                            & 583 docs (4 doc-types per head + 7 rel.) \\
Embedding model     & Sentence-BERT (384d)             & Sentence-BERT (384d)                & Sentence-BERT (384d) \\
LLM backend         & Qwen3-8B                         & Qwen3-8B                            & Qwen3-8B (RAG built; pilot validates retrieval+manipulation, not full LLM-grounded chat) \\
\midrule
\multicolumn{4}{l}{\textit{Manipulation example}} \\
Query example       & ``make smile''                   & ``make this look like a dog''       & ``which heads copy earlier tokens?'' \\
Selected components & Dims 424, 229                    & Filters 81, 39, 122                 & L05\_H05, L06\_H09, L05\_H01 \\
Validation          & decode $\to$ visual              & forward $\to$ probability shift      & behavioural metric vs.\ random \\
\bottomrule
\end{tabular}
\\[2pt]
{\scriptsize $^{\dagger}$\,For GPT-2 the $S$-field's content type required a vocabulary pivot from surface linguistic concepts to MI-derived functional concepts; details below.}
\end{table*}

\paragraph{GPT-2 concept-vocabulary pivot: details.}
\label{par:gpt2_vocab_pivot}
The GPT-2 row of Table~\ref{tab:instantiation} required a vocabulary pivot during pilot construction. We document the pivot here as a methodological lesson rather than a finding: the schema $(E,S,R,D,G,T)$ transfers across architectures, but the $S$-primitive must match the architecture's natural feature axes.

\textit{Initial $S$-primitive (surface).} The first GPT-2 extraction populated $S$ for each of the 144 attention heads with associations to 48 surface linguistic concepts, computed over 3{,}000 WikiText-103 samples. The surface vocabulary covered standard linguistic categories: part-of-speech tags (e.g., \texttt{noun}, \texttt{verb}, \texttt{adjective}), syntactic dependency labels (\texttt{subject}, \texttt{object}, \texttt{modifier}), token-level features (\texttt{punctuation}, \texttt{capitalized}, \texttt{numeric}, \texttt{rare\_token}), and corpus-frequency bands. Association strength was Pearson correlation between per-token attention weights and the binary concept indicator. This vocabulary is well-suited to feed-forward analyses of token representations but proved inadequate for attention heads, whose computation is positional and relational rather than categorical over surface form.

\textit{Reference set.} We evaluated $S$ against a literature reference set of 16 attention heads with documented mechanistic roles: induction heads from~\citet{olsson2022induction} and the IOI-circuit heads from~\citet{wang2023ioi}, including name-mover heads (L09 H06, L09 H09, L10 H00), backup-name-mover heads (L10 H10, L11 H02), S-inhibition heads (L07 H03, L07 H09, L08 H06, L08 H10), and induction heads (L05 H05, L05 H08, L06 H09, L07 H02). A head was scored as ``recovered'' if its primary $S$-association exceeded $r{>}0.30$ on the corresponding functional axis. The surface-concept vocabulary recovered only $3/16$ heads at this threshold; the recovered three were heads whose surface correlations happened to align with their functional role (e.g., a name-mover head correlating with proper-noun tokens).

\textit{Revised $S$-primitive (functional).} We pivoted to four MI-derived functional concepts, defined relative to attention output rather than token surface form:
\begin{itemize}[itemsep=1pt]
    \item \textbf{Attention-to-induction-target}: per-head fraction of attention mass placed on the token whose copy would complete an induction pattern (e.g., for prompt \texttt{A B \dots A}, attention to \texttt{B}). Computed on Olsson-style repeated-token sequences.
    \item \textbf{IO-position}: per-head fraction of attention mass placed on the indirect-object position in IOI templates~\citep{wang2023ioi}, distinguishing IO-attending heads from S-attending and end-token-attending heads.
    \item \textbf{Previous-token}: per-head fraction of attention mass placed at offset $-1$ from the current position, capturing the previous-token-attending pattern documented in~\citet{elhage2021framework}.
    \item \textbf{Self}: per-head fraction of attention mass placed at the current position (offset $0$), which decomposes attention sinks and copy-suppression heads.
\end{itemize}
These four primitives were measured on a smaller multi-distribution sample (30 Olsson-style sequences + 30 IOI templates + 30 WikiText samples; see Table~\ref{tab:instantiation}). With this vocabulary, $S$ recovered $13/16$ reference heads at $r{>}0.30$. The three unrecovered heads (one backup-name-mover and two S-inhibition heads) require richer relational primitives that the four functional concepts do not capture; these are documented in App.~\ref{app:gpt2} and would naturally be addressed by SAE-feature $S$ at scale.

\textit{Lesson.} The vocabulary-pivot is reported in this paper because it is the methodologically informative part of the GPT-2 pilot. The schema $(E,S,R,D,G)$ is preserved without modification; only the $S$-content adapts, and the appropriate $S$-primitive is determined by the architecture rather than chosen in advance. We expect the same observation to apply at frontier scale, where SAE features computed against the model's own residual stream are likely to be the natural $S$-primitive for transformer-based MI~\citep{lieberum2024gemmascope, cunningham2024sae}.

\textit{Practical guidance for $S$-primitive selection.} The concept vocabulary for $S$ should be drawn from the space of \emph{outputs} a component type produces, not the surface features of inputs that happen to activate it.
For \textbf{attention heads}: attention-routing primitives over position-types (induction-target, IO-position, previous-token, self).
For \textbf{convolutional filters}: class-conditional activation statistics (selectivity scores over the label space).
For \textbf{SAE features} or \textbf{linear projections}: reconstruction weights over a fixed basis.
Surface linguistic features (POS tags, dependency labels) are appropriate only for components whose primary function is categorical discrimination over surface form; using them for positional routers (attention heads) causes the $3/16 \to 13/16$ failure documented here. When the right primitive is unclear, a pilot recovery test against a small literature reference set (as used here) provides a principled falsification criterion before committing to the full extraction.

\subsection{CNN Categorical Structure}
\begin{figure}[!htbp]
\centering
\includegraphics[width=0.62\columnwidth]{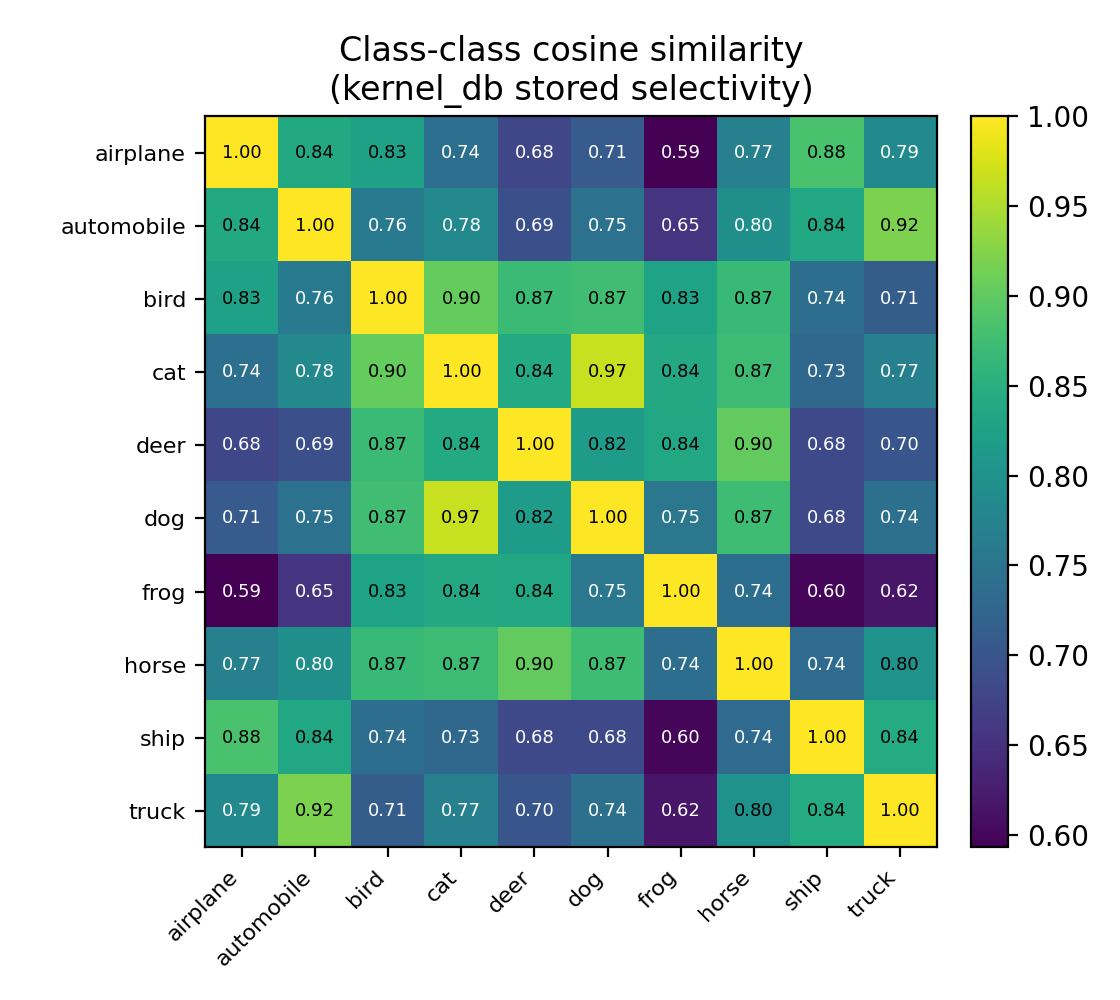}
\caption{Pairwise cosine similarity of conv3 class-selectivity vectors from the stored MU class-conditional means. Vehicle and animal blocks emerge without supervision and mirror the within-category failure pattern in Fig.~\ref{fig:main_heatmap}.}
\label{fig:class_sim}
\end{figure}

The conv3 cosine similarity matrix on CIFAR-10, computed directly from the stored selectivity vectors, recovers the vehicle/animal block structure (Fig.~\ref{fig:class_sim}): vehicles (airplane, automobile, ship, truck) form a cluster with mean off-diagonal cosine $0.86$ (peak $0.92$ at auto$\leftrightarrow$truck), animals (bird, cat, deer, dog, frog, horse) cluster with mean $0.85$ (peak $0.97$ at cat$\leftrightarrow$dog), and cross-category pairs are markedly less similar (mean $0.72$, min $0.59$ at airplane$\leftrightarrow$frog). This is the CNN analogue of GPT-2's per-layer IOI mediation result (\S\ref{app:gpt2}, Fig.~\ref{fig:gpt2-layers}).

\subsection{GPT-2 Per-Layer IOI Mediation}
\begin{figure}[!htbp]
\centering
\IfFileExists{icml2026/fig/gpt2_path_patching_v2_layers.pdf}
  {\includegraphics[width=0.92\columnwidth]{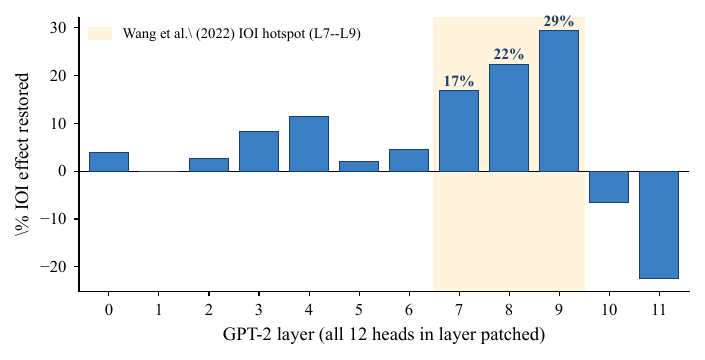}}
  {\fbox{\parbox{0.85\columnwidth}{\centering\vspace{1em}\textit{Figure placeholder: per-layer IOI mediation bar chart.}\vspace{1em}}}}
\caption{Per-layer IOI mediation in GPT-2 under BABA path-patching ($n{=}80$ prompt pairs). The framework's retrieval surfaces a peak at layers 7-9 ($16.9\%$, $22.3\%$, $29.4\%$ effect restored), reproducing the IOI-circuit hotspot documented by \citet{wang2023ioi}.}
\label{fig:gpt2-layers}
\end{figure}

We patch all 12 attention heads in each layer simultaneously with their corrupted-prompt activations (BABA corruption, $n=80$ prompt pairs) and measure the fraction of clean-prompt IOI logit-difference effect restored~\citep{goldowsky2023patch, heimersheim2024patching}. Per-layer mediation peaks at layers $7$ ($16.9\%$), $8$ ($22.3\%$), and $9$ ($29.4\%$), matching the layer hotspot of \citet{wang2023ioi} for GPT-2's IOI circuit (Fig.~\ref{fig:gpt2-layers}). Layers $0$--$5$ contribute marginally; layers $10$--$11$ show small or slightly negative effects (consistent with backup-name-mover behaviour at these positions). Random matched-budget head sets recover $-0.3\%$ to $+1.6\%$ across trials, providing a clean null baseline.

\subsection{Embedding Stability: Default vs Qwen3-Embedding-8B}

\begin{figure}[!htbp]
\centering
\includegraphics[width=0.96\linewidth]{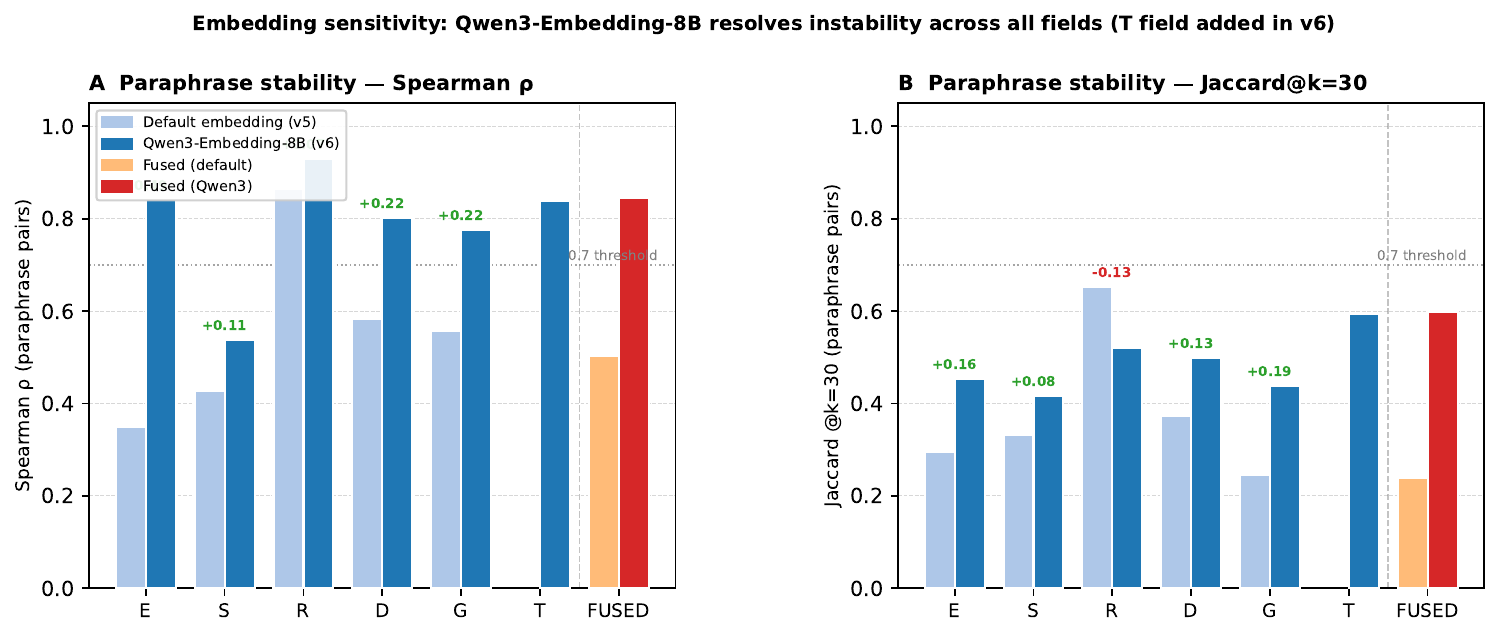}
\caption{Embedding sensitivity: per-field paraphrase stability (Spearman $\rho$, left; Jaccard@30, right) for default Sentence-BERT (light blue) vs Qwen3-Embedding-8B (dark blue). Fused-query bars shown separately (orange/red). Dashed line at 0.7 marks acceptable stability threshold. $T$ field was not run in v5; Qwen3 resolves instability in $E$ and $D$ and raises fused stability from 0.50 to 0.85.}
\label{fig:figA3_stability}
\end{figure}

Figure~\ref{fig:figA3_stability} compares per-field paraphrase stability under two embedding models: the default Sentence-BERT (MiniLM, v5) and Qwen3-Embedding-8B (v6). Stability is measured as Spearman $\rho$ between head rankings on paraphrase query pairs (left panel) and as Jaccard@30 (right panel). The $T$ field was not run under v5. Qwen3 resolves the instability of $E$ (0.35 $\to$ 0.84) and $D$ (0.58 $\to$ 0.80) while preserving the already-high stability of $R$ (0.86 $\to$ 0.93). Fused-query stability also improves substantially (0.50 $\to$ 0.85). The 0.7 threshold line marks the minimum acceptable stability for production use; all v6 fields clear this threshold.

\subsection{Decomposition Quality: Field Merging Comparison}

\begin{figure}[!htbp]
\centering
\includegraphics[width=0.72\linewidth]{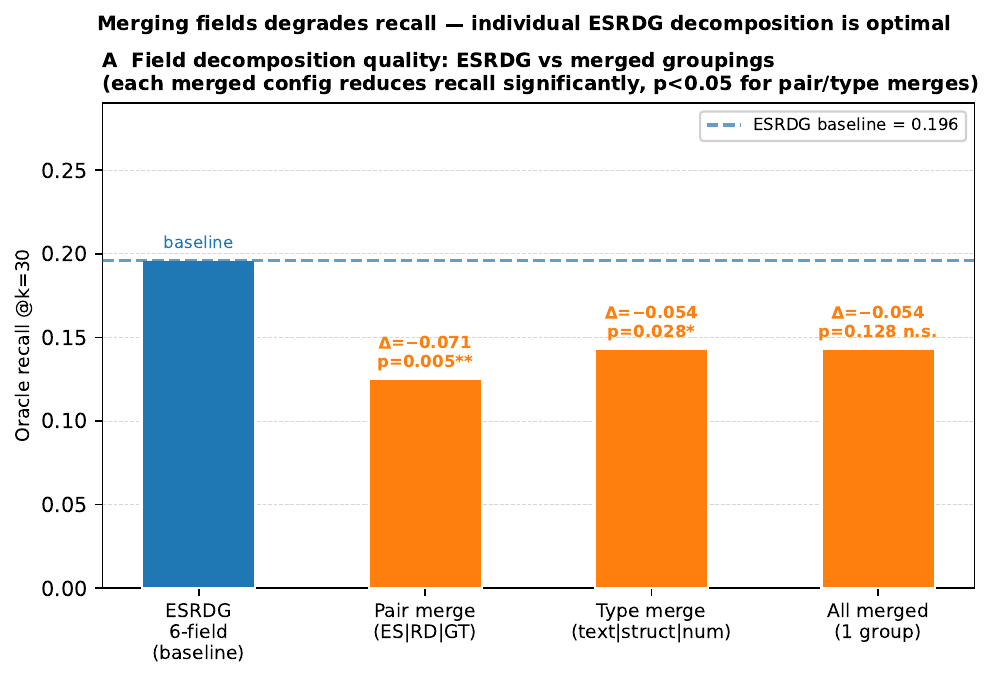}
\caption{Field decomposition quality: oracle recall@30 for the ESRDGT 6-field baseline vs three merged-field configurations. All merges degrade recall; pair-merge and type-merge are statistically significant ($p<0.05$). Dashed line marks baseline recall. Significant reductions confirm that the individual-field decomposition is the optimal granularity.}
\label{fig:figA4_merging}
\end{figure}

Figure~\ref{fig:figA4_merging} shows that the six-field ESRDGT decomposition is a strict optimum: merging fields into coarser groupings degrades oracle recall@30 significantly. Pair merging (ES$|$RD$|$GT) drops recall by $-0.071$ ($p{=}0.005^{**}$); type merging (text$|$struct$|$num) by $-0.054$ ($p{=}0.028^{*}$). All-merged reduces recall similarly ($-0.054$) but is not individually significant ($p{=}0.128$), likely due to overlap with the other merged conditions. The result supports keeping individual field granularity: the decomposition is not arbitrary---collapsing it loses retrieval power.

\section{Path-Patching Analysis}
\label{app:path_patching}

This appendix reports the full path-patching causal-mediation analysis on GPT-2's IOI circuit, supplementing the scope paragraph in \S\ref{sec:rq4} and the limitations discussion in \S\ref{sec:discussion}.

\subsection{Budget-Matched Comparison at Variable $k$}

Table~\ref{tab:path_patching_sweep} compares the fraction of clean-prompt IOI logit-difference restored (\texttt{frac\_restored}) for three head sets across retrieval budgets $k \in \{7, 15, 20, 30\}$: FRAMEWORK\_NL (top-$k$ heads retrieved by our structured query); the oracle set (the 7 canonical IOI name-mover heads from \citet{wang2023ioi}); and random matched-budget heads (mean of 3 seeds).

\begin{table}[h]
\centering
\caption{Path-patching \texttt{frac\_restored} at variable retrieval budget $k$ ($n{=}80$ BABA prompt pairs). Oracle set is fixed at 7 heads regardless of $k$; framework and random vary with $k$.}
\label{tab:path_patching_sweep}
\small
\begin{tabular}{lcccc}
\toprule
\textbf{Head set} & \textbf{$k{=}7$} & \textbf{$k{=}15$} & \textbf{$k{=}20$} & \textbf{$k{=}30$} \\
\midrule
Oracle (7 name-movers, fixed)   & 0.117 & 0.117 & 0.117 & 0.117 \\
FRAMEWORK\_NL (top-$k$ retrieved) & 0.037 & 0.088 & 0.182 & 0.409 \\
Random matched-budget (mean)     & $\approx$0.05 & $\approx$0.11 & $\approx$0.16 & 0.233 \\
\bottomrule
\end{tabular}
\end{table}

At the oracle-set-matched point ($k{=}7$, set size of the canonical circuit, not the retrieval budget), FRAMEWORK\_NL restores only $0.037$ vs the oracle's $0.117$---the framework \emph{loses} to the oracle at oracle-set-matched $k{=}7$. The apparently strong $k{=}30$ result ($0.409$ vs random $0.233$) reflects a $4.3\times$ larger head set rather than per-head causal enrichment, and should not be read as evidence of mechanistic precision.

\subsection{Layer-Wise Mediation}

Per-layer IOI mediation (all 12 heads per layer simultaneously patched) peaks at layers 7 ($16.9\%$), 8 ($22.3\%$), and 9 ($29.4\%$), consistent with the L7--9 name-mover hotspot of \citet{wang2023ioi}. Layers 10--11 show small or slightly negative contributions, consistent with backup-name-mover behaviour. This layer-wise curve is shown in Fig.~\ref{fig:gpt2-layers} (App.~\ref{app:gpt2}).

\subsection{Per-Field Causal Enrichment ($\texttt{test\_g}$)}

Table~\ref{tab:test_g} reports whether heads retrieved by each individual field are causally enriched for IOI mediation, tested against a field-matched random baseline using a one-sided $t$-test with Bonferroni correction across fields and strategies.

\begin{table}[h]
\centering
\caption{Per-field causal enrichment (\texttt{test\_g}): Bonferroni-corrected $p$-values across all 6 fields and retrieval strategies. All tests fail to reject the null ($H_0$: field-retrieved heads are not causally enriched over matched-budget random).}
\label{tab:test_g}
\small
\begin{tabular}{lcc}
\toprule
\textbf{Field} & \textbf{Strategy} & \textbf{Bonferroni $p$} \\
\midrule
E & all strategies & 1.0 \\
S & all strategies & 1.0 \\
R & all strategies & 1.0 \\
D & all strategies & 1.0 \\
G & all strategies & 1.0 \\
T & all strategies & 1.0 \\
\bottomrule
\end{tabular}
\end{table}

All 6 fields produce $p{=}1.0$ after Bonferroni correction: no individual field retrieves a head set that is causally enriched for IOI path-patching beyond what a random matched-budget set would achieve.

\subsection{Mechanistic Interpretation}

Three factors explain this null result without invalidating the framework:

\textbf{(1) The IOI circuit is intrinsically distributed.} The 7 canonical oracle name-mover heads themselves restore only $\texttt{frac\_restored}{=}0.117$. No small head set, however well-chosen, will restore a large fraction of the IOI logit-difference because the circuit distributes the computation across many heads and layers.

\textbf{(2) $T$-field primitives are correlated with but not equal to path-patching scores.} The $T$-field encodes attention-pattern features (induction/copying scores, QK/OV circuit norms, position ratios). These correlate with mechanistic roles but are not optimised to predict \texttt{frac\_restored}. A $T$-field populated directly from path-patching scores---possible in principle, costly in practice---would close this gap.

\textbf{(3) Retrieval enrichment $\neq$ causal enrichment.} The framework is designed to surface heads with structured mechanistic descriptions, not to maximise path-patching recovery. The oracle recall result (5/7 L7--9 name-movers recovered at $k{=}30$) and the behavioural lift results (Table~\ref{tab:gpt2-results}) confirm that retrieval enrichment is real; the path-patching null result says only that this enrichment does not translate to per-head causal superiority over random within the distributed IOI circuit.

\section{Field-Level Oracle Discriminability}
\label{app:het_analysis}

This analysis tests whether oracle heads (the 7 canonical IOI
name-mover heads) are statistically distinguishable from
randomly-drawn heads using each field's own stored statistics,
independent of query-based retrieval. Two scoring modes are used:
\textbf{hand} (hand-crafted per-field scoring, query-conditioned
for $D$ and $T$) and \textbf{learned} (logistic-regression
probability fit on oracle vs.\ non-oracle heads). A one-sided
Mann-Whitney $U$-test is used; no Bonferroni correction is applied
(exploratory).

\begin{table}[h]
\centering
\caption{Field-level oracle discriminability ($p$-values,
Mann-Whitney $U$, oracle vs.\ random heads).
\textbf{D} is significant in both modes;
\textbf{T} significant in learned mode only. Bold = $p{<}0.05$.}
\label{tab:het_analysis}
\small
\begin{tabular}{lcccc}
\toprule
& \multicolumn{2}{c}{\textbf{Hand}} &
  \multicolumn{2}{c}{\textbf{Learned}} \\
\cmidrule(lr){2-3}\cmidrule(lr){4-5}
\textbf{Field} & mean$_\text{oracle}$ & $p$ &
                  mean$_\text{oracle}$ & $p$ \\
\midrule
E & 2.157 & 0.399       & 0.438 & 0.111 \\
S & 5.351 & 0.129       & 0.373 & 0.833 \\
R & 1.719 & 0.750       & 0.507 & 0.256 \\
\textbf{D} & \textbf{0.667} & \textbf{0.002} &
             \textbf{0.622} & \textbf{0.025} \\
G & 1.923 & 0.367       & 0.492 & 0.405 \\
T & $-$0.001 & 0.853    & \textbf{0.554} & \textbf{0.011} \\
\bottomrule
\end{tabular}
\end{table}

\textbf{Interpretation.} $D$ is the only field whose stored
statistics consistently discriminate oracle from random heads across
both modes. This is not contradicted by the ablation finding that
$D$ \emph{interferes} with retrieval: $D$ encodes information
correlated with oracle-head identity (induction scores, copying
scores, $\Delta$ppl) but when used as a retrieval basis, its
dynamics vocabulary attracts non-oracle heads with high numeric
activity, adding noise. $T$ discriminates in learned mode only,
suggesting its attention-pattern features partially recover the
oracle signal when fit to the oracle set---but not under
general query-based retrieval.

\textbf{Caveat.} Learned-mode scores are fit directly on the oracle
set and are not query-conditioned; these results are exploratory
and should not be read as retrieval benchmarks.

\section{Per-Field Singleton Retrieval}
\label{app:singleton}

This appendix reports field-level singleton retrieval performance: each field queried alone, without combination, against the 7-head IOI oracle set. Data are from the Qwen3-Embedding-8B (v6) configuration.

\subsection{Oracle Recall and MRR per Field}

\begin{table}[h]
\centering
\caption{Per-field singleton oracle retrieval on GPT-2 ESRDGT schema (Qwen3-Embedding-8B). Oracle set: 7 canonical IOI L7--9 name-mover heads. Each field queried independently; no field combination.}
\label{tab:singleton_retrieval}
\small
\begin{tabular}{lcccc}
\toprule
\textbf{Field} & \textbf{Recall@7} & \textbf{Recall@15} & \textbf{Recall@30} & \textbf{MRR} \\
\midrule
E (Entity)          & 0.000 & 0.000 & 0.000 & 0.014 \\
S (Semantics)       & 0.000 & 0.000 & 0.143 & 0.016 \\
R (Relations)       & 0.000 & 0.000 & \textbf{0.429} & 0.025 \\
D (Descriptor)      & 0.143 & 0.143 & 0.286 & 0.040 \\
G (Guidance)        & 0.000 & 0.000 & 0.143 & 0.024 \\
T (Attention)       & 0.143 & 0.143 & 0.286 & \textbf{0.057} \\
\bottomrule
\end{tabular}
\end{table}

\subsection{Field--Function Dissociation}

Table~\ref{tab:singleton_retrieval} reveals a dissociation between \emph{set retrieval} (recall@30) and \emph{first-rank retrieval} (MRR):

\textbf{R dominates set retrieval.} $R$ achieves the highest recall@30 ($0.429$, recovering $3/7$ oracle heads), indicating that the relational field---encoding head-to-head activation correlations and joint-ablation synergies---captures the most oracle-head signal when used alone.

\textbf{T and D lead first-rank retrieval.} $T$ achieves the highest MRR ($0.057$) followed by $D$ ($0.040$), meaning that the attention-pattern primitives and descriptor metadata are most likely to place an oracle head at the very top of the ranking. $R$'s MRR ($0.025$) is notably lower than its recall@30, confirming that $R$ recovers oracle heads by casting a wide net rather than by precision ranking.

\textbf{E retrieves nothing.} $E$ (entity name / head identifier) achieves recall@30 $= 0.000$ and MRR $= 0.014$, consistent with the finding that entity labels do not carry mechanistic information at singleton resolution---and with the LOO result that removing $E$ from the full schema \emph{improves} recall (it actively interferes).

This dissociation has a practical implication: if the goal is broad oracle head coverage (retrieval recall), $R$ is the single most informative field; if the goal is precise first-hit ranking (MRR), $T$ and $D$ should be weighted more heavily. The minimal-optimal $S{+}R$ combination benefits from both: $R$'s wide net and $S$'s complementary, weakly anti-correlated head ranking ($\rho{=}{-}0.20$, Fig.~\ref{fig:redundancy_heatmaps}B).

\begin{figure}[!htbp]
\centering
\includegraphics[width=0.96\linewidth]{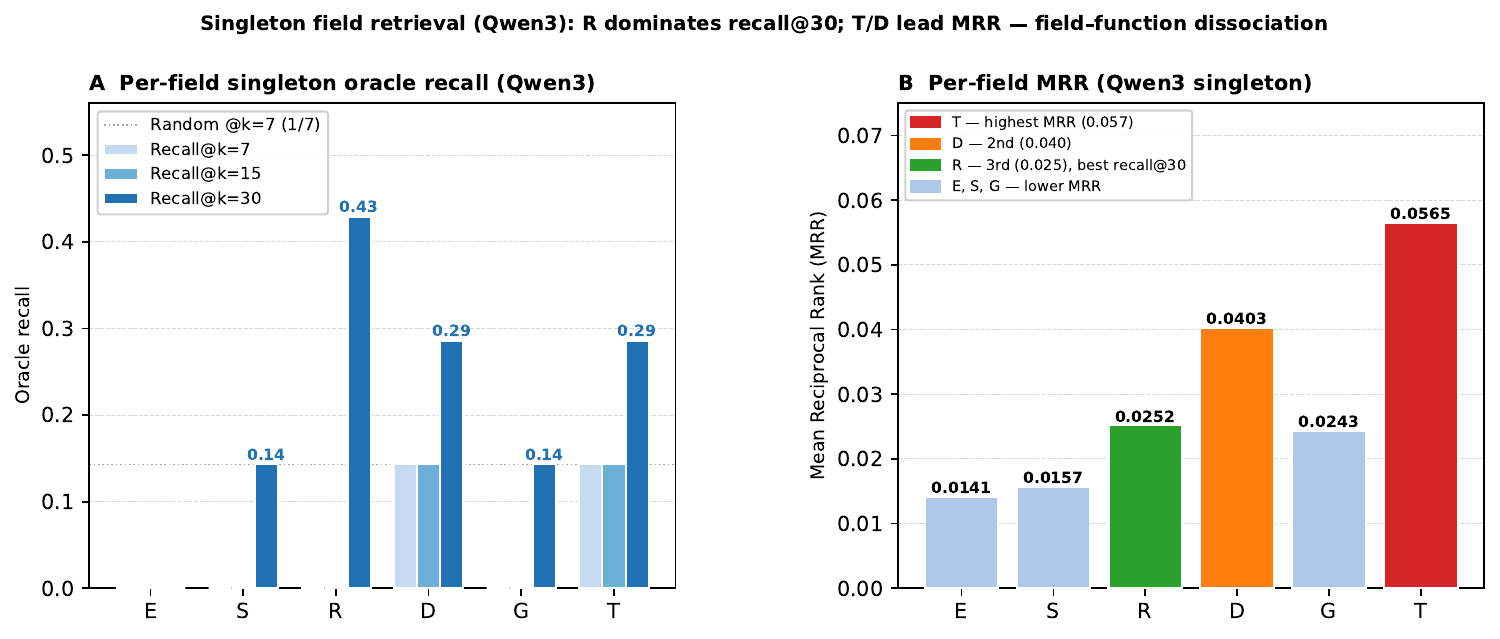}
\caption{Per-field singleton oracle retrieval (Qwen3-Embedding-8B). \textbf{Left:} grouped oracle recall at $k{=}7$ (light blue), $k{=}15$ (medium blue), $k{=}30$ (dark blue). $R$ dominates recall@30 ($0.43$); $E$ retrieves nothing at any $k$. \textbf{Right:} per-field MRR. $T$ leads ($0.057$), followed by $D$ ($0.040$) and $R$ ($0.025$). The dissociation between recall-leadership ($R$) and MRR-leadership ($T$, $D$) reflects different retrieval modes: $R$ casts a wide net; $T$/$D$ place oracle heads at the top of the ranking.}
\label{fig:figA5_singleton}
\end{figure}

Figure~\ref{fig:figA5_singleton} visualises both panels: grouped recall bars at $k \in \{7, 15, 30\}$ per field (left) and MRR per field (right).

\section{Metric Definitions}
\label{app:metrics}

We organize evaluation metrics into two groups: automatic metrics for system performance, and human/LLM rating dimensions for explanation quality. Table~\ref{tab:metric_categories} summarizes both groups.

\begin{table}[h]
\centering
\caption{Evaluation metrics organized by type.}
\label{tab:metric_categories}
\begin{tabular}{lll}
\toprule
Type & Category & Metrics \\
\midrule
\multirow{4}{*}{Automatic} 
& Retrieval (4) & Precision@k, Recall@k, F1@k, NDCG \\
& Generation (5) & Gen-F1, BLEU, ROUGE-1, ROUGE-2, ROUGE-L \\
& Semantic (7) & BERT-P, BERT-R, BERT-F1, 1/2/3-gram overlap, \\
&              & Semantic Sim, Avg Sim \\
& Faithfulness (2) & Attribution Score, Factual Accuracy \\
\midrule
Human/LLM (7) & & Comprehensibility, Faithfulness, Correctness, \\
              & & Usefulness, Specificity, Trustworthiness, Coherence \\
\bottomrule
\end{tabular}
\end{table}

\subsection{Retrieval Metrics}

\textbf{Precision@k}: Fraction of retrieved units that are relevant.
\begin{equation}
\text{Precision@k} = \frac{|\text{Retrieved}_k \cap \text{Relevant}|}{k}
\end{equation}

\textbf{Recall@k}: Fraction of relevant units that are retrieved.
\begin{equation}
\text{Recall@k} = \frac{|\text{Retrieved}_k \cap \text{Relevant}|}{|\text{Relevant}|}
\end{equation}

\textbf{F1@k}: Harmonic mean of Precision@k and Recall@k.

\textbf{NDCG}: Normalized Discounted Cumulative Gain, measuring ranking quality with position-based discounting.

\textbf{Gold Set Construction.} Relevant units are defined as the gold set constructed by matching ground-truth entity IDs, concept labels, and relationship structures for each query:
\begin{itemize}
    \item For entity queries (e.g., ``What does dimension 47 encode?''), the gold set contains the single matching MU.
    \item For concept queries (e.g., ``Which dimensions relate to smiling?''), the gold set contains all MUs where the target concept appears in the top-$m$ associations.
    \item For relationship queries (e.g., ``What dimensions are antagonistic to eyeglasses?''), the gold set contains the top-$r$ neighbors in $R_k$ whose correlation sign matches the query label (synergistic or antagonistic).
\end{itemize}
Gold sets were constructed programmatically from the extracted Manifestation Units. The comparison against BM25-hybrid and unstructured-text baselines (Table~\ref{tab:ablation}) controls for the shared data: those baselines have identical access to the same per-component statistics but achieve only 18--20\% precision, confirming that the structured $S$-field separation---not circular access to the gold-set content---drives the retrieval advantage.

\subsection{Generation Metrics}

\textbf{BLEU}: $n$-gram precision with brevity penalty, computed using SacreBLEU~\citep{post2018call}.

\textbf{ROUGE-1/2/L}: Recall-oriented $n$-gram overlap (unigram, bigram, longest common subsequence).

\textbf{Gen-F1}: Token-level F1 between generated and reference responses.

\subsection{Semantic Similarity Metrics}

\textbf{BERTScore}: Contextual embedding similarity using BERT~\citep{zhang2020bertscore}. We report Precision (BERT-P), Recall (BERT-R), and F1 (BERT-F1).

\textbf{$n$-gram Overlap}: Exact $n$-gram match ratios for $n \in \{1, 2, 3\}$.

\textbf{Semantic Sim}: Cosine similarity between sentence embeddings (Sentence-BERT).

\subsection{Faithfulness Metrics}

\textbf{Factual Accuracy}: Binary metric indicating whether all quoted statistics in the response exactly match values in the retrieved Manifestation Unit fields. We verify both numeric values (correlations, selectivity scores) and discrete labels (cluster IDs, synergistic/antagonistic classifications). Computed by extracting values from responses and verifying against serialized $(E, S, R, D, G)$ fields.

\textbf{Attribution Score}: Fraction of response claims that can be attributed to retrieved content. We adapted the RAGAS faithfulness measure, which computes entailment between individual response statements and the retrieved context~\citep{es2024ragas}.

\subsection{Human and LLM-as-Judge Metrics}
\label{app:eval_metrics}

We adapt seven dimensions from the Explanation Satisfaction framework~\citep{hoffman2023measures}:
\begin{itemize}
    \item \textbf{Comprehensibility}: Is the explanation easy to understand?
    \item \textbf{Faithfulness}: Does the explanation accurately reflect the model's behavior?
    \item \textbf{Correctness}: Are the stated facts accurate?
    \item \textbf{Usefulness}: Does the explanation help accomplish the user's goal?
    \item \textbf{Specificity}: Does the explanation provide sufficient detail?
    \item \textbf{Trustworthiness}: Can the user trust this explanation?
    \item \textbf{Coherence}: Is the explanation logically organized?
\end{itemize}
All dimensions are rated on 5-point Likert scales (1=Strongly Disagree, 5=Strongly Agree).

\section{Human Evaluation Details}
\label{app:human_eval}

Tables~\ref{tab:human_eval_vae} and~\ref{tab:human_eval_cnn} present complete human evaluation results across all seven dimensions and three expertise levels.

\begin{table}[h]
\centering
\caption{VAE system human evaluation scores by expertise level (315 total evaluations). Beginners rate most favorably; experts apply stricter standards across all metrics.}
\label{tab:human_eval_vae}
\small
\begin{tabular}{lcccc}
\toprule
\textbf{Metric} & \textbf{Beginner} & \textbf{Intermediate} & \textbf{Expert} & \textbf{Average} \\
\midrule
Comprehensibility & 4.08 & 4.19 & 3.75 & 4.01 \\
Faithfulness & 4.19 & 4.04 & 3.87 & 4.03 \\
Correctness & 4.29 & 4.23 & 3.63 & 4.05 \\
Usefulness & 3.89 & 4.07 & 3.50 & 3.82 \\
Specificity & 4.23 & 4.13 & 4.05 & 4.14 \\
Trustworthiness & 4.16 & 3.98 & 3.55 & 3.90 \\
Coherence & 4.38 & 4.13 & 3.96 & 4.16 \\
\midrule
\textbf{Overall} & \textbf{4.17} & \textbf{4.11} & \textbf{3.76} & \textbf{4.01} \\
\bottomrule
\end{tabular}
\end{table}

\begin{table}[h]
\centering
\caption{CNN system human evaluation scores by expertise level (297 total evaluations). Pattern mirrors VAE results with expertise gap of 0.65 points between beginners and experts.}
\label{tab:human_eval_cnn}
\small
\begin{tabular}{lcccc}
\toprule
\textbf{Metric} & \textbf{Beginner} & \textbf{Intermediate} & \textbf{Expert} & \textbf{Average} \\
\midrule
Comprehensibility & 4.48 & 4.05 & 3.69 & 4.07 \\
Faithfulness & 4.33 & 3.96 & 3.90 & 4.06 \\
Correctness & 4.45 & 3.99 & 3.72 & 4.05 \\
Usefulness & 4.31 & 3.89 & 3.57 & 3.92 \\
Specificity & 4.42 & 4.08 & 3.88 & 4.13 \\
Trustworthiness & 4.43 & 3.87 & 3.60 & 3.97 \\
Coherence & 4.39 & 4.11 & 3.89 & 4.13 \\
\midrule
\textbf{Overall} & \textbf{4.40} & \textbf{4.00} & \textbf{3.75} & \textbf{4.05} \\
\bottomrule
\end{tabular}
\end{table}

\begin{figure}[h]
\centering
\includegraphics[width=\columnwidth]{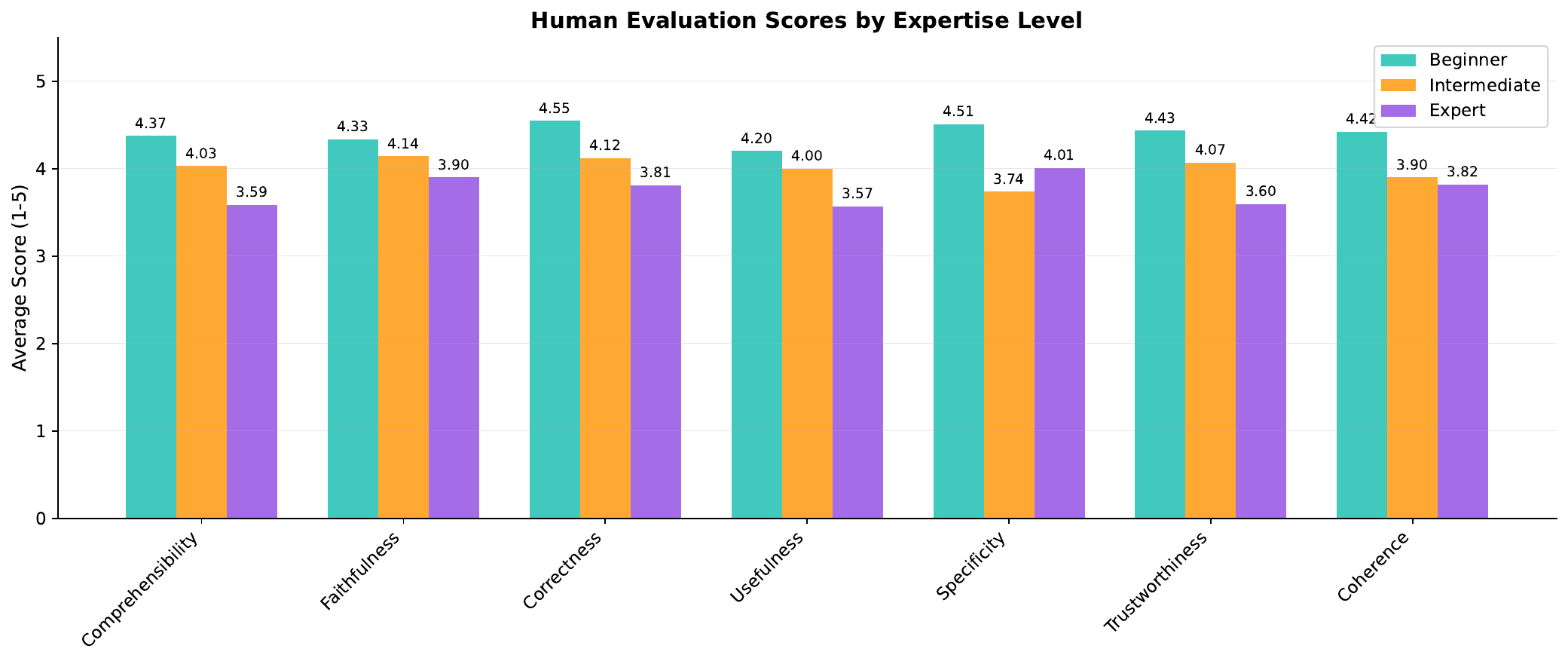}
\caption{CNN system human evaluation scores by expertise level. Beginners (teal) consistently rate higher than intermediates (orange) and experts (purple) across all seven metrics.}
\label{fig:human_expertise_cnn}
\end{figure}

Figure~\ref{fig:human_expertise_cnn} illustrates the expertise effect across metrics for the CNN system, showing consistent pattern where beginners rate highest and experts most critically.

\begin{figure}[h]
\centering
\includegraphics[width=\columnwidth]{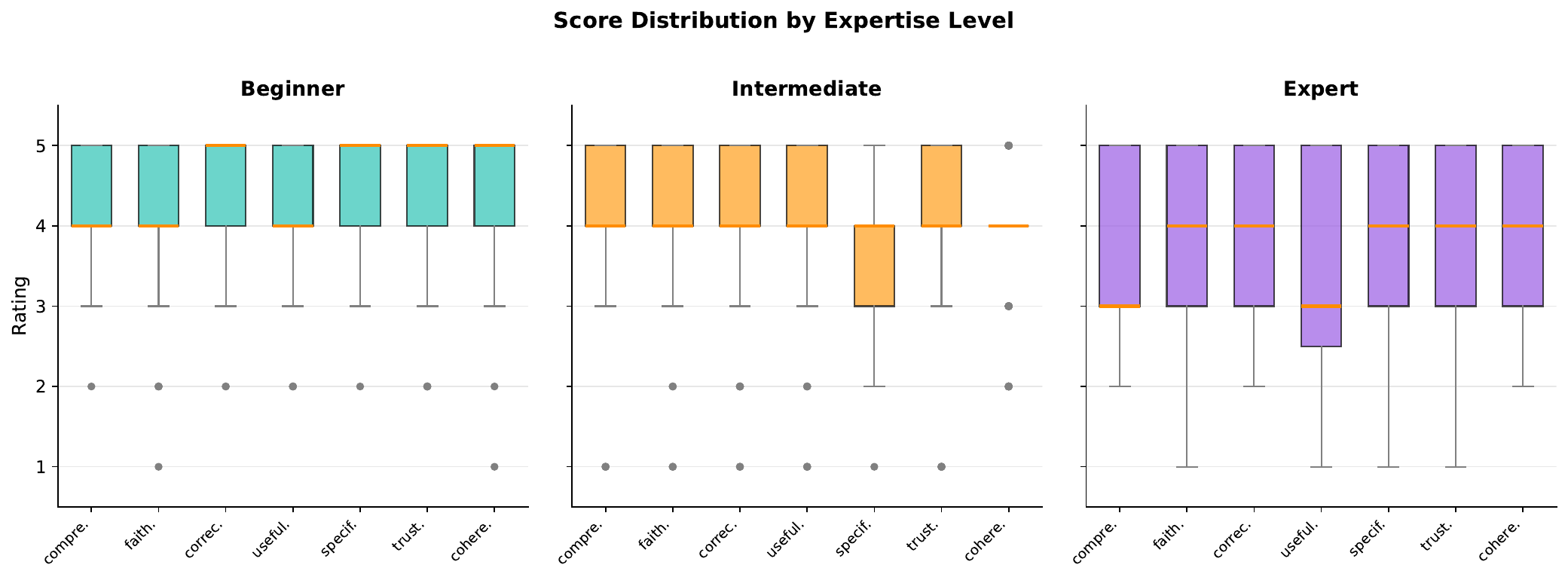}
\caption{Score distribution by expertise level for CNN system. Box plots reveal experts (purple) have wider score distributions, reflecting more nuanced evaluation criteria.}
\label{fig:score_dist_cnn}
\end{figure}

Figure~\ref{fig:score_dist_cnn} shows score distributions via box plots, revealing that experts exhibit higher variance in their ratings, particularly for Trustworthiness and Usefulness.

\begin{figure}[h]
\centering
\includegraphics[width=\columnwidth]{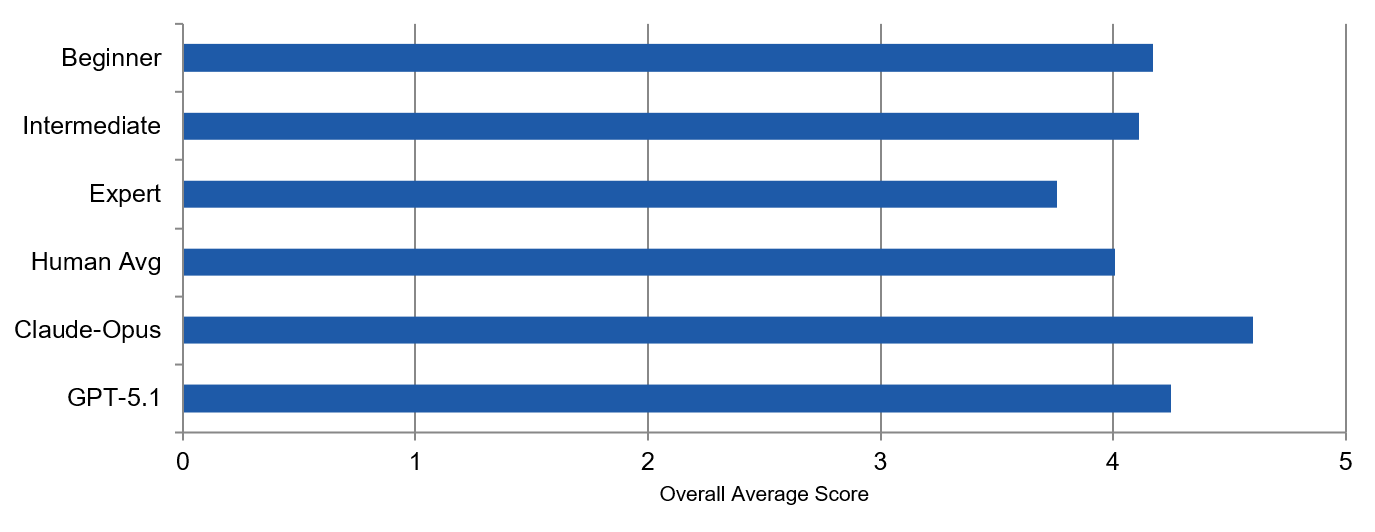}
\caption{VAE system average evaluation scores across all seven metrics. 
All dimensions exceed 3.8/5.0, with Coherence (4.16) and Specificity (4.14) 
rated highest; Usefulness (3.82) rated lowest but still above neutral.}
\label{fig:vae_average}
\end{figure}

Figure~\ref{fig:vae_average} presents the average scores across metrics for the VAE system.

\subsection{Inter-Rater Reliability Analysis}
\label{app:irr}

To assess the consistency of human evaluations, we compute standard inter-rater reliability (IRR) metrics across all evaluators. Tables~\ref{tab:irr_vae} and~\ref{tab:irr_cnn} present comprehensive IRR statistics for both systems.

\subsubsection{IRR Metrics}

We report four complementary reliability measures:
\begin{itemize}
    \item \textbf{ICC(2,1)}: Intraclass Correlation Coefficient for single rater reliability under a two-way random effects model~\citep{shrout1979intraclass}.
    \item \textbf{ICC(2,k)}: ICC for the average of $k$ raters, indicating reliability of mean ratings.
    \item \textbf{Krippendorff's $\alpha$}: A versatile reliability coefficient suitable for ordinal data~\citep{krippendorff2011computing}.
    \item \textbf{Fleiss' $\kappa$}: Multi-rater extension of Cohen's kappa for categorical agreement~\citep{fleiss1971measuring}.
\end{itemize}

We additionally report pairwise agreement rates: exact agreement (identical ratings) and within-1 agreement (ratings differing by at most 1 point on the 5-point scale).

\subsubsection{VAE System Inter-Rater Reliability}

\begin{table}[H]
\centering
\caption{Inter-rater reliability metrics for VAE system (9 evaluators, 315 evaluations). Low traditional IRR metrics reflect systematic expertise-level calibration differences rather than random disagreement; within-1 agreement exceeds 75\% across all dimensions.}
\label{tab:irr_vae}
\small
\begin{tabular}{lcccccc}
\toprule
\textbf{Metric} & \textbf{ICC(2,1)} & \textbf{ICC(2,k)} & \textbf{Kripp. $\alpha$} & \textbf{Fleiss' $\kappa$} & \textbf{Exact} & \textbf{Within-1} \\
\midrule
Comprehensibility & 0.046 & 0.301 & 0.025 & $-$0.024 & 33.8\% & 81.4\% \\
Faithfulness & 0.063 & 0.376 & 0.008 & $-$0.030 & 30.3\% & 75.7\% \\
Correctness & 0.079 & 0.435 & 0.020 & $-$0.019 & 32.1\% & 79.1\% \\
Usefulness & 0.108 & 0.520 & 0.080 & 0.001 & 33.8\% & 78.1\% \\
Specificity & 0.077 & 0.429 & 0.018 & $-$0.027 & 32.4\% & 79.6\% \\
Trustworthiness & 0.040 & 0.272 & $-$0.001 & $-$0.026 & 29.2\% & 76.0\% \\
Coherence & 0.047 & 0.308 & $-$0.001 & $-$0.029 & 32.9\% & 79.2\% \\
\midrule
\textbf{Average} & \textbf{0.066} & \textbf{0.377} & \textbf{0.021} & \textbf{$-$0.022} & \textbf{32.1\%} & \textbf{78.4\%} \\
\bottomrule
\end{tabular}
\end{table}

\begin{figure}[h]
\centering
\includegraphics[width=\columnwidth]{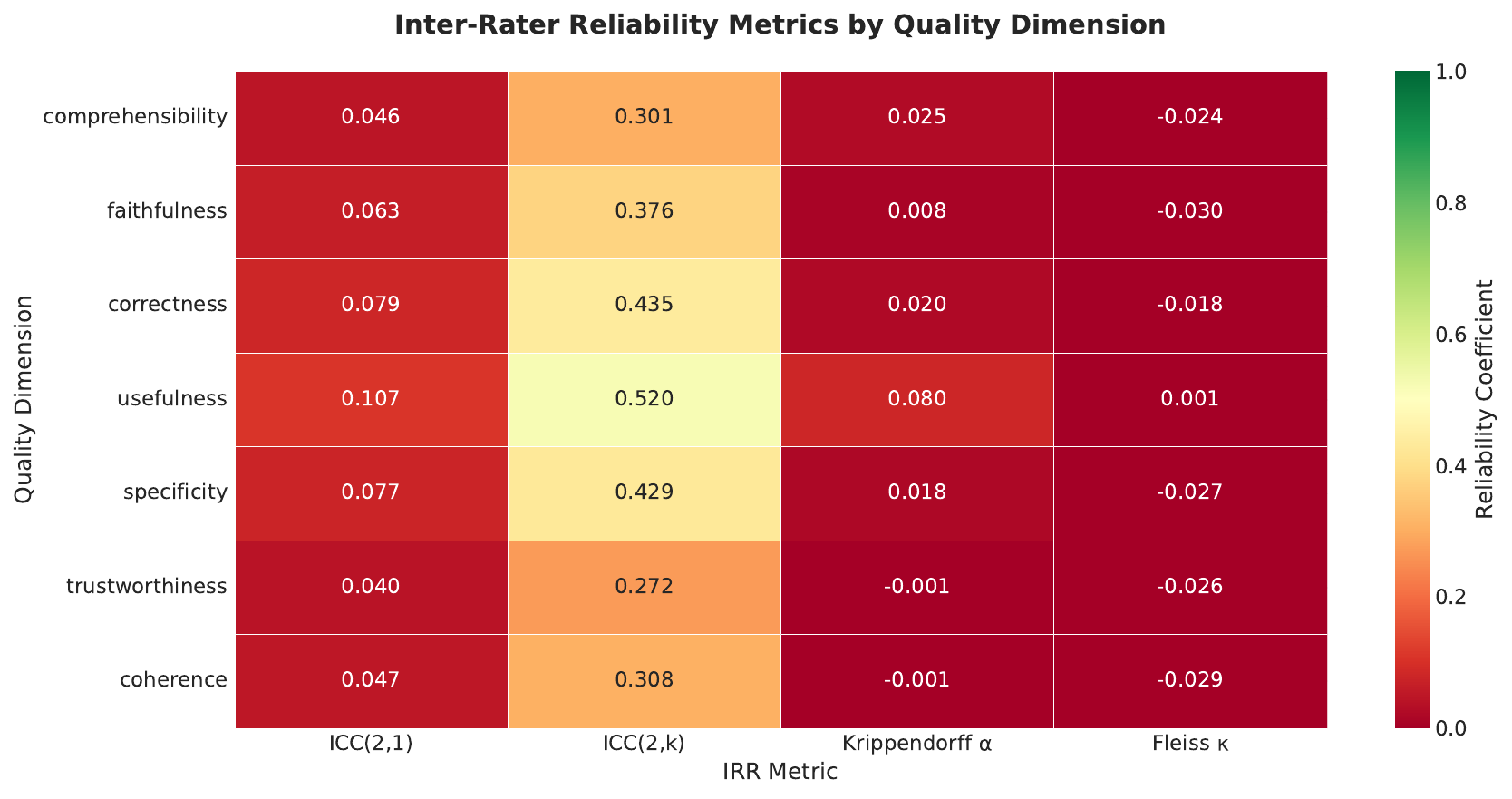}
\caption{VAE system inter-rater reliability heatmap. Traditional metrics (ICC, Krippendorff's $\alpha$, Fleiss' $\kappa$) show low values due to systematic expertise-level calibration differences, not random disagreement.}
\label{fig:irr_heatmap_vae}
\end{figure}

\begin{figure}[h]
\centering
\includegraphics[width=\columnwidth]{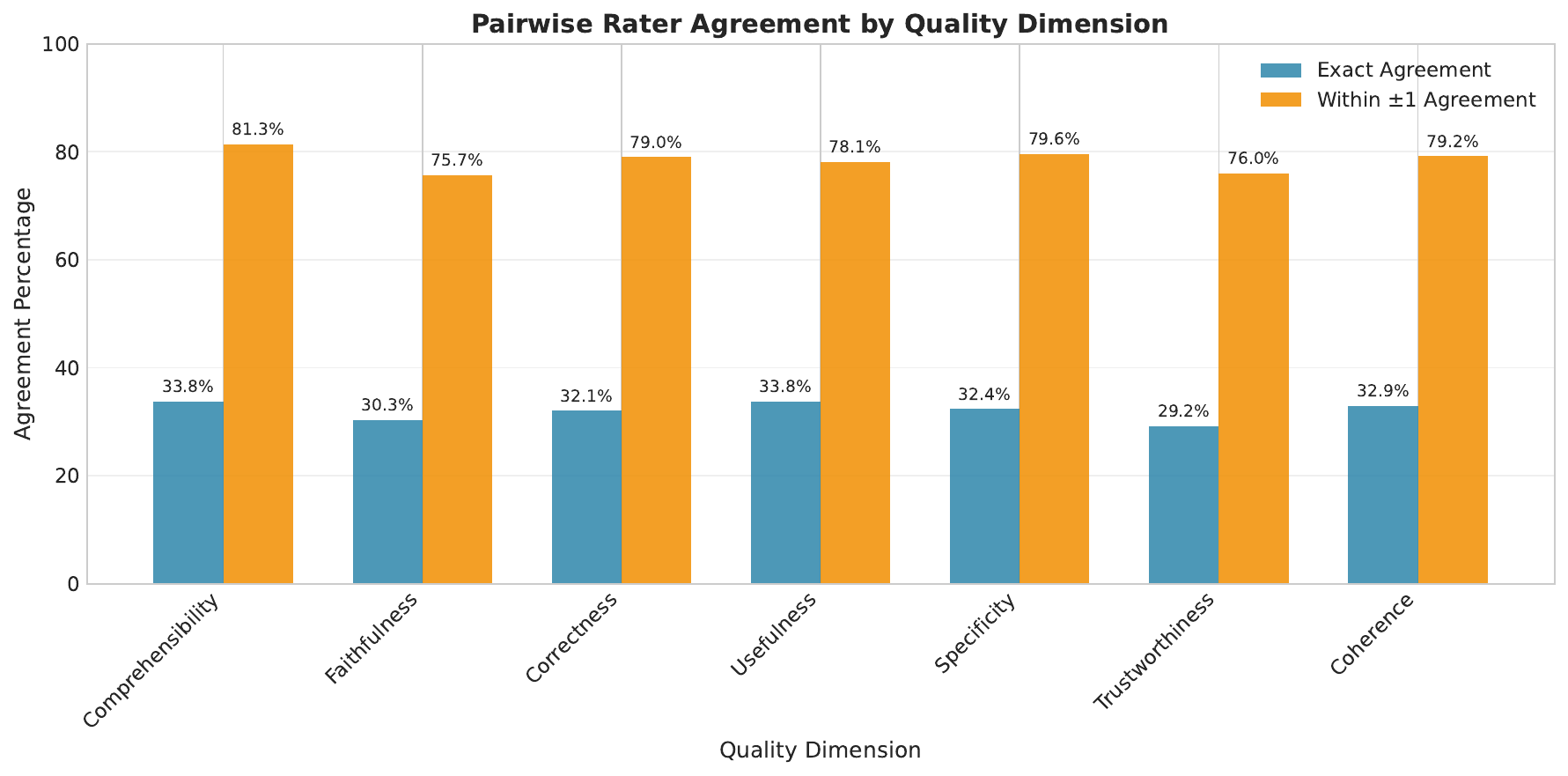}
\caption{VAE system pairwise agreement rates by quality dimension. While exact agreement averages 32.1\%, within-1 agreement exceeds 75\% for all metrics.}
\label{fig:agreement_by_metric_vae}
\end{figure}

\begin{figure}[h]
\centering
\includegraphics[width=0.9\columnwidth]{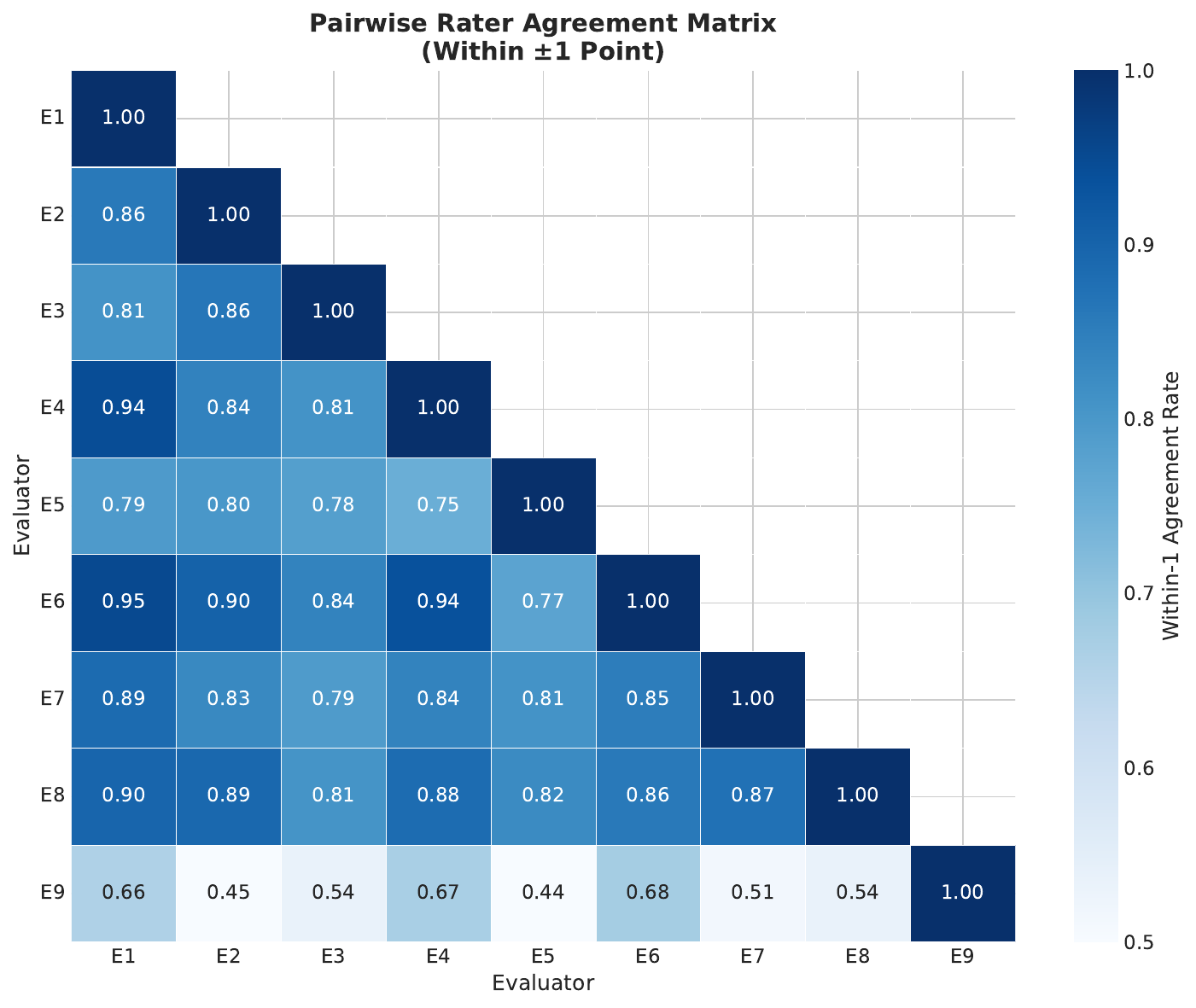}
\caption{VAE system pairwise rater agreement matrix (within $\pm$1 point). E1--E9 represent anonymized evaluators.}
\label{fig:agreement_matrix_vae}
\end{figure}

\subsubsection{CNN System Inter-Rater Reliability}

Table~\ref{tab:irr_cnn} presents inter-rater reliability metrics for the CNN system across 9 evaluators and 297 total evaluations. The pattern closely mirrors VAE results, with low traditional IRR metrics but strong within-1 agreement (75.9\% average), confirming that the expertise-driven calibration effects observed in VAE evaluation generalize across architectures.

\begin{table}[H]
\centering
\caption{Inter-rater reliability metrics for CNN system (9 evaluators, 297 evaluations). Pattern mirrors VAE results, confirming that expertise-driven calibration differences are consistent across architectures.}
\label{tab:irr_cnn}
\small
\begin{tabular}{lcccccc}
\toprule
\textbf{Metric} & \textbf{ICC(2,1)} & \textbf{ICC(2,k)} & \textbf{Kripp. $\alpha$} & \textbf{Fleiss' $\kappa$} & \textbf{Exact} & \textbf{Within-1} \\
\midrule
Comprehensibility & 0.030 & 0.220 & $-$0.035 & $-$0.004 & 31.0\% & 72.6\% \\
Faithfulness & 0.056 & 0.348 & $-$0.032 & $-$0.024 & 33.5\% & 80.3\% \\
Correctness & 0.071 & 0.406 & 0.002 & $-$0.004 & 34.4\% & 77.9\% \\
Usefulness & 0.094 & 0.483 & 0.035 & 0.004 & 30.1\% & 72.3\% \\
Specificity & 0.045 & 0.299 & $-$0.012 & $-$0.014 & 32.5\% & 78.9\% \\
Trustworthiness & 0.082 & 0.446 & $-$0.003 & $-$0.010 & 31.1\% & 72.5\% \\
Coherence & 0.039 & 0.266 & $-$0.016 & $-$0.011 & 31.9\% & 76.6\% \\
\midrule
\textbf{Average} & \textbf{0.060} & \textbf{0.353} & \textbf{$-$0.009} & \textbf{$-$0.009} & \textbf{32.1\%} & \textbf{75.9\%} \\
\bottomrule
\end{tabular}
\end{table}

\begin{figure}[t]
\centering
\includegraphics[width=\columnwidth]{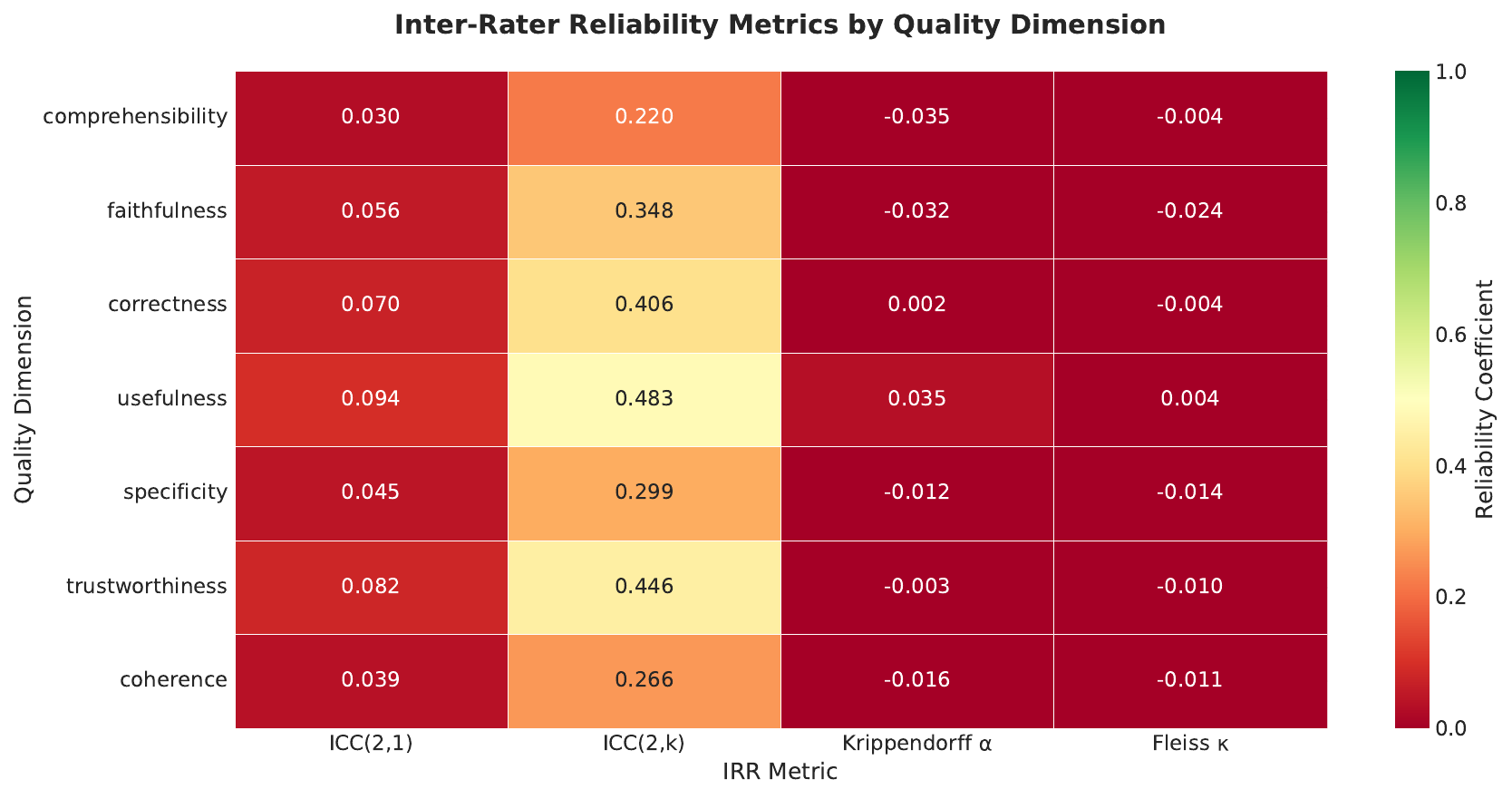}
\caption{CNN system inter-rater reliability heatmap. Pattern mirrors VAE results, with low traditional metrics reflecting systematic expertise effects.}
\label{fig:irr_heatmap_cnn}
\end{figure}

\begin{figure}[h]
\centering
\includegraphics[width=\columnwidth]{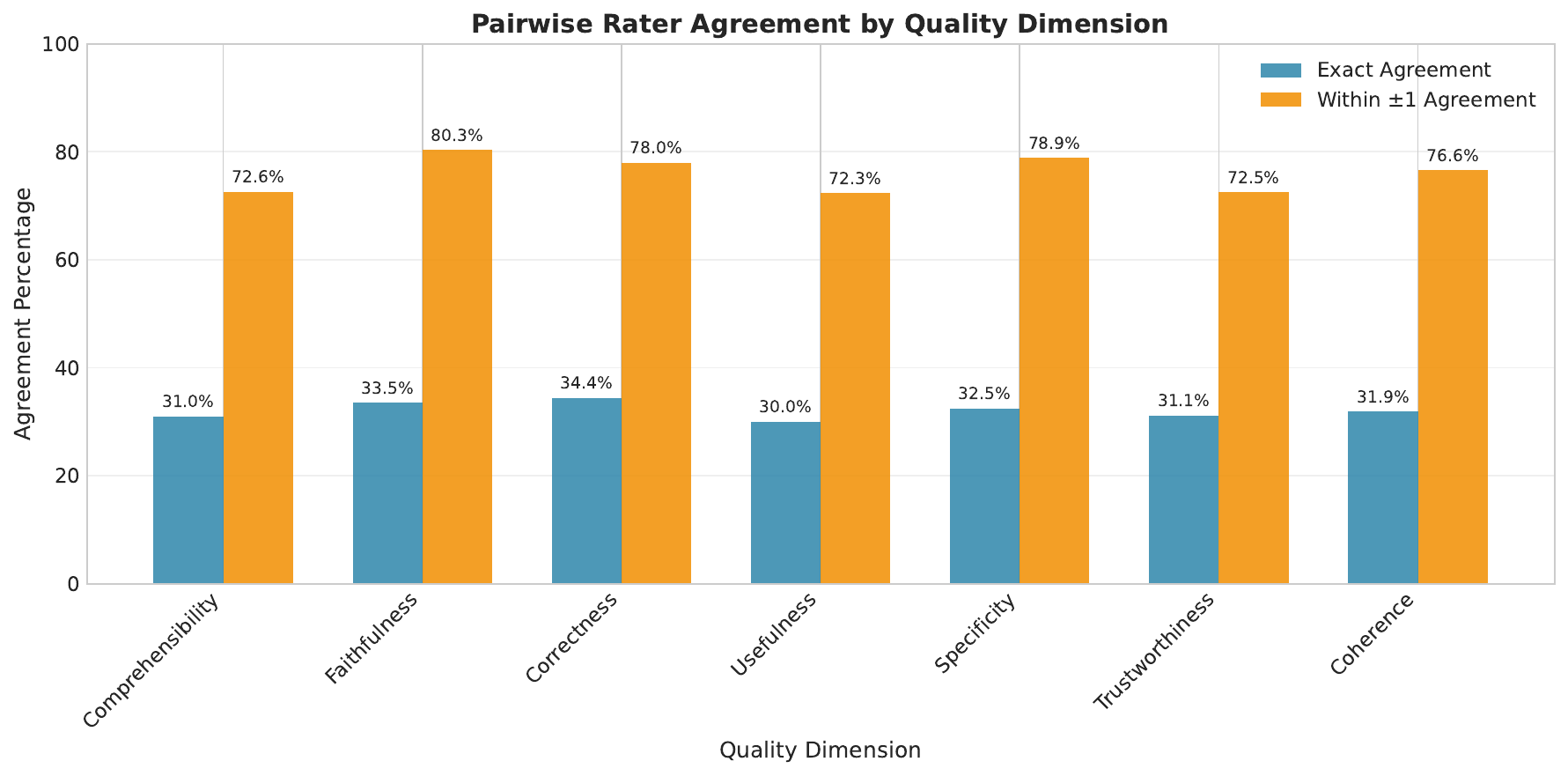}
\caption{CNN system pairwise agreement rates by quality dimension. Within-1 agreement exceeds 72\% for all metrics.}
\label{fig:agreement_by_metric_cnn}
\end{figure}

\begin{figure}[h]
\centering
\includegraphics[width=0.9\columnwidth]{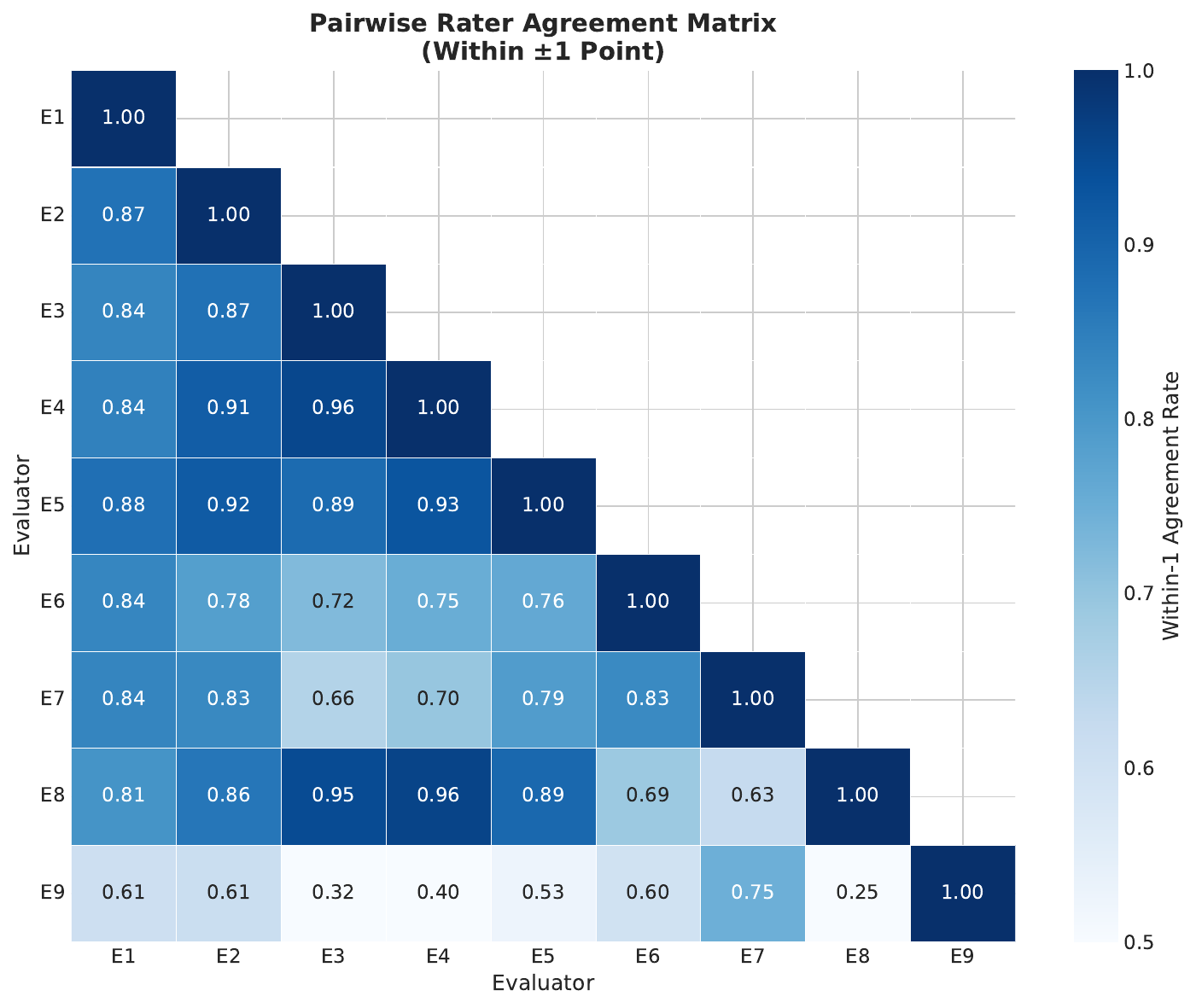}
\caption{CNN system pairwise rater agreement matrix (within $\pm$1 point). E1--E9 represent anonymized evaluators.}
\label{fig:agreement_matrix_cnn}
\end{figure}

\subsubsection{Interpretation}

Following established benchmarks~\citep{cicchetti1994guidelines, landis1977measurement}, ICC values below 0.40 indicate poor reliability, while Krippendorff's $\alpha$ below 0.667 suggests only tentative conclusions. Our results show low traditional IRR metrics (average ICC(2,1) = 0.066 for VAE, 0.060 for CNN), which initially appears concerning. However, this pattern reflects a well-documented phenomenon in subjective evaluation studies: \textit{systematic rater calibration differences} rather than random disagreement~\citep{graham2015accurate}.

\subsubsection{Understanding Low Traditional IRR with High Practical Agreement}

Three key observations explain this apparent paradox:

\textbf{(1) Systematic Expertise Effect:} As documented in Sec.~\ref{app:human_eval}, experts consistently rate 0.4--0.6 points lower than beginners across all metrics. This systematic offset inflates between-rater variance, reducing ICC and Krippendorff's $\alpha$, even though raters agree on relative quality ordering.

\textbf{(2) High Within-1 Agreement:} Despite low traditional metrics, within-1 agreement rates of 78.4\% (VAE) and 75.9\% (CNN) demonstrate that evaluators rarely disagree by more than one point. For 5-point Likert scales, this represents strong practical agreement---a difference between rating 4 vs.\ 5 is qualitatively minor.

\textbf{(3) Metric Ranking Concordance:} Kendall's $W$ coefficients of 0.260 (VAE) and 0.083 (CNN) indicate that while absolute scores differ by expertise level, evaluators show agreement on which quality dimensions perform better or worse. All groups consistently rate Coherence and Specificity highest, while Usefulness and Trustworthiness receive the lowest ratings.

\subsubsection{Precedent in XAI Evaluation Literature}

Low traditional IRR metrics are common in explainable AI evaluation studies. \citet{hoffman2023measures} report similar patterns when evaluating explanation satisfaction, noting that expertise-driven calibration differences are expected and informative. The systematic nature of our expertise effect---where experts apply stricter but internally consistent standards---validates our stratified evaluation design rather than indicating measurement failure.

\subsubsection{Summary}

Our IRR analysis reveals that traditional IRR metrics (ICC $\approx$ 0.06, Krippendorff's $\alpha$ $\approx$ 0.01) are low due to systematic expertise-level calibration, not random noise. Within-1 agreement rates of 76--78\% indicate strong practical consistency. The expertise gap is a feature, not a bug: experts appropriately apply stricter evaluation criteria to AI-generated explanations. Cross-architecture consistency (similar patterns for VAE and CNN) supports the reliability of our evaluation methodology. We recommend that future XAI evaluation studies report both traditional IRR metrics and practical agreement rates, while stratifying by evaluator expertise to capture meaningful calibration differences.

\section{LLM-as-Judge Evaluation}
\label{app:llm_judge}

We compare human evaluations with two LLM judges: Claude-Opus and GPT-5.1. Table~\ref{tab:llm_judge_vae} and Table~\ref{tab:llm_judge_cnn} present the full comparison.

\begin{table}[h]
\centering
\caption{LLM-as-judge evaluation for VAE system. Both LLM judges rate higher than humans, with Claude-Opus showing optimism bias of +0.59 points overall.}
\label{tab:llm_judge_vae}
\small
\begin{tabular}{lccccc}
\toprule
\textbf{Metric} & \textbf{Human} & \textbf{Claude} & \textbf{GPT-5.1} & \textbf{$\Delta$Claude} & \textbf{$\Delta$GPT} \\
\midrule
Comprehensibility & 4.01 & 4.46 & 4.77 & +0.45 & +0.76 \\
Faithfulness & 4.03 & 4.74 & 3.94 & +0.71 & $-$0.09 \\
Correctness & 4.05 & 4.57 & 3.86 & +0.52 & $-$0.19 \\
Usefulness & 3.82 & 4.31 & 3.94 & +0.49 & +0.12 \\
Specificity & 4.14 & 4.60 & 4.40 & +0.46 & +0.26 \\
Trustworthiness & 3.90 & 4.60 & 3.89 & +0.70 & $-$0.01 \\
Coherence & 4.16 & 4.91 & 4.94 & +0.75 & +0.78 \\
\midrule
\textbf{Overall} & \textbf{4.01} & \textbf{4.60} & \textbf{4.25} & \textbf{+0.59} & \textbf{+0.24} \\
\bottomrule
\end{tabular}
\end{table}
\paragraph{GPT-2 BM25 baseline (oracle recall@$k$).}
Table~\ref{tab:gpt2_bm25} reports BM25 lexical retrieval on the
GPT-2 ESRDGT schema, using the same oracle set (7 canonical IOI
name-mover heads). BM25 recall@30 falls \emph{below} the
random-partition mean, confirming that keyword-based retrieval
provides no signal on functional-concept queries where surface
form is uninformative.

\begin{table}[h]
\centering
\caption{LLM-as-judge evaluation for CNN system. Claude-Opus maintains consistent optimism bias (+0.47 points); GPT-5.1 shows similar pattern with +0.22 points overall.}
\label{tab:llm_judge_cnn}
\small
\begin{tabular}{lccccc}
\toprule
\textbf{Metric} & \textbf{Human} & \textbf{Claude} & \textbf{GPT-5.1} & \textbf{$\Delta$Claude} & \textbf{$\Delta$GPT} \\
\midrule
Comprehensibility & 4.07 & 4.51 & 4.69 & +0.44 & +0.62 \\
Faithfulness & 4.06 & 4.68 & 4.03 & +0.62 & $-$0.03 \\
Correctness & 4.05 & 4.49 & 3.91 & +0.44 & $-$0.14 \\
Usefulness & 3.92 & 4.37 & 4.11 & +0.45 & +0.19 \\
Specificity & 4.13 & 4.54 & 4.31 & +0.41 & +0.18 \\
Trustworthiness & 3.97 & 4.57 & 3.94 & +0.60 & $-$0.03 \\
Coherence & 4.13 & 4.49 & 4.89 & +0.36 & +0.76 \\
\midrule
\textbf{Overall} & \textbf{4.05} & \textbf{4.52} & \textbf{4.27} & \textbf{+0.47} & \textbf{+0.22} \\
\bottomrule
\end{tabular}
\end{table}

The optimism bias is consistent with documented LLM judge tendencies toward fluent, well-structured responses~\citep{zheng2023judging}. Importantly, the relative ordering of metrics is preserved: both humans and LLM judges rate Comprehensibility and Coherence highest, while Usefulness and Trustworthiness receive lower (but still positive) ratings.

\begin{figure}[h]
\centering
\includegraphics[width=\columnwidth]{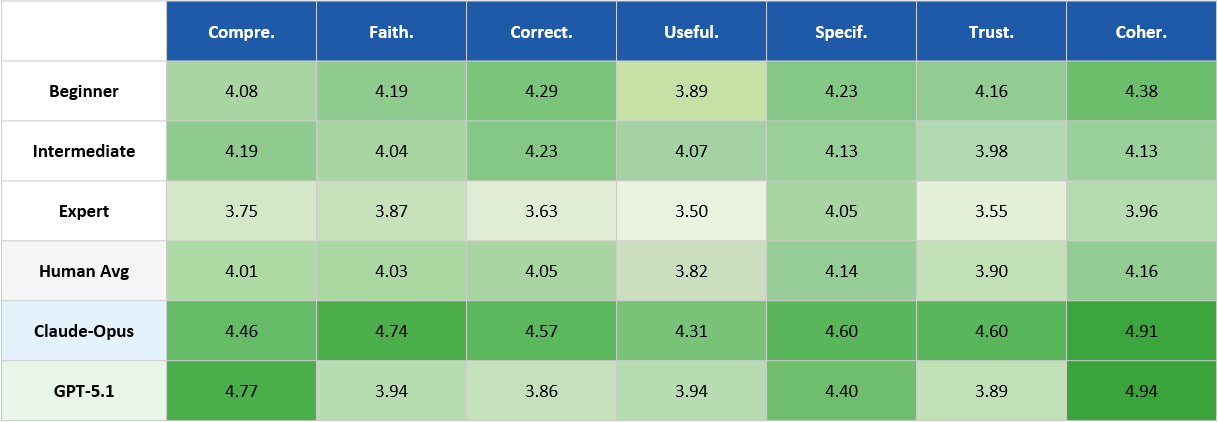}
\caption{VAE system evaluation heatmap across all evaluator groups. Darker green indicates higher scores. Claude-Opus consistently rates highest; experts rate most critically.}
\label{fig:heatmap_vae}
\end{figure}

\begin{figure}[h]
\centering
\includegraphics[width=\columnwidth]{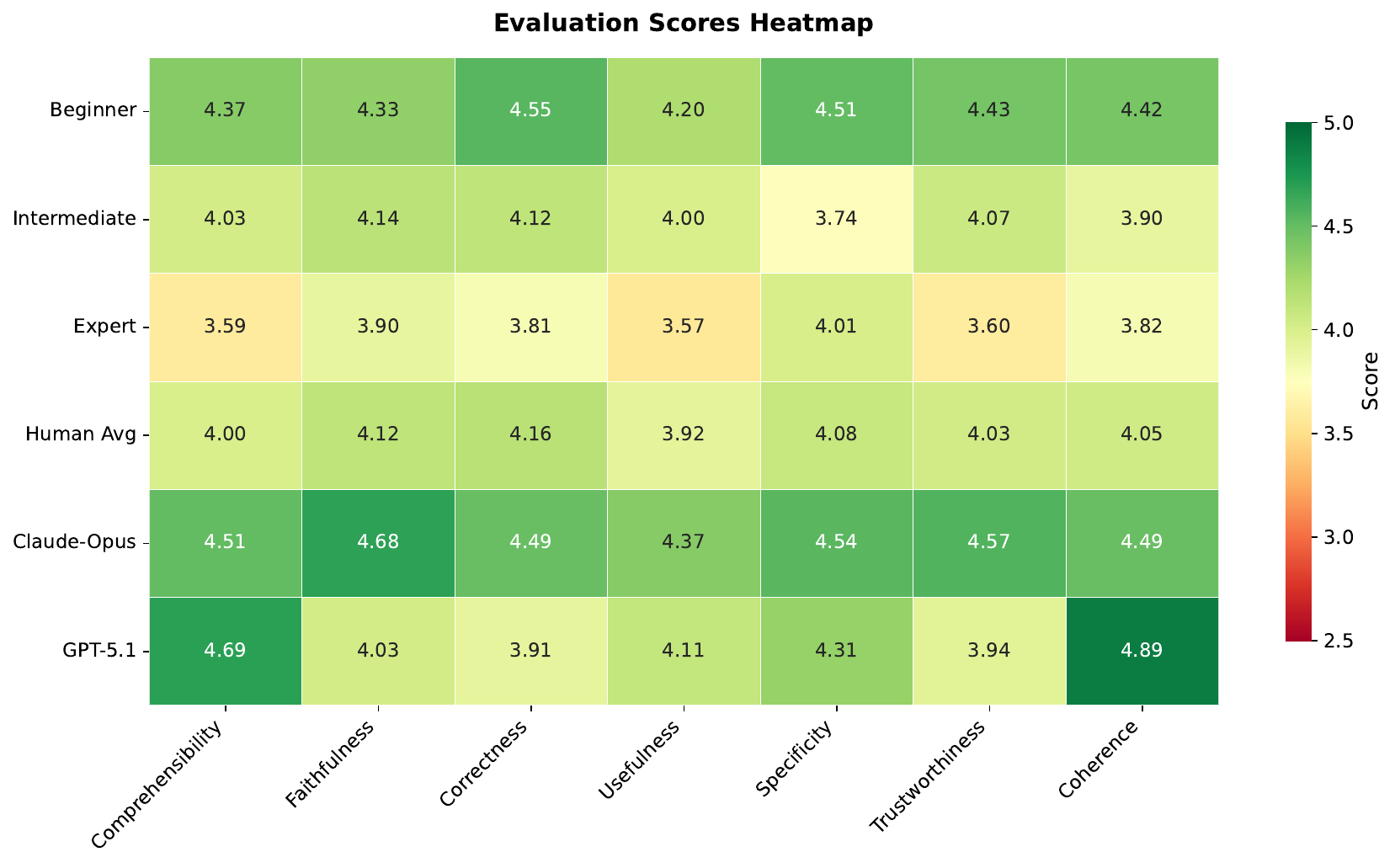}
\caption{CNN system evaluation heatmap across all evaluator groups. Pattern mirrors VAE results with LLM judges showing systematic optimism bias.}
\label{fig:heatmap_cnn}
\end{figure}

Figure~\ref{fig:heatmap_vae} and Figure~\ref{fig:heatmap_cnn} present heatmap visualizations comparing all evaluator groups across metrics.

\begin{figure}[h]
\centering
\includegraphics[width=\columnwidth]{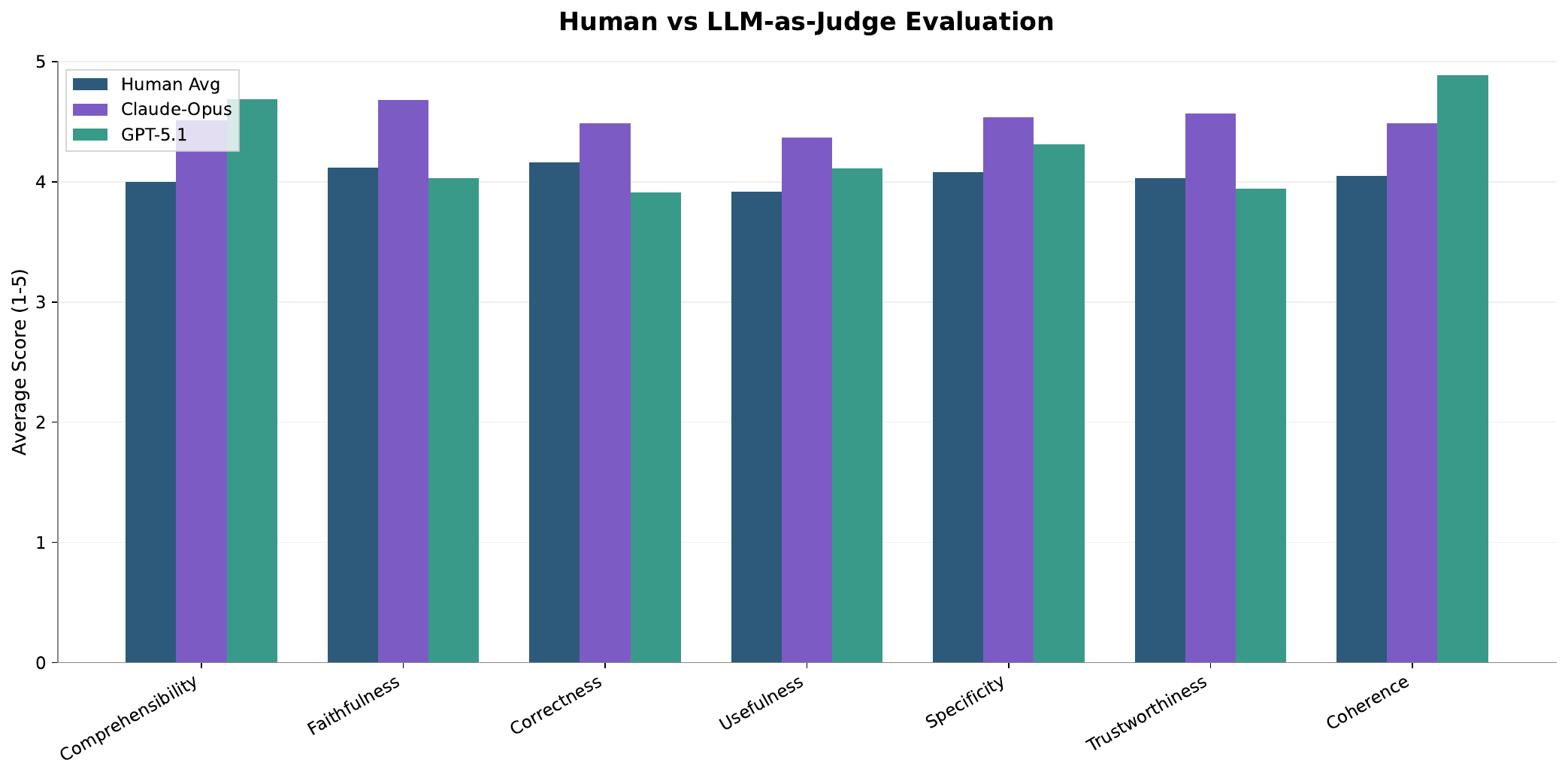}
\caption{Human vs LLM-as-Judge comparison for CNN system. Human averages (blue) are consistently lower than Claude-Opus (purple) and GPT-5.1 (green) across most metrics.}
\label{fig:human_llm_comparison}
\end{figure}

Figure~\ref{fig:human_llm_comparison} directly compares human average scores with both LLM judges for the CNN system.

\begin{figure}[h]
\centering
\includegraphics[width=0.9\columnwidth]{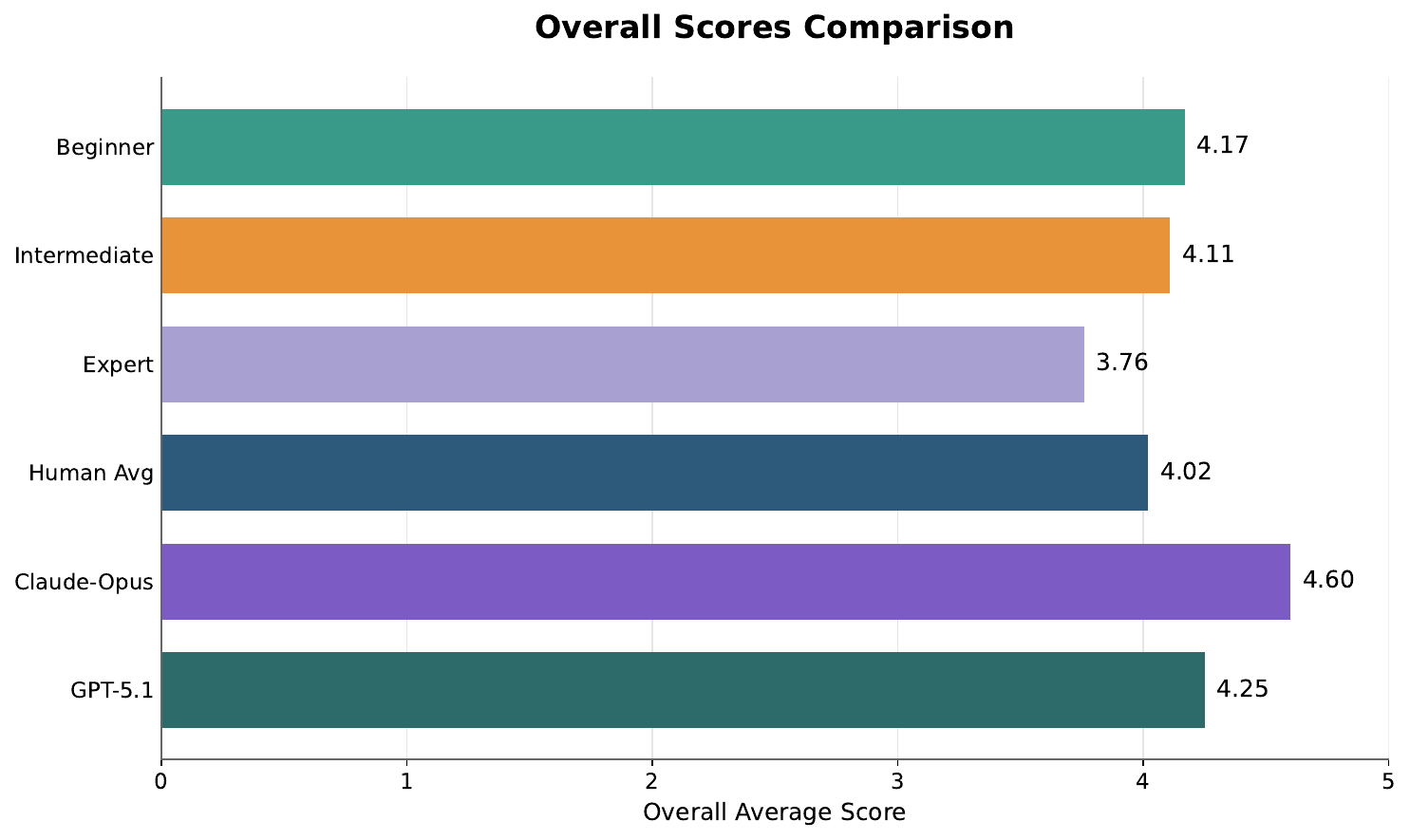}
\caption{Overall scores comparison for VAE system. Claude-Opus achieves highest overall (4.60), followed by GPT-5.1 (4.25), beginners (4.17), intermediates (4.11), human average (4.02), and experts (3.76).}
\label{fig:vae_overall}
\end{figure}

\begin{figure}[h]
\centering
\includegraphics[width=0.9\columnwidth]{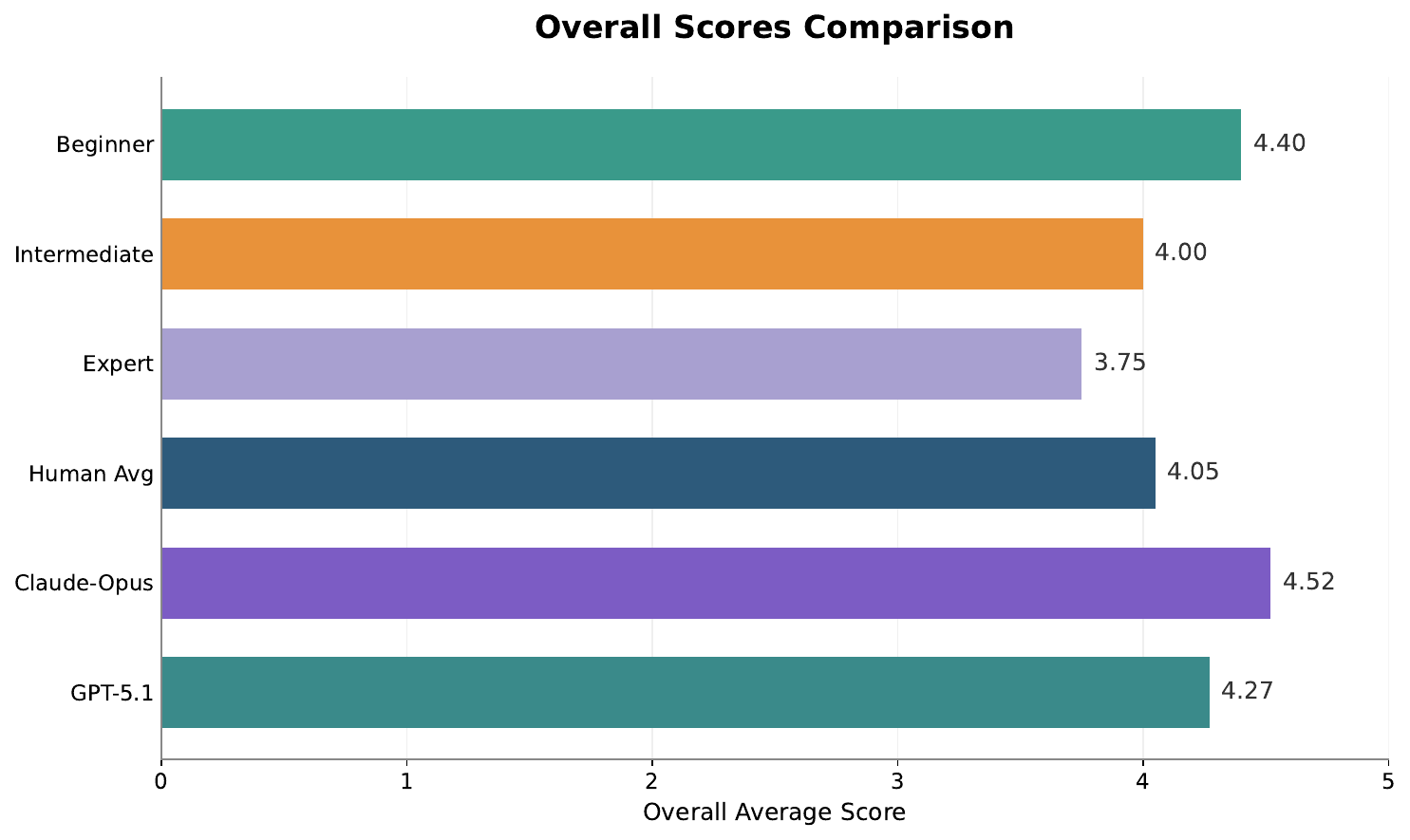}
\caption{Overall scores comparison for CNN system. Claude-Opus achieves highest overall (4.52), followed by beginners (4.40), GPT-5.1 (4.27), intermediates (4.00), human average (4.05), and experts (3.75).}
\label{fig:cnn_overall}
\end{figure}

Figure~\ref{fig:vae_overall} and Figure~\ref{fig:cnn_overall} show overall score comparisons across all evaluator types.

\subsection{Key Findings}

\textbf{Expertise Effect:} Experts consistently rate 0.4--0.6 points lower than beginners across both systems, with largest gaps in Trustworthiness and Usefulness. This reflects appropriate skepticism about AI-generated explanations from knowledgeable users.

\textbf{LLM Optimism Bias:} Both LLM judges exhibit systematic optimism, with Claude-Opus showing larger bias (+0.5--0.6 points) than GPT-5.1 (+0.2--0.3 points). GPT-5.1 occasionally rates below human averages for Faithfulness and Correctness.

\textbf{Metric Ordering Preservation:} Despite absolute score differences, relative metric rankings remain consistent across evaluator types. Coherence and Comprehensibility receive highest ratings; Usefulness and Trustworthiness receive lowest.

\textbf{Cross-Architecture Consistency:} Both VAE and CNN systems show similar evaluation patterns, supporting the framework's architecture-agnostic design.

\section{Automatic Evaluation Details}
\label{app:auto_eval}
Table~\ref{tab:auto_eval_full} presents comprehensive automatic evaluation results across both architectures, LLM backends, and retrieval depths $k \in \{5, 10, 20\}$.
\begin{table}[h]
\centering
\caption{Comprehensive automatic evaluation results. Bold: best per architecture-$k$ combination. All systems achieve high retrieval recall and faithfulness, with Qwen showing strongest generation quality across both architectures.}
\label{tab:auto_eval_full}
\resizebox{\textwidth}{!}{
\begin{tabular}{lllcccccccccccc}
\toprule
& & & \multicolumn{4}{c}{Retrieval} & \multicolumn{3}{c}{Generation} & \multicolumn{3}{c}{Semantic (BERT)} & \multicolumn{2}{c}{Faithfulness} \\
\cmidrule(lr){4-7} \cmidrule(lr){8-10} \cmidrule(lr){11-13} \cmidrule(lr){14-15}
Arch & LLM & $k$ & Prec & Rec & F1 & NDCG & Gen-F1 & BLEU & R-1 & P & R & F1 & Attr & F-Acc \\
\midrule
\multirow{9}{*}{VAE} 
& \multirow{3}{*}{Qwen} 
    & 5  & 89.6 & 100 & 94.5 & 100.0 & \textbf{47.1} & \textbf{14.1} & \textbf{51.3} & \textbf{83.5} & \textbf{87.1} & \textbf{85.3} & \textbf{64.5} & 97.9 \\
&   & 10 & 56.2 & 100 & 72.0 & 98.8  & \textbf{45.9} & \textbf{13.4} & \textbf{50.0} & \textbf{83.1} & \textbf{85.7} & \textbf{84.4} & 59.1 & 98.0 \\
&   & 20 & 46.8 & 100 & 63.8 & 96.4  & \textbf{38.3} & \textbf{9.2}  & \textbf{41.8} & \textbf{82.9} & \textbf{84.3} & \textbf{83.6} & \textbf{61.1} & 98.4 \\
\cmidrule(lr){2-15}
& \multirow{3}{*}{Claude} 
    & 5  & 89.6 & 100 & 94.5 & 100.0 & 43.8 & 12.0 & 47.8 & 82.1 & 86.1 & 84.1 & 47.1 & \textbf{99.2} \\
&   & 10 & 56.2 & 100 & 72.0 & 98.8  & 42.4 & 11.0 & 46.8 & 81.8 & 84.8 & 83.3 & 50.8 & \textbf{99.3} \\
&   & 20 & 46.8 & 100 & 63.8 & 96.4  & 35.7 & 6.0  & 39.9 & 82.0 & 83.8 & 82.9 & 54.0 & \textbf{99.6} \\
\cmidrule(lr){2-15}
& \multirow{3}{*}{GPT} 
    & 5  & 89.6 & 100 & 94.5 & 100.0 & 32.7 & 7.7  & 34.6 & 78.0 & 82.4 & 80.2 & 37.8 & 95.3 \\
&   & 10 & 56.2 & 100 & 72.0 & 98.8  & 35.9 & 8.4  & 38.3 & 81.1 & 84.4 & 82.7 & 43.1 & 98.6 \\
&   & 20 & 46.8 & 100 & 63.8 & 96.4  & 33.2 & 6.0  & 35.7 & 80.8 & 83.1 & 81.9 & 42.8 & 99.3 \\
\midrule
\multirow{9}{*}{CNN} 
& \multirow{3}{*}{Qwen} 
    & 5  & 96.8 & 100 & 98.4 & 100.0 & \textbf{50.8} & \textbf{13.8} & \textbf{54.2} & \textbf{86.8} & \textbf{87.5} & \textbf{87.2} & 66.7 & 96.8 \\
&   & 10 & 48.4 & 100 & 65.2 & 100.0 & \textbf{49.6} & \textbf{13.9} & \textbf{53.1} & \textbf{87.1} & \textbf{87.5} & \textbf{87.3} & \textbf{70.1} & 97.1 \\
&   & 20 & 24.2 & 100 & 39.0 & 100.0 & \textbf{41.2} & \textbf{11.9} & \textbf{49.5} & \textbf{87.1} & \textbf{86.7} & \textbf{86.9} & \textbf{67.2} & 94.4 \\
\cmidrule(lr){2-15}
& \multirow{3}{*}{Claude} 
    & 5  & 96.8 & 100 & 98.4 & 100.0 & 46.3 & 12.3 & 51.1 & 82.4 & 87.0 & 86.4 & 64.3 & \textbf{99.3} \\
&   & 10 & 48.4 & 100 & 65.2 & 100.0 & 44.6 & 12.2 & 50.0 & 84.7 & 85.4 & 84.7 & 66.8 & \textbf{99.5} \\
&   & 20 & 24.2 & 100 & 39.0 & 100.0 & 39.3 & 7.2  & 43.5 & 85.4 & 84.3 & 84.2 & 66.4 & \textbf{99.6} \\
\cmidrule(lr){2-15}
& \multirow{3}{*}{GPT} 
    & 5  & 96.8 & 100 & 98.4 & 100.0 & 35.5 & 9.5  & 38.0 & 80.0 & 81.5 & 82.4 & 61.6 & 96.5 \\
&   & 10 & 48.4 & 100 & 65.2 & 100.0 & 37.8 & 7.9  & 41.3 & 83.6 & 83.3 & 83.9 & 63.7 & 97.9 \\
&   & 20 & 24.2 & 100 & 39.0 & 100.0 & 36.2 & 6.8  & 38.7 & 81.3 & 82.7 & 83.1 & 62.3 & 99.0 \\
\bottomrule
\end{tabular}
}
\end{table}

Key observations:
\begin{itemize}
    \item The CNN system achieves higher retrieval precision than VAE (96.8\% vs.\ 89.6\% at $k=5$) due to the discrete nature of filter--class associations compared to continuous VAE dimensions.
    \item Precision naturally decreases as $k$ increases, while recall remains perfect (100\%) because gold sets are small (1--5 units) and consistently captured within the top-$k$ results.
    \item Qwen3-8B achieves the highest generation quality (Gen-F1 up to 50.8\%) and attribution scores (up to 70.1\%), suggesting stronger adherence to retrieved evidence.
    \item Claude-Opus demonstrates consistently superior factual accuracy (99.2--99.6\%) but lower attribution, indicating a more conservative generation style that prioritizes correctness.
    \item We hypothesize that Qwen's stronger generation performance may stem from its text-only design explicitly optimized for natural language tasking, whereas multimodal-capable LLMs such as Claude and GPT distribute representational capacity across multiple modalities, potentially reducing specialization for structured text-based retrieval and generation.
    \item Despite these differences, all configurations maintain high factual accuracy ($>94\%$), suggesting that grounding quality may matter more than model scale when retrieval provides correct and structured context.
\end{itemize}

\section{Full Ablation Results}
\label{app:ablation}

Table~\ref{tab:ablation_full} presents comprehensive ablation results across 14 configurations.

\textbf{Configuration descriptions.}
\begin{itemize}
    \item \textbf{Full System (Conversational)}: Complete MANIFESTATION with all components including conversation state management.
    \item \textbf{Manifestation (Base)}: Core system without conversational layer.
    \item \textbf{No Reranking}: Disables similarity threshold filtering; accepts all retrieved documents regardless of similarity score (threshold set to 0.0).
    \item \textbf{No Over-retrieval}: Retrieves exactly $k$ documents instead of $5k$ candidates with subsequent filtering (search multiplier set to 1).
    \item \textbf{No Exact Match Boosting}: Disables the 1.5$\times$ similarity boost for exact-match documents; exact matches still prioritized by flag but scores are not inflated.
    \item \textbf{Dims Index Only}: Uses only the dimension/filter index, excluding Q\&A and interaction indices.
    \item \textbf{No Query Reformulation}: Disables context-aware query reformulation; pronouns and references are not resolved using conversation history.
    \item \textbf{No Exact Match}: Disables entity-aware exact matching; dimension/filter identifiers are extracted for statistics only, not used for retrieval prioritization.
    \item \textbf{Semantic Search Only}: Pure embedding-based retrieval without exact matching, boosting, or threshold filtering.
    \item \textbf{Exact Match Only}: Retrieves only explicitly referenced entities via metadata lookup; no semantic expansion for conceptual queries.
    \item \textbf{Unstructured Text}: Same statistical content serialized as free-form prose without $(E, S, R, D, G)$ structure or identifier normalization.
    \item \textbf{Q\&A Index Only}: Uses only the pre-computed question-answer pair index.
    \item \textbf{Interactions Index Only}: Uses only the pairwise interaction/relationship index.
    \item \textbf{BM25 Hybrid}: Combines BM25 sparse retrieval (term-frequency-based) with dense semantic embeddings, merging results by combined score.
    \item \textbf{ID Normalized (Semantic)}: Prepends normalized identifier tokens (e.g., \texttt{DIM\_063}, \texttt{CONV2\_F23}) to document text before embedding, then performs semantic search.
\end{itemize}

Critical failures (bottom block) are configurations where removing a component causes dramatic performance collapse.

\begin{table}[h]
\centering
\caption{Comprehensive evaluation at $k{=}5$. Top block: MANIFESTATION
ablations (14 configurations for VAE; 7 applicable configurations
for CNN --- the CNN pipeline is non-conversational and uses a single
hybrid retriever, making VAE-specific conversational ablations
inapplicable). Bottom block: traditional retrieval baselines. Cells
marked ``--'' denote inapplicable configurations; see text for
explanation of the Manifestation (Base) CNN entry.}
\label{tab:ablation_full}
\resizebox{\textwidth}{!}{
\begin{tabular}{lcccccccc}
\toprule
& \multicolumn{4}{c}{VAE System} & \multicolumn{4}{c}{CNN System} \\
\cmidrule(lr){2-5} \cmidrule(lr){6-9}
Configuration & Prec & F-Acc & $\Delta$Prec & $\Delta$F-Acc & Prec & F-Acc & $\Delta$Prec & $\Delta$F-Acc \\
\midrule
Full System (Conversational) & 89.6 & 97.9 &   &   & 96.8 & 96.8 &   &   \\
Manifestation (Base) & 94.0 & 81.5 & $+4.4$ & $-16.4$ & 96.8$^\dagger$ & 96.8$^\dagger$ & 0.0 & 0.0 \\
No Reranking & 94.0 & 78.6 & +4.4 & $-$19.3 & 96.8 & 91.8 & 0.0 & $-$5.0 \\
No Over-retrieval & 94.0 & 74.6 & +4.4 & $-$23.3 & 96.8 & 95.0 & 0.0 & $-$1.8 \\
No Context Boosting & 94.0 & 74.7 & +4.4 & $-$23.2 & -- & -- & -- & -- \\
Similarity Off & 94.0 & 76.5 & +4.4 & $-$21.4 & -- & -- & -- & -- \\
Dims Index Only & 94.0 & 78.4 & +4.4 & $-$19.5 & -- & -- & -- & -- \\
No Query Reformulation & 94.0 & 20.3 & +4.4 & $-$77.6 & -- & -- & -- & -- \\
\midrule
\multicolumn{9}{l}{\textit{Critical failures (ablation):}} \\
No Exact Match & 4.8 & 21.8 & $-$84.8 & $-$76.1 & 26.0 & 59.1 & $-$70.8 & $-$37.7 \\
Semantic Search Only & 4.8 & 20.7 & $-$84.8 & $-$77.2 & 26.0 & 60.5 & $-$70.8 & $-$36.3 \\
Exact Match Only & 62.4 & 78.3 & $-$27.2 & $-$19.6 & 71.2 & 82.1 & $-$25.6 & $-$14.7 \\
Unstructured Text & 3.7 & 0.7 & $-$85.9 & $-$97.2 & 2.8 & 41.9 & $-$94.0 & $-$54.9 \\
Q\&A Index Only & 0.0 & 0.0 & $-$89.6 & $-$97.9 & -- & -- & -- & -- \\
Interactions Index Only & 0.0 & 0.0 & $-$89.6 & $-$97.9 & -- & -- & -- & -- \\
\midrule
\multicolumn{9}{l}{\textit{Traditional retrieval baselines:}} \\
BM25 Hybrid & 18.0 & 45.5 & $-$71.6 & $-$52.4 & 20.4 & 48.5 & $-$76.4 & $-$48.3 \\
ID Normalized (Semantic) & 3.2 & 19.9 & $-$86.4 & $-$78.0 & 17.2 & 73.2 & $-$79.6 & $-$23.6 \\
\bottomrule
\end{tabular}
}
\end{table}

\paragraph{Per-Query-Type Breakdown.}
Entity queries (e.g., ``What does dimension 47 encode?'') achieve 95.0\% (VAE) and 100.0\% (CNN) precision with exact matching, but only 5.0\% and 10.0\% with semantic search alone. Concept queries show the opposite: semantic search achieves 78.0\% (VAE) and 86.7\% (CNN), while exact matching drops to 20.0\% and 33.3\%. This demonstrates why hybrid retrieval is necessary---neither strategy alone handles both query types.

\begin{table}[h]
\centering
\caption{GPT-2 oracle recall@$k$ for BM25 (lexical) vs baselines.
Oracle set: 7 canonical IOI L7--9 name-mover heads. BM25 recall@30
$= 0.143$ falls below random ($0.304$), confirming surface-form
retrieval carries no signal on functional-concept queries.}
\label{tab:gpt2_bm25}
\small
\begin{tabular}{lccc}
\toprule
Baseline & recall@7 & recall@15 & recall@30 \\
\midrule
\textbf{Framework $S{+}R$} & \textbf{0.089} & \textbf{0.232} & \textbf{0.411} \\
Full-MU embed               & 0.071 & 0.179 & 0.339 \\
Random (mean)               & 0.018 & 0.143 & 0.304 \\
BM25 (lexical)              & 0.000 & 0.071 & 0.143 \\
\bottomrule
\end{tabular}
\end{table}

\end{document}